\PassOptionsToPackage{table}{xcolor}
\documentclass{article}
\usepackage{iclr2027_conference,times}
\usepackage{amsmath,amsfonts,bm}

\def\eqref#1{equation~\ref{#1}}

\def\1{\bm{1}}

\DeclareMathAlphabet{\mathsfit}{\encodingdefault}{\sfdefault}{m}{sl}
\SetMathAlphabet{\mathsfit}{bold}{\encodingdefault}{\sfdefault}{bx}{n}

\usepackage[hidelinks]{hyperref}
\usepackage{url}
\usepackage{booktabs}
\usepackage{amsfonts}       
\usepackage{nicefrac}       
\usepackage{microtype}      
\usepackage{xcolor} 
\usepackage{cleveref}
\usepackage{color, colortbl,graphicx, amssymb,enumitem}
\usepackage{tcolorbox}
\definecolor{LightCyan}{rgb}{0.88,1,1}
\usepackage{multirow} 
\usepackage{array}
\usepackage{longtable}
\usepackage{rotating}
\usepackage{graphicx, wrapfig}
\usepackage[table]{xcolor}
\usepackage{pifont}
\usepackage{arydshln}
\usepackage{caption}
\usepackage{dsfont}
\usepackage{roboto}
\usepackage{graphicx}
\usepackage{subcaption}

\newtheorem{theoremdef}{Theorem}
\newtheorem{definition}[theoremdef]{Definition}
\newtheorem{definition*}{Problem}

\definecolor{BestColor}{HTML}{FFF2CC}
\newcommand{\best}[1]{\cellcolor{BestColor}\textbf{#1}}
\hypersetup{colorlinks,linkcolor={red},citecolor={blue},urlcolor={red}}  

\title{\texttt{VLM4Cluster}: Benchmarking Image Clustering in the Era of Pre-trained Vision–Language Models}

\author{%
\begin{minipage}[t]{\dimexpr\textwidth-2\tabcolsep\relax}
\raggedright
{\normalfont\bfseries
Yuanwei Hu\textsuperscript{1},
Bo Peng\textsuperscript{2},
Yuheng Jia\textsuperscript{3},
Xinting Hu\textsuperscript{4},
Yadan Luo\textsuperscript{2},
Wenjie Zhu\textsuperscript{5}}\\[0.8em]
{\normalfont\fontsize{8.5pt}{10pt}\selectfont
\textsuperscript{1}University College London \hspace{0.5em}
\textsuperscript{2}The University of Queensland \hspace{0.5em}
\textsuperscript{3}Southeast University\\
\textsuperscript{4}University of Science and Technology of China \hspace{0.5em}
\textsuperscript{5}Auckland University of Technology}
\end{minipage}%
}

\newif\ificlrpreprint
\iclrpreprinttrue
\ificlrpreprint\iclrfinalcopy\fi

\begin{document}

\maketitle
\ificlrpreprint\lhead{Preprint}\fi

\begin{abstract}

Vision–language pre-training has reshaped image clustering, giving rise to language-assisted image clustering (LaIC), which leverages textual semantics to complement visual representations.
Despite the rapid proliferation of LaIC methods, it remains unclear how much LaIC has actually advanced image clustering, as existing studies generally suffer from major limitations, including inconsistent experimental settings, inadequate dataset selection, and limited evaluation dimensions.
To address this gap, we introduce \texttt{VLM4Cluster}, a comprehensive benchmark for image clustering in the era of pre-trained vision–language models (VLMs). 
\texttt{VLM4Cluster} implements \textbf{17} representative methods spanning classical, deep, and language-assisted image clustering, and evaluates them on \textbf{20} datasets covering classical, challenging, fine-grained, large-scale, and out-of-distribution settings. 
Beyond effectiveness, \texttt{VLM4Cluster} systematically investigates image clustering along three complementary dimensions: robustness to adversarial perturbations, generalization under distribution shifts, and computational efficiency.
Our study shows that LaIC substantially advances the clustering performance frontier on many semantically demanding benchmarks, generally exhibits stronger generalization under distribution shifts, and achieves a more favorable effectiveness–efficiency trade-off. However, its gains become less consistent on large-scale and fine-grained datasets, while language assistance does not systematically reduce sensitivity to adversarial perturbations.
\texttt{VLM4Cluster} is released at \href{https://github.com/YuanweiHuu/VLM4Cluster}{\textcolor{red}{here}}.
\end{abstract}

\section{Introduction}
As a fundamental problem in machine learning, image clustering~\citep{lu2024survey} aims to partition a set of unlabeled images into groups such that images within the same cluster are semantically similar.
Classical image clustering methods rely on strong assumptions about the data distribution, e.g., compactness~\citep{compactness}, connectivity~\citep{connectivity}, sparsity~\citep{sparsity}, and low-rankness~\citep{lowrank}. This renders them ill-suited for complex, high-dimensional data.
As an improvement, deep image clustering non-linearly transforms the original samples into a latent embedded space for successfully providing more effective features and promising performance. The pioneers in deep image clustering focus on jointly learning feature representations and cluster assignments with priors induced by image reconstruction~\citep{8,32,33}, probabilistic modeling~\citep{9,12,13}, mutual information maximization~\citep{34,35}, data augmentation~\citep{36,37,21,6,47,22,58}, and neighborhood consistency~\citep{19,yu2023contextually}. 

Advances in representation learning have driven major breakthroughs in deep image clustering. Observing that pretext tasks from representation learning can be used to obtain semantically meaningful features, SCAN~\citep{van2020scan} leverages the pretext features as a prior for clustering the images. To avoid learning feature representations from scratch for each image dataset of interest, advanced methods such as TEMI~\citep{adaloglou2023exploring}, CPP~\citep{chu2024image}, and PRO-DSC~\citep{meng2025exploring} build their clustering pipelines on top of Vision Transformers (ViTs)~\citep{dosovitskiy2020image} pre-trained on large-scale image datasets. Despite these advances, prior works tend to depend exclusively on visual self-supervision signals, resulting in inherent limitations, particularly in cases where images are visually similar to but semantically different from each other.

In recent years, the emergence of vision–language pre-training, exemplified by CLIP~\citep{radford2021learning} and SigLIP~\citep{zhai2023sigmoid}, has fundamentally reshaped the landscape of image clustering, shifting it beyond a vision-only paradigm toward a \textit{language-assisted} regime. 
By mapping visual and textual inputs into a shared embedding space, Vision–Language Models (VLMs) capture cross-modal correspondences and thereby enrich the semantic structure of image representations.
Building on this capacity, the recently emerging Language-assisted Image Clustering (LaIC) frameworks leverage textual semantics as the supervision signal to guide clustering in the visual domain. SIC~\citep{cai2023semantic} uses textual semantics to enhance image pseudo-labeling, while TAC~\citep{li2024image} focuses on leveraging textual semantics to enhance feature discriminability. To reduce noisy semantics filtered from WordNet~\citep{WordNet}, GradNorm~\citep{peng2025provable} estimates noun positiveness via gradients. NTK-SC~\citep{peng2025provable} constructs vision-language spectral affinities by coupling visual proximity with semantic overlap, and SEIC~\citep{li2026self} mines consistency between image-text pairs at the instance, cluster assignment, and cluster center levels. Differently, SAC~\citep{zhang2026semantic} leverages Multi-modal Large Language Models (MLLMs) to generate textual descriptions for each image, and subsequently integrates both visual and textual modalities to improve image clustering performance through multi-modal collaboration. MAGIC~\citep{zhangmagic} designs information bottleneck-inspired semantic adapters that adaptively refine and compress the semantically dense features into clustering-friendly representations under task guidance.

However, despite the growing popularity of image clustering, the lack of a standardized benchmark hinders systematic evaluation and limits a clear understanding of progress in this field. Specifically:
\begin{itemize}
    \item Existing works on LaIC directly base their comparisons on reported baseline results without controlled reproduction, which undermines the validity of their claimed improvements given inconsistencies in dataset splits, backbone choices, and image resolutions.
    \item Datasets widely used in the literature are either too outdated to reflect real-world scenarios or insufficiently challenging given the powerful capabilities of pre-trained VLMs. Therefore, evaluations on such datasets may overestimate performance on realistic, yet harder tasks.
    \item Current literature overwhelmingly prioritizes effectiveness when evaluating clustering performance, leaving other aspects relevant to real-world deployment underexplored.
\end{itemize}

To address these limitations, this paper releases \texttt{VLM4Cluster}, a comprehensive benchmark that is specially designed for systematically and reproducibly evaluating image clustering methods in the era of pre-trained VLMs. In particular, our work makes the following key contributions:
\begin{itemize}
    \item \texttt{VLM4Cluster} integrates \textbf{17} representative methods from all major paradigms (4 from classical image clustering, 5 from deep image clustering, 8 from LaIC) within a unified codebase to enable fair and reproducible comparison. 
    \item \texttt{VLM4Cluster} evaluates all reproduced methods across \textbf{20} widely used datasets with diverse characteristics, including 5 classical, 3 challenging, 5 fine-grained, 2 large-scale, and 5 out-of-distribution datasets.
    \item \texttt{VLM4Cluster} extends beyond effectiveness by incorporating three complementary dimensions: robustness to adversarial perturbations, generalization under distribution shifts, and computational efficiency.
    \item \texttt{VLM4Cluster} is released as an extensible open-source platform, together with a leaderboard for convenient comparison across methods, datasets, and VLMs, to promote fair and reproducible benchmarking and accelerate progress in the field.
\end{itemize}

\begin{table}[tb]
\centering
\caption{\textbf{Overview of the \texttt{VLM4Cluster} benchmark.}
  The benchmark covers 17 clustering algorithms, 20 datasets,
  and VLM configurations spanning vision backbones, pretraining
  data, and training objectives. Evaluation encompasses
  effectiveness, robustness, generalization, and efficiency.}
\label{overview}
\begingroup

\roboto
\fontsize{10}{12}\selectfont
\color[HTML]{202124}
\renewcommand{\arraystretch}{1.0}
\setlength{\extrarowheight}{0pt}
\setlength{\tabcolsep}{6pt}

\resizebox{0.95\textwidth}{!}{%
\begin{tabular}{l|l}

\multicolumn{2}{c}{\cellcolor[HTML]{E8F0FE}%
  \textcolor[HTML]{1967D2}{\large\textbf{Algorithms}}} \\
\textbf{Classical Image Clustering}
  & $k$-means, Spectral Clustering (SC), EnSC-ORGEN, SSC-OMP \\
\textbf{Deep Image Clustering}
  & IDC, SCAN, CPP, TEMI, PRO-DSC \\
\textbf{Language-assisted Image Clustering (LaIC)}
  & SIC, TAC, TAC++, SAC, GradNorm, NTK-SC, SEIC, MAGIC \\

\multicolumn{2}{c}{\cellcolor[HTML]{E6F4EA}%
  \textcolor[HTML]{188038}{\large\textbf{Datasets}}} \\
\textbf{Classical}
  & CIFAR-10, CIFAR-20, STL-10, ImageNet-10, ImageNet-Dogs \\
\textbf{Challenging}
  & DTD, UCF-101, CIFAR-100 \\
\textbf{Fine-grained}
  & Aircraft, Foods, Pets, Cars, Flowers \\
\textbf{Large-scale}
  & ImageNet-1K, Places-365 \\
\textbf{Out-of-distribution}
  & ImageNet-C, ImageNet-V2, ImageNet-S, ImageNet-A, ImageNet-R \\

\multicolumn{2}{c}{\cellcolor[HTML]{FEF7E0}%
  \textcolor[HTML]{B06000}{\large\textbf{VLMs}}} \\
\textbf{Model Scale}
  & ViT-B/32, ViT-B/16, ViT-L/14 \\
\textbf{Data Scale}
  & LAION-400M, LAION-2B \\
\textbf{Training Objective}
  & CLIP, SigLIP \\

\multicolumn{2}{c}{\cellcolor[HTML]{FCE8E6}%
  \textcolor[HTML]{C5221F}{\large\textbf{Evaluation}}} \\
\textbf{Effectiveness}
  & Correspondence between cluster assignments and class labels \\
\textbf{Robustness}
  & Effectiveness under adversarial perturbations \\
\textbf{Generalization}
  & Effectiveness under distribution shifts \\
\textbf{Efficiency}
  & Time consumption \\

\end{tabular}%
}
\endgroup
\end{table}

\section{Benchmark Design}
In this section, we introduce \texttt{VLM4Cluster} in terms of algorithms ($\triangleright$ Section~\ref{alg}), datasets ($\triangleright$ Section~\ref{dataset}), VLMs ($\triangleright$ Section~\ref{vlm}), evaluation strategy ($\triangleright$ Section~\ref{eva}), and the comparison with related benchmarks ($\triangleright$ Section~\ref{comp}). The overview of \texttt{VLM4Cluster} is shown in Table~\ref{overview}.
\subsection{Algorithms}
\label{alg}
To form our core benchmark, we carefully select \textbf{17} representative, state-of-the-art image clustering methods that span three main paradigms:
\begin{itemize}
    \item \textbf{Classical Image Clustering}: $k$-means~\citep{dhillon2001concept}, Spectral Clustering~\citep{ng2001spectral}, EnSC~\citep{you2016oracle}, and SSC-OMP~\citep{you2016scalable}
    \item \textbf{Deep Image Clustering}\footnote{We acknowledge that numerous deep image clustering methods have been proposed since 2020, including CC~\citep{21}, NMM~\citep{19}, LFSS~\citep{li2025learning}, MiCE~\citep{37}, TCC~\citep{58}, and ProPos~\citep{6}. However, many of these methods rely heavily on carefully designed input-space data augmentations. Consequently, they are not directly applicable in their original form to our benchmark, which operates on features extracted from pre-trained VLMs rather than raw images.}:
    IDC~\citep{liu2024interactive}, SCAN~\citep{van2020scan}, CPP~\citep{chu2024image}, TEMI~\citep{adaloglou2023exploring}, and PRO-DSC~\citep{meng2025exploring}
    \item \textbf{Language-assisted Image Clustering}: SIC~\citep{cai2023semantic}, TAC~\citep{li2024image}, TAC++~\citep{li2024image}, SAC~\citep{zhang2026semantic}, GradNorm~\citep{peng2025provable}, NTK-SC~\citep{peng2025provable}, SEIC~\citep{li2026self}, and MAGIC~\citep{zhangmagic}
\end{itemize}
Detailed descriptions of these methods are provided in \textbf{Appendix~\ref{appendix_a}}.



\subsection{Datasets}
\label{dataset} 
To evaluate algorithmic performance, we curate a comprehensive benchmark suite of \textbf{20} datasets that are widely adopted in the computer vision literature. To ensure broad and systematic coverage of different clustering scenarios, we organize these datasets into five categories:
\begin{itemize}
    \item \textbf{Classical Datasets}: CIFAR-10~\citep{alex2009learning}, CIFAR-20~\citep{alex2009learning}, STL-10~\citep{coates2011analysis}, ImageNet-10~\citep{chang2017deep}, and ImageNet-Dogs~\citep{chang2017deep}
    \item \textbf{Challenging Datasets}: DTD~\citep{cimpoi2014describing}, UCF-101~\citep{soomro2012ucf101}, and CIFAR-100~\citep{alex2009learning}
    \item \textbf{Fine-grained Datasets}: Pets~\citep{pets}, Cars~\citep{krause20133d}, Aircraft~\citep{maji2013fine}, Foods~\citep{bossard2014food}, and Flowers~\citep{nilsback2008automated}
    \item \textbf{Large-scale Datasets}: ImageNet-1K~\citep{ImageNet} and Places-365~\citep{zhou2017places}
    \item \textbf{Out-of-distribution Datasets}: ImageNet-C~\citep{hendrycks2019benchmarking}, ImageNet-V2~\citep{recht2019imagenet}, ImageNet-S~\citep{wang2019learning}, ImageNet-A~\citep{hendrycks2021natural}, and ImageNet-R~\citep{hendrycks2021many}
\end{itemize}
Detailed statistics and dataset descriptions are provided in \textbf{Appendix~\ref{appendix_b}}.

\subsection{Vision Language Models}
\label{vlm} 
To maximize compatibility and reproducibility, we build our benchmark on OpenCLIP~\citep{cherti2023reproducible}, taking advantage of its open-source pre-trained VLMs across diverse model sizes, data scales, and training objectives. Specifically, we evaluate methods on VLMs utilizing ViT-B/32, ViT-B/16, or ViT-L/14 as visual encoders (paired with correspondingly scaled text encoders), pre-trained on LAION-400M~\citep{schuhmann2021laion} or LAION-2B~\citep{schuhmann2022laion} datasets, and optimized using CLIP~\citep{radford2021learning} or SigLIP~\citep{zhai2023sigmoid}.

\subsection{Evaluation Dimensions}
\label{eva} 
To comprehensively evaluate algorithms across diverse real-world scenarios, \texttt{VLM4Cluster} covers 4 critical dimensions tailored to image clustering: effectiveness, robustness, generalization, and efficiency. Each dimension is associated with carefully designed evaluation protocols, reflecting \texttt{VLM4Cluster}'s goal of serving as a flexible, rigorous, and high-standard benchmark.

\textbf{Effectiveness} measures how an algorithm's output reflects the intrinsic structure of the data. An effective algorithm should produce cluster assignments that closely correspond to the true class labels.

\textbf{Robustness} assesses how well image clustering algorithms perform when exposed to adversarial perturbations. A robust algorithm should maintain its clustering performance even when inputs are subjected to imperceptible yet malicious perturbations.

\textbf{Generalization}  assesses how well image clustering algorithms perform on out-of-distribution data. Algorithms with strong generalization capabilities should exhibit only limited degradation in clustering performance under distribution shifts.

\textbf{Efficiency} measures the speed of image clustering algorithms. An efficient algorithm should achieve fast clustering while minimizing computing resource consumption.

\subsection{Discussion on Existing Benchmarks}
\label{comp} 
Most clustering benchmarks in the literature are designed for \textit{non-image data}. For example, \cite{javed2020benchmark} benchmark classical clustering methods on time-series datasets, while \cite{wu2026dgcbench} focus on deep clustering for graph-structured data and introduces PyDGC, a unified framework for fair and reproducible comparison. In contrast, relatively little attention has been devoted to benchmarking image clustering. To the best of our knowledge, the main exceptions are ClustPy~\citep{leiber2023benchmarking} and CLUBench~\citep{xiao2026clubench}. Despite their valuable contributions, several limitations remain: 1) existing evaluations are often restricted to relatively simple datasets and a limited set of evaluation metrics; 2) comprehensive comparisons spanning classical, deep, and language-assisted image clustering methods are still lacking; and 3) there is no easy-to-use toolbox for reproducible benchmarking in the emerging era of vision--language pre-training.

\section{Experiments}

\subsection{Research Questions}
To translate the evaluation dimensions outlined in Section~\ref{eva} into a concrete empirical study, we formulate four corresponding research questions (RQs). 

\underline{\textbf{RQ1: How much progress are made by LaIC?}}

\textbf{Motivation:} Although prior studies on LaIC have reported promising results, they are often conducted under experimental setups that differ substantially from those used by traditional vision-only paradigms, making direct and fair comparisons difficult. Furthermore, we notice that most existing methods are evaluated on relatively simple datasets, some of which are already nearly saturated, rendering marginal performance gains less indicative of meaningful methodological progress.

\underline{\textbf{RQ2: Are LaIC methods more robust to adversarial perturbations?}}

\textbf{Motivation:} 
Recent studies~\citep{lu2023set,xu2024highly,zhou2023advclip} have shown that VLMs are vulnerable to adversarial perturbations, where even visually imperceptible changes can substantially distort the learned visual representations on which image clustering relies. Despite this vulnerability, existing image clustering methods are evaluated only on clean images. It therefore remains unclear whether LaIC methods, by leveraging multi-modal representations from pre-trained VLMs, can be more robust to adversarial perturbations than their vision-only counterparts.

\underline{\textbf{RQ3: Do LaIC methods generalize better under distribution shifts?}}

\textbf{Motivation:} 
Existing image clustering paradigms are built on the closed-world assumption that the training and test data follow the same statistical pattern, which is mathematically referred to as Independent and Identically Distributed (\textit{i.i.d.}). However, in practice, this \textit{i.i.d.} assumption often fails to hold due to unforeseen distributional shifts between training and test data. It therefore remains unclear whether LaIC methods, owing to their multi-modal nature, can generalize better under distribution shifts than their vision-only counterparts.

\underline{\textbf{RQ4: Are existing LaIC methods computationally efficient?}}

\textbf{Motivation:} The efficiency of image clustering methods in terms of computation is critical to their real-world deployment. Although incorporating language assistance may benefit image clustering, the additional computational overhead it incurs has received limited attention in existing research. This omission hinders a comprehensive assessment of the practical applicability of LaIC.

\begin{table*}[htbp]
\centering
\caption{Clustering results on classical datasets using CLIP ViT-B/32 pretrained on LAION-400M. The best value in each metric column is highlighted in \colorbox[HTML]{FFF2CC}{\textbf{bold}}.}
\label{tab:classical_b32}
\resizebox{\textwidth}{!}{%
\begin{tabular}{l|ccc|ccc|ccc|ccc|ccc}
\toprule
Dataset
& \multicolumn{3}{c|}{STL-10}
& \multicolumn{3}{c|}{CIFAR-10}
& \multicolumn{3}{c|}{CIFAR-20}
& \multicolumn{3}{c|}{ImageNet-10}
& \multicolumn{3}{c}{ImageNet-Dogs}
\\
\midrule
Metric
& NMI & ACC & ARI
& NMI & ACC & ARI
& NMI & ACC & ARI
& NMI & ACC & ARI
& NMI & ACC & ARI
\\
\midrule


\multicolumn{16}{c}{\cellcolor{gray!40}\textbf{Classical Image Clustering}}
\\

$k$-means
& 93.0 & 96.1 & 93.4
& 74.8 & 73.5 & 63.6
& 59.2 & 48.6 & 35.6
& 94.2 & 96.0 & 93.3
& 64.0 & 60.3 & 48.1
\\

SC
& 90.2 & 95.3 & 90.1
& 73.9 & 78.2 & 67.5
& 53.4 & 52.8 & 39.8
& 94.1 & 96.4 & 92.4
& 44.1 & 43.9 & 27.6
\\

SSC-OMP
& 78.6 & 83.4 & 72.9
& 71.7 & 74.0 & 62.5
& 55.5 & 47.6 & 35.2
& 89.6 & 94.0 & 87.4
& 30.3 & 34.3 & 15.6
\\

EnSC-ORGEN
& 78.8 & 83.8 & 73.0
& 75.8 & 82.2 & 70.8
& 58.9 & 52.1 & 40.3
& 93.3 & 96.4 & 92.3
& 34.9 & 42.4 & 19.7
\\

\midrule

\multicolumn{16}{c}{\cellcolor{gray!40}\textbf{Deep Image Clustering}}
\\

IDC
& \best{94.8} & 97.4 & \best{95.3}
& 84.4 & 91.9 & 82.7
& 55.9 & 53.4 & 37.0
& 97.6 & 98.8 & 97.3
& 64.2 & 64.5 & 47.3
\\

SCAN
& 91.4 & 95.4 & 90.6
& 79.8 & 85.0 & 74.4
& 61.8 & 57.8 & 44.8
& 97.0 & 98.4 & 96.5
& 63.6 & 61.7 & 46.2
\\

CPP
& 91.7 & 96.2 & 91.7
& 84.7 & 92.3 & 83.8
& 63.9 & 60.5 & 43.9
& 97.1 & 98.2 & 96.1
& 63.7 & 67.2 & 42.7
\\

TEMI
& 94.1 & 97.5 & 94.5
& 86.9 & 93.3 & 86.1
& 62.2 & 58.7 & 44.4
& 97.8 & 98.8 & 97.3
& 67.2 & 66.1 & 52.8
\\

PRO-DSC
& 93.9 & 97.3 & 94.3
& 85.9 & 93.0 & 85.3
& 63.7 & 58.1 & 45.4
& 95.2 & 97.4 & 94.3
& 66.3 & 64.8 & 49.1
\\

\midrule

\multicolumn{16}{c}{\cellcolor{gray!40}\textbf{Language-assisted Image Clustering}}
\\

SIC
& 94.3 & 97.6 & 94.8
& 87.8 & 94.1 & 87.6
& 59.9 & 53.2 & 40.3
& \best{98.4} & \best{99.2} & \best{98.2}
& 65.2 & 63.2 & 51.7
\\

TAC
& 92.6 & 96.4 & 92.4
& 86.3 & 93.3 & 86.0
& 65.3 & 56.4 & 42.5
& 98.0 & 98.8 & 97.4
& 75.8 & 72.9 & 60.6
\\

TAC++
& 94.0 & 97.4 & 94.3
& \best{87.9} & \best{94.3} & \best{88.0}
& 62.6 & 60.2 & 46.4
& 97.8 & 98.8 & 97.3
& 78.0 & 79.9 & 69.0
\\

SAC
& 93.4 & 97.0 & 93.7
& 85.7 & 93.1 & 85.6
& 63.0 & 62.1 & 47.0
& 96.7 & 98.2 & 96.1
& 72.7 & 73.5 & 60.7
\\

GradNorm
& 93.6 & 97.3 & 94.1
& 85.9 & 93.0 & 85.3
& 65.4 & 60.4 & 45.2
& 95.6 & 97.4 & 94.4
& 73.8 & 74.9 & 61.5
\\

NTK-SC
& 94.5 & \best{97.7} & 95.0
& 87.8 & 94.2 & 87.8
& 64.0 & 52.7 & 36.1
& 97.9 & 98.8 & 97.4
& \best{80.1} & 80.9 & 70.7
\\

SEIC
& 94.4 & \best{97.7} & 95.0
& 86.7 & 93.8 & 86.9
& 60.1 & 60.3 & 45.7
& 98.3 & \best{99.2} & \best{98.2}
& 68.3 & 70.3 & 56.8
\\

MAGIC
& 94.3 & 97.5 & 94.6
& 87.7 & 94.1 & 87.6
& \best{67.4} & \best{66.7} & \best{53.2}
& \best{98.4} & \best{99.2} & \best{98.2}
& 79.9 & \best{82.4} & \best{71.5}
\\

\bottomrule
\end{tabular}%
}
\end{table*}

\begin{table*}[t]
\centering
\begin{minipage}[t]{0.55\textwidth}

\begin{minipage}[t][4.5em][t]{\linewidth}
\centering
\captionof{table}{
Clustering results on challenging datasets using CLIP ViT-B/32 pretrained on LAION-400M. The best result is highlighted in \colorbox[HTML]{FFF2CC}{\textbf{bold}}.}
\label{tab:challenging_b32}
\end{minipage}

\scriptsize
\setlength{\tabcolsep}{2pt}
\renewcommand{\arraystretch}{1.0}

\makebox[\linewidth][c]{%
\begin{tabular}{l|ccc|ccc|ccc}
\toprule
Dataset
 & \multicolumn{3}{c|}{DTD}
 & \multicolumn{3}{c|}{UCF-101}
 & \multicolumn{3}{c}{CIFAR-100} \\
\midrule

Metric
 & NMI & ACC & ARI
 & NMI & ACC & ARI
 & NMI & ACC & ARI \\
\midrule

\multicolumn{10}{c}{\cellcolor{gray!40}\textbf{Classical Image Clustering}} \\

$k$-means
& 66.1 & 50.9 & 37.4
& 81.0 & 58.7 & 50.0
& 69.1 & 52.7 & 39.1 \\

SC
& 64.6 & 54.5 & 40.2
& 81.4 & 63.0 & 55.0
& 67.2 & 52.0 & 41.9 \\

SSC-OMP
& 57.3 & 45.6 & 29.7
& 66.0 & 39.2 & 29.7
& 59.8 & 44.1 & 29.9 \\

EnSC-ORGEN
& 66.1 & 54.0 & 39.0
& 74.2 & 48.2 & 36.7
& 67.3 & 52.4 & 37.8 \\

\midrule
\multicolumn{10}{c}{\cellcolor{gray!40}\textbf{Deep Image Clustering}} \\

IDC
& 67.2 & 59.5 & 40.8
& 78.6 & 63.3 & 49.7
& 67.1 & 55.5 & 36.8 \\

SCAN
& 67.4 & 55.7 & 41.2
& 73.6 & 39.4 & 34.4
& 65.9 & 43.8 & 33.9 \\

CPP
& 67.2 & 57.7 & 41.3
& 79.4 & 59.7 & 48.1
& 73.6 & 61.0 & 39.0 \\

TEMI
& 67.8 & 57.1 & 42.2
& 80.1 & 64.0 & 54.6
& 75.2 & 66.8 & \best{53.1} \\

PRO-DSC
& 66.9 & 56.2 & 40.3
& 81.8 & 65.5 & 56.0
& 75.1 & 65.2 & 50.4 \\

\midrule
\multicolumn{10}{c}{\cellcolor{gray!40}\textbf{Language-assisted Image Clustering}} \\

SIC
& 67.7 & 56.4 & 41.5
& 81.0 & 65.3 & 54.6
& 72.2 & 58.1 & 45.3 \\

TAC
& 66.4 & 54.5 & 39.0
& 80.7 & 61.0 & 52.6
& 73.2 & 59.7 & 46.9 \\

TAC++
& 67.9 & 56.9 & 42.5
& 80.7 & 67.9 & 57.3
& 73.5 & 65.6 & 51.3 \\

SAC
& 66.7 & 56.9 & 42.5
& 79.8 & 66.3 & 56.7
& 73.7 & 66.1 & 51.6 \\

GradNorm
& 67.9 & 56.5 & 41.3
& 82.1 & 64.3 & 54.9
& 73.1 & 61.0 & 42.8 \\

NTK-SC
& \best{69.1} & \best{60.0} & \best{44.5}
& \best{84.5} & \best{72.2} & \best{63.5}
& \best{76.1} & \best{68.0} & 52.5 \\

SEIC
& 65.7 & 55.7 & 39.3
& 79.4 & 64.9 & 54.9
& 72.7 & 65.3 & 50.1 \\

MAGIC
& 62.0 & 44.8 & 32.6
& 73.3 & 51.4 & 41.2
& 72.2 & 58.4 & 45.6 \\

\bottomrule
\end{tabular}%
}

\end{minipage}%
\hspace{0.02\textwidth}%
\begin{minipage}[t]{0.42\textwidth}

\begin{minipage}[t][4.5em][t]{\linewidth}
\centering
\captionof{table}{
Clustering results on large-scale datasets with CLIP ViT-B/32 pretrained on LAION-400M. The best result is in \colorbox[HTML]{FFF2CC}{\textbf{bold}}.}
\label{tab:largescale_b32}
\end{minipage}

\scriptsize
\setlength{\tabcolsep}{2pt}
\renewcommand{\arraystretch}{1.0}

\makebox[\linewidth][c]{%
\begin{tabular}{l|ccc|ccc}
\toprule
Dataset
 & \multicolumn{3}{c|}{ImageNet-1K}
 & \multicolumn{3}{c}{Places-365} \\
\midrule

Metric
 & NMI & ACC & ARI
 & NMI & ACC & ARI \\
\midrule

\multicolumn{7}{c}{\cellcolor{gray!40}\textbf{Classical Image Clustering}} \\

$k$-means
& 73.5 & 40.6 & 28.8
& 58.4 & 30.8 & 18.2 \\

SC
& 72.3 & 38.4 & 27.7
& 56.8 & 29.5 & 18.1 \\

SSC-OMP
& 64.2 & 29.1 & 11.9
& 51.1 & 23.4 & 11.0 \\

EnSC-ORGEN
& 73.1 & 43.6 & 24.3
& 59.7 & 31.2 & 19.1 \\

\midrule
\multicolumn{7}{c}{\cellcolor{gray!40}\textbf{Deep Image Clustering}} \\

IDC
& 73.1 & 42.1 & 27.1
& 58.1 & 31.2 & 17.2 \\

SCAN
& 75.3 & 45.7 & 33.2
& 59.3 & 32.5 & 19.2 \\

CPP
& 70.5 & 39.5 & 17.4
& 54.1 & 27.5 & 12.4 \\

TEMI
& \textbf{78.1} & 53.2 & 38.9
& 60.5 & \best{34.5} & \best{20.6} \\

PRO-DSC
& 71.5 & 41.4 & 23.4
& 54.1 & 25.4 & 12.7 \\

\midrule
\multicolumn{7}{c}{\cellcolor{gray!40}\textbf{Language-assisted Image Clustering}} \\

SIC
& 77.3 & 49.6 & 36.7
& 60.0 & 30.8 & 20.0 \\

TAC
& 77.4 & 48.8 & 36.0
& 59.4 & 32.7 & 19.3 \\

TAC++
& 77.8 & \best{54.3} & 39.4
& 59.0 & 33.6 & 19.4 \\

SAC
& 77.9 & 53.6 & \best{39.6}
& 58.9 & 34.1 & \best{20.6} \\

GradNorm
& 77.5 & 49.3 & 36.1
& 59.3 & 32.2 & 19.1 \\

NTK-SC
& 77.4 & 51.9 & 38.1
& \best{60.6} & 33.7 & 19.5 \\

SEIC
& 75.3 & 45.0 & 29.8
& 59.2 & 31.2 & 17.6 \\

MAGIC
& 74.4 & 40.5 & 28.9
& 60.0 & 33.5 & 20.1 \\

\bottomrule
\end{tabular}%
}
\end{minipage}
\end{table*}

\begin{table*}[tb]
\centering
\caption{Clustering results on fine-grained datasets using CLIP ViT-B/32 pretrained on LAION-400M. The best value in each metric column is highlighted in \colorbox[HTML]{FFF2CC}{\textbf{bold}}.}
\label{tab:finegrained_b32}
\resizebox{\textwidth}{!}{%
\begin{tabular}{l|ccc|ccc|ccc|ccc|ccc}
\toprule
Dataset
& \multicolumn{3}{c|}{Aircraft}
& \multicolumn{3}{c|}{Cars}
& \multicolumn{3}{c|}{Flowers}
& \multicolumn{3}{c|}{Food}
& \multicolumn{3}{c}{Pets}
\\
\midrule
Metric
& NMI & ACC & ARI
& NMI & ACC & ARI
& NMI & ACC & ARI
& NMI & ACC & ARI
& NMI & ACC & ARI
\\
\midrule

\multicolumn{16}{c}{\cellcolor{gray!40}\textbf{Classical Image Clustering}}
\\

$k$-means
& 48.2 & 21.6 & 11.7
& 80.4 & 53.4 & 46.7
& 88.3 & 72.4 & 68.6
& 74.7 & 57.6 & 54.1
& 79.9 & 68.2 & 60.9
\\

SC
& 45.2 & 21.6 & 11.3
& 75.0 & 45.4 & 39.2
& 86.4 & 72.9 & 70.6
& 66.1 & 55.3 & 44.4
& 71.5 & 67.1 & 60.9
\\

SSC-OMP
& 41.8 & 17.9 & 6.6
& 62.4 & 37.4 & 21.8
& 51.5 & 31.4 & 22.0
& 47.9 & 36.2 & 23.0
& 42.9 & 30.5 & 17.6
\\

EnSC-ORGEN
& 45.8 & 22.3 & 9.8
& 77.3 & 55.8 & 41.3
& 86.0 & 71.7 & 65.4
& 71.7 & 63.9 & 48.2
& 61.2 & 54.0 & 37.3
\\

\midrule

\multicolumn{16}{c}{\cellcolor{gray!40}\textbf{Deep Image Clustering}}
\\

IDC
& 47.0 & 23.0 & 12.1
& 78.5 & 55.7 & 43.3
& 63.2 & 37.8 & 27.3
& 69.0 & 60.9 & 42.1
& 79.1 & 70.3 & 58.6
\\

SCAN
& 40.5 & 13.7 & 7.2
& 72.2 & 26.2 & 26.2
& 75.8 & 49.8 & 44.7
& 66.7 & 43.7 & 37.1
& 77.7 & 69.1 & 58.8
\\

CPP
& 47.1 & 22.5 & 11.3
& 75.2 & 52.5 & 38.9
& 84.7 & 69.9 & 60.6
& 67.0 & 59.5 & 41.8
& 74.9 & 69.3 & 52.9
\\

TEMI
& 48.2 & 23.8 & 12.6
& \best{83.2} & \best{61.6} & \best{52.4}
& 70.5 & 49.6 & 41.9
& \best{78.7} & \best{75.2} & \best{62.8}
& 78.2 & 70.3 & 58.8
\\

PRO-DSC
& 45.8 & 22.1 & 10.6
& 80.9 & 57.9 & 46.2
& 83.8 & 68.7 & 62.0
& 68.4 & 58.0 & 33.2
& 82.6 & 76.9 & 67.7
\\

\midrule

\multicolumn{16}{c}{\cellcolor{gray!40}\textbf{Language-assisted Image Clustering}}
\\

SIC
& 47.4 & 21.8 & 11.4
& 79.0 & 52.2 & 44.5
& 81.9 & 68.4 & 61.0
& 74.0 & 66.8 & 53.3
& 77.4 & 68.4 & 57.9
\\

TAC
& 48.7 & 23.3 & 12.4
& 80.3 & 54.2 & 46.3
& \best{89.5} & 74.5 & 71.1
& 69.3 & 60.5 & 43.9
& 81.0 & 72.9 & 63.0
\\

TAC++
& 47.5 & 23.5 & 11.7
& 77.3 & 51.0 & 41.2
& 84.7 & 76.6 & 68.4
& 71.1 & 66.2 & 50.5
& 83.3 & 78.1 & \best{68.7}
\\

SAC
& 47.6 & 23.3 & 12.4
& 79.8 & 56.5 & 47.5
& 80.3 & 61.6 & 56.0
& 71.2 & 65.8 & 51.1
& 80.1 & 76.9 & 65.7
\\

GradNorm
& 48.0 & 22.7 & 11.6
& 79.5 & 53.8 & 45.4
& 88.9 & 75.6 & \best{71.7}
& 73.6 & 65.7 & 50.7
& 82.9 & 75.2 & 66.2
\\

NTK-SC
& \best{50.4} & \best{26.3} & \best{14.3}
& 82.2 & 61.1 & 51.4
& 88.2 & \best{80.9} & 71.2
& 76.2 & 69.9 & 54.2
& \best{83.7} & \best{79.2} & 68.6
\\

SEIC
& 44.6 & 19.7 & 9.5
& 71.3 & 44.8 & 31.6
& 74.4 & 56.7 & 49.4
& 70.3 & 64.0 & 48.7
& 72.1 & 51.2 & 39.2
\\

MAGIC
& 43.2 & 16.8 & 7.9
& 61.6 & 25.0 & 17.8
& 60.9 & 34.1 & 28.1
& 71.2 & 60.5 & 48.0
& 77.5 & 66.4 & 57.0
\\

\bottomrule
\end{tabular}%
}
\end{table*}

\subsection{Effectiveness Analysis (RQ1)}
\label{effectiveness}
In this subsection, we re-implement and benchmark all representative methods described in Section~\ref{alg} under a strictly unified experimental setup, as detailed in \textbf{Appendix~\ref{appendix_d}}. Beyond the classical datasets commonly used in prior work, we further evaluate LaIC and vision-only methods on the challenging, large-scale, and fine-grained datasets introduced in Section~\ref{dataset}, enabling a rigorous comparison across increasingly challenging scenarios and an assessment of the progress brought by language assistance. Following standard practice in image clustering, we evaluate performance using Accuracy (ACC), Normalized Mutual Information (NMI), and Adjusted Rand Index (ARI), with detailed definitions provided in \textbf{Appendix~\ref{appendix_c}}. The results on classical, challenging, large-scale, and fine-grained datasets are reported in Tables~\ref{tab:classical_b32}, \ref{tab:challenging_b32}, \ref{tab:largescale_b32}, and \ref{tab:finegrained_b32}, respectively.

Comparing the best LaIC result with the best vision-only result for each dataset and metric, LaIC advances the performance frontier in 33 out of 45 dataset--metric comparisons. The gains are not confined to a particular evaluation metric: LaIC outperforms the strongest vision-only baselines on 11 out of 15 datasets in NMI, 12 out of 15 in ACC, and 10 out of 15 in ARI. Moreover, on 9 out of 15 datasets, LaIC establishes a new best result across all three metrics. These results demonstrate that language assistance has made substantial progress beyond traditional vision-only clustering, although the magnitude of the improvement varies considerably across dataset regimes.

The largest gains occur on several classical and challenging benchmarks where semantic cues can provide complementary information beyond visual similarity. On ImageNet-Dogs, for example, the best vision-only results are 67.2\% NMI, 67.2\% ACC, and 52.8\% ARI, whereas LaIC improves them to 80.1\%, 82.4\%, and 71.5\%, corresponding to gains of 12.9, 15.2, and 18.7 percentage points, respectively. Similarly, LaIC improves CIFAR-20 by 3.5/6.2/7.8 points in NMI/ACC/ARI and UCF-101 by 2.7/6.7/7.5 points. These results show that language assistance can substantially raise the clustering performance ceiling on selected benchmarks.

However, this advantage becomes much less pronounced at scale. On ImageNet-1K, LaIC improves the best vision-only ACC and ARI by only 1.1 and 0.7 points, respectively, while its best NMI is 0.2 points lower. On Places-365, the two paradigms perform almost identically: LaIC changes the best NMI and ACC by only +0.1 and -0.4 points, respectively, while attaining the same best ARI of 20.6\%. Thus, the substantial gains observed on several smaller benchmarks do not directly translate to large-scale clustering, where the large number of categories, high intra-class diversity, and complex visual semantics remain challenging.

A similar lack of consistency appears on fine-grained datasets. LaIC improves all three metrics on Aircraft and Pets, and on Flowers it raises the best ACC from 72.9\% to 80.9\%, an improvement of 8.0 points. In contrast, on Cars, the strongest LaIC results remain below the vision-only frontier on all three metrics. The degradation is more pronounced on Food, where LaIC trails the best vision-only results by 2.5, 5.3, and 8.6 points in NMI, ACC, and ARI, respectively. This mixed performance suggests a potential mismatch between high-level language semantics and the subtle visual attributes needed to distinguish certain fine-grained categories.

To further assess the generality of these observations, we evaluate additional VLMs with different backbone architectures, pre-training data scales, and training objectives, with detailed results provided in \textbf{Appendix~\ref{appendix_e}}. Across these settings, we observe similar trends, indicating that the effectiveness patterns above are broadly consistent across different VLM configurations.

\begin{tcolorbox}[center, width=140mm, colback=blue!5!white,colframe=blue!75!black,colbacktitle=red!80!black]
\textbf{Takeaway 1:} LaIC clearly advances the performance frontier beyond vision-only models on semantically demanding datasets. However, scalability and fine-grained discrimination are two major limitations of current LaIC methods, indicating significant room for improvement.
\end{tcolorbox}

\begin{figure*}[tb]
    \centering
    \includegraphics[width=\linewidth]{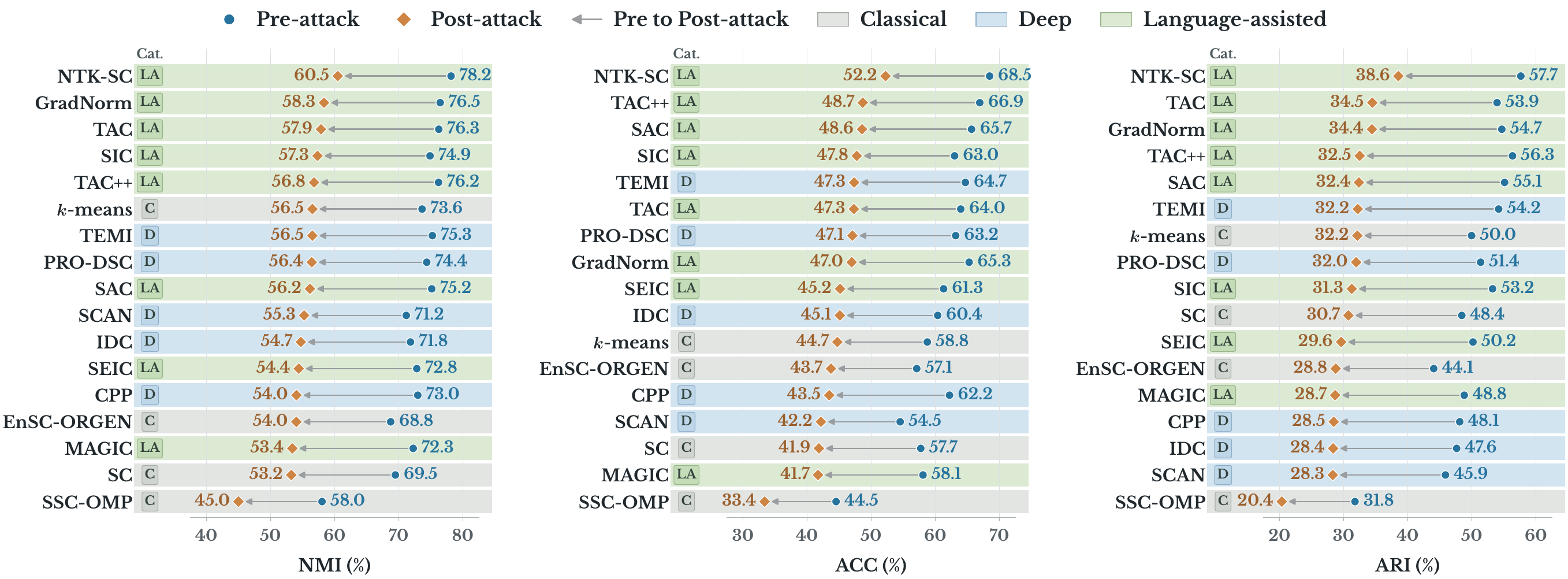}
    \caption{Average pre-attack and post-attack clustering performance across 15 datasets using CLIP ViT-B/32 pretrained on LAION-400M, evaluated w.r.t. NMI (left), ACC (middle), and ARI (right).}
    \label{fig:robustness_1}
\end{figure*}

\begin{figure*}[tb]
    \centering
    \IfFileExists{robustness_bar5.pdf}{%
        \includegraphics[width=\linewidth,height=2in,keepaspectratio]{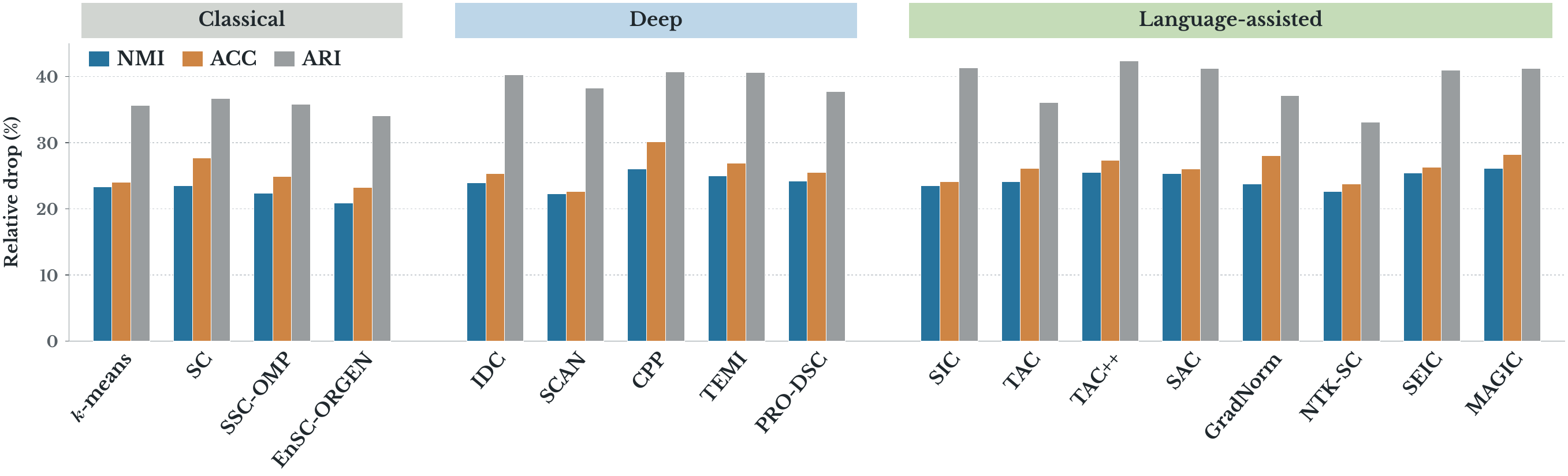}%
    }{%
        \fbox{\parbox[c][2in][c]{0.92\linewidth}{\centering Missing figure: \texttt{robustness\_bar5.pdf}}}%
    }
    \caption{Comparison of the relative drop in clustering performance from clean to adversarial conditions, evaluated on CLIP ViT-B/32 pretrained on LAION-400M using NMI, ACC, and ARI.}
    \label{fig:robustness_2}
\end{figure*}

\subsection{Robustness Analysis (RQ2)}
\label{robustness}
In this subsection, we evaluate all methods outlined in Section~\ref{alg} under controlled adversarial conditions. Specifically, for each image dataset, we fit each method on the clean training samples following the same experimental setup as in \textbf{RQ1}. We then employ AnyAttack~\citep{zhang2025anyattack} to generate adversarial examples from the corresponding test samples with a perturbation radius of $8/255$, and evaluate the fitted methods on these perturbed test images. We assess adversarial robustness from two complementary perspectives across the 15 datasets used in Section~\ref{effectiveness}: absolute post-attack clustering performance and the relative performance drop from clean to adversarial conditions. The corresponding results are visualized in Figure~\ref{fig:robustness_1} and Figure~\ref{fig:robustness_2}, respectively, while detailed results for individual datasets are provided in \textbf{Appendix~\ref{appendix_f}}.

A clear advantage of LaIC methods emerges in terms of post-attack performance relative to vision-only methods. Among classical and deep image clustering approaches, the best post-attack NMI, ACC, and ARI scores are \(56.5\%\), \(47.3\%\), and \(32.2\%\), respectively. By comparison, 6 of the 8 LaIC methods match or exceed the best vision-only result on at least two of the three metrics. Notably, TAC, TAC++, and NTK-SC match or surpass the best vision-only results across all three metrics, while SIC, SAC, and GradNorm remain competitive on two of the three. 

The trend is less consistent when robustness is measured by relative performance drop. LaIC methods do not systematically exhibit smaller relative declines than classical or deep clustering approaches; instead, their degradation is broadly comparable, with substantial variation across methods and metrics. In particular, several LaIC methods suffer ARI drops exceeding 40\%, while their NMI and ACC declines generally remain within the range observed for vision-only baselines. 



\begin{tcolorbox}[center, width=140mm, colback=blue!5!white,colframe=blue!75!black,colbacktitle=red!80!black]
\textbf{Takeaway 2:} Language assistance generally raises the post-attack performance floor of image clustering, but does not by itself guarantee reduced sensitivity to adversarial attacks.
\end{tcolorbox}

\subsection{Generalization Analysis (RQ3)}
In this subsection, we use ImageNet-1K as a case study. Since classical image clustering methods are fully transductive and cannot deal with out-of-sample data, we restrict our evaluation to deep image clustering and LaIC methods. Following the experimental setup outlined in \textbf{RQ1}, we start with fiting each method using the ImageNet-1K training split as the in-distribution (ID) data. Subsequently, we assess their generalization capabilities across the out-of-distribution (OOD) datasets detailed in Section~\ref{dataset}. In addition to reporting OOD clustering performance, we, following~\cite{mao2023coco}, adopt \emph{effective robustness}~\citep{taori2020measuring} as a complementary metric to measures performance under distribution shift
relative to that expected from the corresponding ID clustering performance. The results are summarized in Table~\ref{tab:ood_b32}, where effective robustness is computed based on the average clustering performance across the five OOD datasets.

LaIC methods generally achieve stronger averaged performance under distribution shifts. The best results on all three averaged metrics are obtained by LaIC methods: TAC achieves the highest average NMI of 69.0\%, while NTK-SC achieves the highest average ACC and ARI of 38.5\% and 25.5\%, respectively. In comparison, the strongest deep image clustering baseline, TEMI, achieves 64.4\% NMI, 33.8\% ACC, and 14.8\% ARI. More broadly, 7 of the 8 LaIC methods outperform TEMI in both NMI and ARI, while 4 of them surpass TEMI across all three metrics. This indicates that the advantage is not limited to a single LaIC method. A similar pattern emerges w.r.t. effective robustness: nearly all deep image clustering methods exhibit negative effective robustness across all three metrics. By contrast, 6 of the 8 LaIC methods achieve positive effective robustness on at least two metric, with NTK-SC, SEIC, and MAGIC showing positive robustness consistently across all three metrics. 

At the dataset level, however, the leading methods vary substantially across datasets and metrics: TAC achieves the highest NMI on ImageNet-A and ImageNet-R, NTK-SC performs particularly well on ImageNet-A, ImageNet-R, and ImageNet-S, while SAC achieves the best results across all three metrics on ImageNet-V2. On ImageNet-C, even the deep clustering baseline TEMI attains the highest ACC. These results suggest that different distribution shifts pose distinct challenges to clustering methods, and strong generalization to one type of shift does not necessarily transfer to others.
\begin{table*}[tb]
\centering
\caption{Clustering performance on OOD datasets using CLIP ViT-B/32 pretrained on LAION-400M. The best result in each metric column is highlighted in \colorbox[HTML]{FFF2CC}{\textbf{bold}}. Positive and negative effective robustness values are shown in \textcolor{red}{red} and \textcolor{green!50!black}{green}, respectively.}
\label{tab:ood_b32}
\resizebox{\textwidth}{!}{%
\begin{tabular}{l|ccc|ccc|ccc|ccc|ccc|ccc|ccc}
\toprule
Dataset
 & \multicolumn{3}{c|}{ImageNet-A}
 & \multicolumn{3}{c|}{ImageNet-C}
 & \multicolumn{3}{c|}{ImageNet-R}
 & \multicolumn{3}{c|}{ImageNet-V2}
 & \multicolumn{3}{c|}{ImageNet-S}
 & \multicolumn{3}{c|}{Average}
 & \multicolumn{3}{c}{Effective Robustness} \\
\midrule
Metric & NMI & ACC & ARI
 & NMI & ACC & ARI
 & NMI & ACC & ARI
 & NMI & ACC & ARI
 & NMI & ACC & ARI
 & NMI & ACC & ARI
 & NMI & ACC & ARI \\
\midrule

\multicolumn{22}{c}{\cellcolor{gray!40}\textbf{Deep Image Clustering}} \\

IDC
& 47.6 & 12.2 & 4.1
& 66.9 & 34.8 & 12.8
& 53.2 & 31.8 & 14.2
& 76.8 & 38.2 & 18.1
& 61.0 & 25.1 & 2.8
& 61.1 & 28.4 & 10.4
& \textcolor{red}{+1.59}
& \textcolor{red}{+0.39}
& \textcolor{green!50!black}{-0.98} \\

SCAN
& 49.0 & 12.3 & 4.1
& 69.2 & 39.0 & 11.0
& 53.1 & 32.1 & 9.6
& 78.2 & 41.0 & 23.1
& 59.3 & 24.6 & 1.9
& 61.8 & 29.8 & 9.9
& \textcolor{green!50!black}{-1.32}
& \textcolor{green!50!black}{-0.92}
& \textcolor{green!50!black}{-5.60} \\

CPP
& 36.4 & 11.4 & 2.1
& 63.8 & 31.8 & 9.0
& 34.7 & 20.9 & 5.1
& 74.6 & 35.8 & 7.3
& 55.8 & 20.7 & 3.7
& 53.1 & 24.1 & 5.4
& \textcolor{green!50!black}{-2.34}
& \textcolor{green!50!black}{-2.02}
& \textcolor{green!50!black}{-0.54} \\

TEMI
& 50.3 & 12.4 & 4.2
& 72.3 & \best{45.7} & 23.2
& 54.7 & 33.6 & 14.7
& 80.1 & 46.6 & 27.1
& 64.5 & 30.7 & 5.0
& 64.4 & 33.8 & 14.8
& \textcolor{green!50!black}{-3.34}
& \textcolor{green!50!black}{-2.86}
& \textcolor{green!50!black}{-5.16} \\

PRO-DSC
& 37.9 & 11.1 & 2.9
& 64.9 & 32.8 & 14.2
& 36.8 & 20.3 & 5.5
& 75.6 & 36.6 & 15.4
& 59.4 & 24.3 & 4.5
& 54.9 & 25.0 & 8.5
& \textcolor{green!50!black}{-2.04}
& \textcolor{green!50!black}{-2.50}
& \textcolor{green!50!black}{-0.65} \\

\midrule

\multicolumn{22}{c}{\cellcolor{gray!40}\textbf{Language-assisted Image Clustering}} \\

SIC
& 49.3 & 12.4 & 4.3
& 71.1 & 42.7 & 13.7
& 52.3 & 33.4 & 8.3
& 79.4 & 43.6 & 25.4
& 63.0 & 30.7 & 2.3
& 63.0 & 32.6 & 10.8
& \textcolor{green!50!black}{-3.36}
& \textcolor{green!50!black}{-1.19}
& \textcolor{green!50!black}{-7.41} \\

TAC
& \best{55.9} & 11.2 & 3.5
& 71.5 & 39.9 & 25.7
& \best{64.8} & 28.1 & 19.0
& 78.2 & 38.2 & 22.0
& 74.5 & 40.2 & 27.8
& \best{69.0} & 31.5 & 19.6
& \textcolor{red}{+2.43}
& \textcolor{green!50!black}{-1.60}
& \textcolor{red}{+1.94} \\

TAC++
& 50.7 & 12.3 & 4.1
& 71.7 & 45.2 & 26.8
& 63.2 & 43.0 & 30.0
& 80.2 & 47.5 & 27.7
& 70.9 & 38.4 & 20.3
& 67.3 & 37.3 & 21.8
& \textcolor{red}{+0.12}
& \textcolor{green!50!black}{-0.30}
& \textcolor{red}{+1.36} \\

SAC
& 50.5 & 12.1 & 3.6
& 71.8 & 45.0 & 27.3
& 62.1 & 42.1 & 27.3
& \best{80.3} & \best{47.6} & \best{28.1}
& 70.6 & 37.9 & 20.7
& 67.1 & 36.9 & 21.4
& \textcolor{green!50!black}{-0.32}
& \textcolor{green!50!black}{-0.06}
& \textcolor{red}{+0.81} \\

GradNorm
& 45.2 & 15.5 & 6.1
& 71.7 & 41.0 & 26.8
& 62.7 & 46.5 & 33.4
& 78.5 & 38.5 & 22.7
& 74.6 & 40.7 & 28.2
& 66.5 & 36.4 & 23.4
& \textcolor{green!50!black}{-0.18}
& \textcolor{red}{+2.93}
& \textcolor{red}{+5.70} \\

NTK-SC
& 47.7 & \best{18.9} & \best{8.4}
& \best{73.6} & 43.8 & \best{28.5}
& 62.4 & \best{48.3} & \best{34.9}
& 78.6 & 38.7 & 25.8
& \best{76.4} & \best{42.7} & \best{29.7}
& 67.7 & \best{38.5} & \best{25.5}
& \textcolor{red}{+1.19}
& \textcolor{red}{+2.88}
& \textcolor{red}{+6.12} \\

SEIC
& 50.4 & 13.1 & 4.1
& 69.9 & 38.1 & 21.2
& 63.3 & 40.6 & 29.2
& 78.0 & 40.8 & 21.3
& 69.0 & 31.6 & 16.7
& 66.1 & 32.8 & 18.5
& \textcolor{red}{+3.04}
& \textcolor{red}{+2.65}
& \textcolor{red}{+5.36} \\

MAGIC
& 51.1 & 11.2 & 3.4
& 69.8 & 35.5 & 22.2
& 61.3 & 33.2 & 23.6
& 78.0 & 38.4 & 21.3
& 68.8 & 31.6 & 17.1
& 65.8 & 30.0 & 17.5
& \textcolor{red}{+4.19}
& \textcolor{red}{+3.12}
& \textcolor{red}{+4.98} \\

\bottomrule
\end{tabular}
}
\end{table*}
\begin{tcolorbox}[center, width=140mm, colback=blue!5!white,colframe=blue!75!black,colbacktitle=red!80!black]
\textbf{Takeaway 3:} LaIC generally exhibit stronger generalization under distribution shifts, as evidenced by higher averaged OOD clustering performance and stronger effective robustness. However, no single LaIC method uniformly generalizes well across diverse distribution shifts.
\end{tcolorbox}

\subsection{Efficiency Analysis (RQ4)}
\label{efficiency}
\begin{wrapfigure}{r}{0.55\textwidth}
    \centering
    \includegraphics[width=\linewidth]{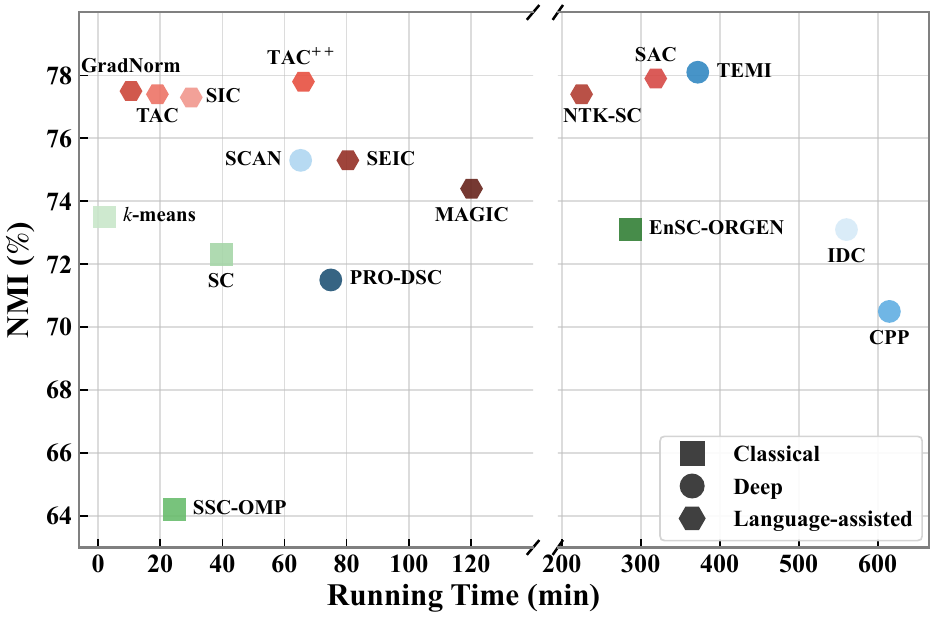}
    \caption{Time consumption and NMI of different image clustering methods on ImageNet-1K.}
    \label{fig:efficiency}
\end{wrapfigure}

In this subsection, we evaluate the computational efficiency of the image clustering methods described in Section~\ref{alg} on ImageNet-1K by comparing the total wall-clock time required by their entire clustering pipelines. The results are shown in Figure~\ref{fig:efficiency}.

As we can observe, LaIC methods generally occupy the upper-left region, indicating high clustering effectiveness with relatively low computational cost. In particular, GradNorm, TAC, and SIC achieve $\geq77.3\%$ NMI within only 10-30 minutes, substantially outperforming classical image clustering methods with comparable running times. For example, although $k$-means is faster, it achieves only 73.5\% NMI, while SC and SSC-OMP obtain 72.3\% and 64.2\% NMI, respectively. TAC++ further achieves 77.8\% NMI in about 66.1 minutes, outperforming deep image clustering methods such as SCAN and PRO-DSC under a similar computational budget. Although TEMI obtains the highest NMI of 78.1\%, it requires around 370 minutes, more than five times the running time of TAC++ for a marginal performance improvement. Other deep image clustering methods, such as IDC and CPP, require more than 500 minutes while exhibiting considerably lower NMI.

\begin{tcolorbox}[center, width=140mm, colback=blue!5!white,colframe=blue!75!black,colbacktitle=red!80!black]
\textbf{Takeaway 4:} LaIC methods achieve a more favorable trade-off between time efficiency and clustering effectiveness. 
\end{tcolorbox}

\section{Conclusion}
We introduce \texttt{VLM4Cluster}, a comprehensive and unified benchmark for image clustering in the era of pre-trained VLMs. Our framework systematically evaluates 17 representative methods spanning three paradigms along four evaluation dimensions across 20 datasets with diverse characteristics. 
Through fair comparisons and extensive analyses, our benchmark reveals several key insights into this promising research direction. We hope that \texttt{VLM4Cluster} will serve as a useful foundation and contribute to the continued development of this emerging field. Potential extensions of the benchmark and broader directions for future research are discussed in \textbf{Appendix~\ref{appendix_H}}.

\bibliographystyle{iclr2027_conference}
\IfFileExists{reference.bib}{%
    \bibliography{reference}%
}{%
    \IfFileExists{iclr2027_conference.bib}{\bibliography{iclr2027_conference}}{}%
}

\appendix
\newpage

\section{Preliminary}
\textbf{Notation.} We denote matrices and vectors by boldface uppercase and lowercase letters, respectively. Throughout this paper, $\mathbf{A}[i,j]$ denotes the $(i,j)$-th element of a matrix $\mathbf{A}$, and $\mathbf{a}[i]$ denotes the $i$-th element of a vector $\mathbf{a}$. We use $\mathbf{I}_{K}$ to represent the $K \times K$ identity matrix and $\mathbf{1}_{K}$ for the $K$-dimensional all-ones vector. We define $\sigma(\cdot)$ as the softmax function and $H(\cdot)$ as the entropy function. 
Let $\Phi(\mathbf{a}_i, \mathbf{a}_j) = \Vert{}\mathbf{a}_i - \mathbf{a}_j\Vert{}_2^2$ denote the squared Euclidean distance. We define $\mathcal{N}_q(\mathbf{a}, \Phi, \mathbf{A})$ as the index set of the $q$ nearest neighbors of $\mathbf{a}$ among the rows of $\mathbf{A}$, measured by the distance function $\Phi(\cdot, \mathbf{a})$. The indicator function $\mathds{1}\{\cdot\}$ evaluates to $1$ when its argument is true and $0$ otherwise.

\textbf{Zero-shot Classification.} Let $\mathcal{X}$ and $\mathcal{T}$ be the visual and textual input space respectively, modern {VLMs} adopt a dual-stream architecture with one text encoder $f_{\mathcal{T}}$ and one image encoder $f_{\mathcal{X}}$ to map inputs of two modalities into an uni-modal hyper-spherical feature space $\mathbb{S}^{d-1}=\left \{\mathbf{z}\in\mathbb{R}^{d}|\left \|\mathbf{z}\right \|_2=1\right \}$.  Considering a $K$-way image classification task with known class names $\left\{\mathbf{y}_1,\ldots,\mathbf{y}_K\right\}$, pre-trained VLMs make class prediction for any input $\mathbf{x}\in\mathcal{X}$ by computing
\begin{equation}
\label{eq1}
\arg\max_{i=1,\ldots K}\frac{\exp\big[ f_{\mathcal{X}}(\mathbf{x})^\top f_{\mathcal{T}}\big(\Delta(\mathbf{y}_i)\big)\big]}{\sum_{j=1}^K\exp\big[ f_{\mathcal{X}}(\mathbf{x})^\top f_{\mathcal{T}}\big(\Delta(\mathbf{y}_j)\big)\big]},
\end{equation}
where $\Delta(\mathbf{y}_j)\in\mathcal{T}$ with $\Delta(\cdot)$ as the prompt template for the input class name. 

\begin{definition*}[Image Clustering]\label{P1}
Given an unlabeled image dataset of interest $\mathbb{D}_{\mathcal{X}}=\left\{\mathbf{x}_1,\ldots\mathbf{x}_M\right\}$ comprising $K$ classes, the aim of image clustering is to group images into $K$ semantically distinct clusters $\mathbb{G}_1,\cdots,\mathbb{G}_K$. Ideally, each resulting cluster corresponds to a class.
\end{definition*}

Despite its remarkable effectiveness~\citep{radford2021learning} and provable guarantees~\citep{chen2024understanding}, the zero-shot paradigm in Eq.~(\ref{eq1}) relies on prior knowledge of the true class names, rendering it inapplicable in unsupervised settings.
To address this limitation, a popular strategy~\citep{cai2023semantic, li2024image, peng2025provable, peng2026delving, li2026self} is to 
leveraging wild textual data which can be collected almost for free in the open world. However, it is important to note that wild textual data inevitably contains a mixture of \textit{positive}\footnote{By definition, \textit{positive} nouns are those semantically \textit{relevant/similar} to \textit{any} class in a dataset while \textit{negative} nouns are those semantically \textit{irrelevant/dissimilar} to \textit{all} the classes.} and \textit{negative semantics} regarding to the image dataset of interest. In view of this, we propose to use the Huber contamination model~\citep{huber1992robust} to model the marginal distribution of wild textual data as follows:
\begin{definition}[Wild Data Distribution]
\label{D1}
Let $\mathbb{P}_{\text{pos}}$ and $\mathbb{P}_{\text{neg}}$ be the distributions of positive and negative textual data defined over $\mathcal{T}$, respectively. According to the Huber contamination model~\citep{huber1992robust}, the unlabeled textual data distribution $\mathbb{P}_{\text{wild}}$ can be modeled as follows:
\begin{equation}
\label{eq2}
\mathbb{P}_{\text{wild}}\triangleq\pi\cdot\mathbb{P}_{\text{pos}}+(1-\pi)\cdot\mathbb{P}_{\text{neg}},
\end{equation}
where $\pi\in(0,1]$ is typically unknown in practice.
\end{definition}
\begin{definition}[Empirical Wild Dataset]
\label{D2}
An empirical wild textual dataset $\mathcal{D}_{\mathcal{T}}$ is sampled independently and identically distributed (i.i.d.) from the wild data distribution $\mathbb{P}_{\text{wild}}$.
\end{definition}

\begin{definition*}[Language-assisted Image Clustering]\label{P2}
Given an unlabeled image dataset of interest $\mathbb{D}_{\mathcal{X}}=\left\{\mathbf{x}_1,\ldots\mathbf{x}_M\right\}$ comprising $K$ classes, the aim of language-assisted image clustering is to group images into $K$ semantically distinct clusters $\mathbb{G}_1,\cdots,\mathbb{G}_K$ under the guidance of the wild textual data $\mathcal{D}_{\mathcal{T}}=\left\{\mathbf{t}_1,\ldots,\mathbf{t}_N\right\}$. Ideally, each resulting cluster aligns with a single class.
\end{definition*}

For clarity of exposition, we denote $\mathbf{Z} = [\mathbf{z}_1, \dots, \mathbf{z}_M] \in \mathbb{R}^{D \times M}$ and $\mathcal{V} = \{\mathbf{v}_1, \dots, \mathbf{v}_N\}$ as the feature matrix and feature set for the image dataset $\mathbb{D}_{\mathcal{X}} = \{\mathbf{x}_1, \dots, \mathbf{x}_M\}$ and the wild textual dataset $\mathcal{D}_{\mathcal{T}} = \{\mathbf{t}_1, \dots, \mathbf{t}_N\}$, respectively, where $\mathbf{z}_i = f_{\mathcal{X}}(\mathbf{x}_i)$ and $\mathbf{v}_i = f_{\mathcal{T}}(\mathbf{t}_i)$. Following prior work~\citep{cai2023semantic, li2024image, peng2025provable, peng2026delving, li2026self}, our benchmark simulates the wild dataset $\mathcal{D}_{\mathcal{T}}$ by utilizing WordNet~\citep{WordNet}. To be specific, we define $\mathbf{t}_i = \Delta(\mathbf{c}_i)$, where $\{\mathbf{c}_1, \dots, \mathbf{c}_N\}$ is a set of nouns fetched from WordNet.

\section{Algorithm Details}
\label{appendix_a}
\subsection{Classical Image Clustering}
\textbf{$k$-means}~\citep{dhillon2001concept} seeks to find the hard cluster assignments $\{\mathbf{p}^*_i\}_{i=1}^{M}$ and cluster prototypes $\{\boldsymbol{\mu}^*_k\}_{k=1}^{K}$ that minimize the within-cluster sum of squares, i.e., 
\begin{equation}
\begin{split}
\{\mathbf{p}^*_i\}_{i=1}^{M}, \{\boldsymbol{\mu}^*_k\}_{k=1}^{K} = \underset{\{\mathbf{p}_i\}_{i=1}^{M}, \{\boldsymbol{\mu}_k\}_{k=1}^{K}}{\arg\min}&~\sum_{i=1}^{N} \sum_{k=1}^{K} \mathbf{p}_i[k]\cdot\Vert{}\mathbf{z}_i - \boldsymbol{\mu}_k\Vert{}^2_2 \\ \text{s.t.} & \quad \mathbf{p}_i \in [0,1]^K, \quad \sum_{k=1}^K \mathbf{p}_i[k] = 1, \quad \forall i=1,\dots,M, \\ & \quad \Vert{}\boldsymbol{\mu}_k\Vert{}_2 = 1, \quad \forall k=1,\dots,K. 
\end{split}
\end{equation}

\textbf{SC}~\citep{ng2001spectral} converts the image clustering task as a normalized graph cut problem given by:
\begin{equation}
\mathbf{Y}^\star=\underset{\mathbf{Y}\in\mathbb{R}^{M\times K}}{\arg\min}~\text{Tr}\left(\mathbf{Y}^\top\mathbf{L}\mathbf{Y}\right),\quad \text{s.t.} \quad \mathbf{Y}^\top\mathbf{Y}=\mathbf{I}_{K},
\end{equation}
where $\mathbf{I}_{K}$ is a $K\times K$ identity matrix, $\mathbf{A}\in\mathbb{R}^{M\times M}$ is an affinity matrix where the element $A[i,j]$ represents the similarity between $\mathbf{x}_i$ and $\mathbf{x}_j$, $\mathbf{L}=\mathbf{I}_{M}-\mathbf{D}^{-1/2}\mathbf{A}\mathbf{D}^{-1/2}$ is a normalized Laplacian matrix, $\mathbf{D}$ is a diagonal matrix of $\mathbf{D}[i,i]=\sum_{j}\mathbf{A}[i,j]$, and $\text{tr}(\cdot)$ denotes the trace of a matrix. The optimal spectral embedding matrix $\mathbf{Y}^\star$ consists of the top-$K$ minimum eigenvectors of $\mathbf{L}$. 

Following the most popular practice, we compute $\mathbf{A}$ by applying a Radial Basis Function (RBF) kernel to the latent representations, defined as $\mathbf{A}_{\text{RBF}}[i,j] = e^{-\Phi(\mathbf{z}_i,\mathbf{z}_j)/\tau}$, where $\tau>0$ is a hyperparameter.

\textbf{SSC-OMP}~\citep{you2016scalable} leverages the Orthogonal Matching Pursuit (OMP) algorithm to efficiently solve the following large-scale constrained optimization problem for each $\mathbf{x}_i\in\mathbb{D}_{\mathcal{X}}$:
\begin{equation}
    \mathbf{c}_i^{*}
    =
    \underset{\mathbf{c}_i \in \mathbb{R}^{M}}{\arg\min}~
    \left\|
        \mathbf{z}_i-\mathbf{Z}\mathbf{c}_i
    \right\|_2^2
    \quad
    \text{s.t.}~
    \|\mathbf{c}_i\|_0 \leq \epsilon,~\mathbf{c}_{i}[i]=0,
\end{equation}
where $\epsilon>0$ is a hyper-parameter. Once the optimal coefficient matrix $\mathbf{C}^*=\left[\mathbf{c}_1^{*},\ldots,\mathbf{c}_M^{*}\right]$ is obtained, the segmentation of images can be found by applying SC to the affinity matrix $\mathbf{A}^*=\vert{}\mathbf{C}^*\vert{} + \vert{}\mathbf{C}^{*\top}\vert{}$.

\textbf{EnSC-ORGEN}~\citep{you2016oracle} designs an oracle-based active set algorithm to 
to efficiently solve the following large-scale regularized optimization problem for each $\mathbf{x}_i\in\mathbb{D}_{\mathcal{X}}$:
\begin{equation}
    \mathbf{c}_i^{*}
    =
    \underset{\mathbf{c}_i \in \mathbb{R}^{M}}{\arg\min}~
    \lambda\|\mathbf{c}_i\|_1
    +
    \frac{1-\lambda}{2}\|\mathbf{c}_i\|_2^2
    +
    \frac{\gamma}{2}
    \left\|
        \mathbf{z}_i-\mathbf{Z}\mathbf{c}_i
    \right\|_2^2,
\end{equation}
where $\lambda\in[0,1)$ and
$\gamma>0$ are two hyper-parameters. Once the optimal coefficient matrix $\mathbf{C}^*=\left[\mathbf{c}_1^{*},\ldots,\mathbf{c}_M^{*}\right]$ is obtained, the segmentation of images can be found by applying SC to the affinity matrix $\mathbf{A}^*=\vert{}\mathbf{C}^*\vert{} + \vert{}\mathbf{C}^{*\top}\vert{}$.

\subsection{Deep Image Clustering}
\textbf{IDC}~\citep{liu2024interactive} is a plug-and-play method designed to boost the clustering performance of off-the-shelf clustering models with minimal interaction overhead. To begin with, IDC quantitatively evaluates sample values based on the scoring function $S(\cdot)$ defined as follows:
\begin{equation}
S(\mathbf{x}_i)=\log\left(1-\mathbf{z}_i^{\top}\boldsymbol{\mu}^{(i)}_1
        +\mathbf{z}_i^{\top}\boldsymbol{\mu}^{(i)}_2\right)-\log\sum_{j\in\mathcal{N}_q(\mathbf{z}_i,\Phi)}\Phi(\mathbf{z}_i,\mathbf{z}_j)+\min_{j\in\mathcal{J}}\log\!\left(1-\mathbf{z}_i^{\top}\mathbf{z}_{j}\right),
\end{equation}
where $\boldsymbol{\mu}^{(i)}_1$ and $\boldsymbol{\mu}^{(i)}_2$ denote the closest and second-closest cluster prototypes to $\mathbf{z}_i$, respectively, and $\mathcal{J}$ represents the indices of the selected samples. In accordance with the original IDC implementation, cluster prototypes are derived by applying $k$-means to $\mathbf{Z}$. Subsequently, the sample $\mathbf{x}_i$ yielding the highest score $S(\mathbf{x}_i)$ is iteratively selected until the maximum size $|\mathcal{J}| = L$ is reached. 

For each selected sample, IDC presents $T$ candidate centers and prompts the user to identify the candidate that is semantically consistent with the sample. According to the user feedback that could be either positive (selecting a candidate) or negative (rejecting all candidates) to the given sample, we further partition the index set $\mathcal{J}$ into $\mathcal{J}=\mathcal{J}_{\text{pos}}\cup\mathcal{J}_{\text{neg}}$.

Finally, IDC learns a clustering head $g(\cdot;\boldsymbol{\theta}):\mathbb{R}^D\rightarrow\mathbb{R}^K$ by minimizing the following objective function 
\begin{equation}
\small
\mathcal{L}_{\text{IDC}}=
-\frac{1}{|\mathcal{J}_{\text{pos}}|}\sum_{i\in\mathcal{J}_{\text{pos}}}\sum_{j=1}^K y_{ij}\log \mathbf{p}_{i}[j]
-\frac{1}{|\mathcal{J}_{\text{neg}}|}\sum_{i\in\mathcal{J}_{\text{neg}}}\sum_{j=1}^K \tilde{y}_{ij}\log \mathbf{p}_{i}[j]
-\frac{1}{M}\sum_{i=1}^M\sum_{j:\mathbf{p}_{i}[j]>\epsilon}\log \mathbf{p}_{i}[\hat{j}].
\end{equation}
where $\mathbf{p}_{i}=\sigma\left(g(\mathbf{z}_i;\boldsymbol{\theta})\right)$, $\hat{j}=\arg\max_{j={1,\ldots,K}}\mathbf{p}_{i}[j]$, and $\epsilon\in[0,1)$ is a hyperparameter. For each $i\in\mathcal{J}_{\text{pos}}$, $y_{ij}=1$ if the user assigns sample $\mathbf{x}_{i}$ to the $j$-th cluster, and $y_{ij}=0$ otherwise. For each $i\in\mathcal{J}_{\text{neg}}$, $\tilde{y}_{ij}=1$ if the $j$-th cluster is randomly assigned to sample $\mathbf{x}_i$, and $\tilde{y}_{ij}=0$ otherwise.

\textbf{SCAN}~\citep{van2020scan} learns a clustering head $g(\cdot;\boldsymbol{\theta}):\mathbb{R}^D\rightarrow\mathbb{R}^K$ by minimizing the following objective function
\begin{equation}
\mathcal{L}_{\text{SCAN}}
=
-\frac{1}{M}\sum_{i=1}^{M}
\sum_{j\in\mathcal{N}_q(\mathbf{z}_i,\Phi)}
\log\left(\mathbf{p}_i^{\top}\mathbf{p}_j\right)
+
\lambda H(\bar{\mathbf{p}}),
\end{equation}
where $\mathbf{p}_{i}=\sigma\left(g(\mathbf{z}_i;\boldsymbol{\theta})\right)$, $\bar{\mathbf{p}}=\frac{1}{M}\sum_{i=1}^M\mathbf{p}_{i}$, and $\lambda>0$ is a hyper-parameter. Following the original SCAN implementation, $L$ separate projection heads are trained in parallel, where $L=1$ for classical datasets and $L=10$ for all others.

\textbf{CPP}~\citep{chu2024image} seeks to learn (1) a projected embedding space for the feature matrix $\mathbf{Z}$ via a projection head $g(\cdot;\boldsymbol{\theta}):\mathbb{R}^D\rightarrow\mathbb{R}^{d}$, and (2) a cluster membership matrix $\mathbf{C}=[\mathbf{c}_1,\ldots,\mathbf{c}_M]\in\mathbb{R}^{M\times M}$ generated by another projection head $h(\cdot;\boldsymbol{\eta}):\mathbb{R}^D\rightarrow\mathbb{R}^{d}$ and a doubly stochastic projection $\Psi(\cdot)$, by minimizing the following objective function:
\begin{equation}
\mathcal{L}_{\text{CPP}}
=\log\det\left(\mathbf{I}_{d}+\frac{d}{M\epsilon}\hat{\mathbf{Z}}\hat{\mathbf{Z}}^\top\right)-\frac{1}{M}\sum_{i=1}^M\log\det\left(\mathbf{I}_{d}+\frac{d}{\epsilon}\hat{\mathbf{Z}}\text{diag}(\mathbf{c}_i)\hat{\mathbf{Z}}^\top\right),
\end{equation}
where $\epsilon>0$ is a hyper-parameter, $\hat{\mathbf{Z}}=[g(\mathbf{z}_1;\boldsymbol{\theta}),\ldots,g(\mathbf{z}_M;\boldsymbol{\theta})]\in\mathbb{R}^{d\times M}$, and $\mathbf{C}=\Psi(\hat{\mathbf{R}}^\top\hat{\mathbf{R}})$ with $\hat{\mathbf{R}}=[h(\mathbf{z}_1;\boldsymbol{\eta}),\ldots,h(\mathbf{z}_M;\boldsymbol{\eta})]\in\mathbb{R}^{d\times M}$. In accordance with the original CPP implementation, $\Psi(\cdot)$ is given by solving the following optimization problem:
\begin{equation}
\Psi(\boldsymbol{{\Pi}})=\min_{\mathbf{C}\in\mathbb{R}_{\geq0}^{M\times M}}~\operatorname{Tr}\left(\boldsymbol{{\Pi}}^\top\mathbf{C}\right)+\gamma\sum_{i=1}^MH(\mathbf{c}_i),~~\text{s.t.}~~ \mathbf{C}^\top\mathbf{1}_{M}=\mathbf{C}\mathbf{1}_{M}=\mathbf{1}_{M},
\end{equation}
where $\gamma>0$ is a hyper-parameter.

Once the training process is done, the optimized projection head $h(\cdot;\boldsymbol{\eta}^*)$ is used to compute $\hat{\mathbf{V}}^*$, which subsequently determines the final cluster membership matrix $\mathbf{C}^*$. SC is then applied to $\mathbf{C}^*$ to partition the images into $K$ clusters.

\textbf{TEMI}~\citep{adaloglou2023exploring} involves self-distillation training of a student clustering head $g(\cdot;\boldsymbol{\theta}):\mathbb{R}^D\rightarrow\mathbb{R}^{d}$ and a teacher clustering head $g(\cdot;\hat{\boldsymbol{\theta}}):\mathbb{R}^D\rightarrow\mathbb{R}^{d}$ by minimizing the following objective function: 
\begin{equation}
\mathcal{L}_{\text{TEMI}}
=\frac{1}{M}\sum_{i=1}^M\sum_{j\in\mathcal{N}_q(\mathbf{z}_i,\Phi)}w_{ij}\left[\log\sum_{k=1}^K\frac{\left(\mathbf{p}_{i}[k]\hat{\mathbf{p}}_{j}[k]\right)^\beta}{\bar{\mathbf{p}}[k]}+\log\sum_{k=1}^K\frac{\left(\mathbf{p}_{j}[k]\hat{\mathbf{p}}_{i}[k]\right)^\beta}{\bar{\mathbf{p}}[k]}\right],
\end{equation}
where $\beta>0$ is a hyper-parameter, $\mathbf{p}_{i}=\sigma\left(g(\mathbf{z}_i;\boldsymbol{\theta})/\tau\right)$, $\hat{\mathbf{p}}_{i}=\sigma\left(g(\mathbf{z}_i;\hat{\boldsymbol{\theta}})/\tau\right)$, $w_{ij}=\hat{\mathbf{p}}_{i}^\top\hat{\mathbf{p}}_{j}$, and $\bar{\mathbf{p}}=\frac{1}{M}\sum_{i=1}^M\mathbf{p}_{i}$. During training, the parameters of the student projection head $g(\cdot;\boldsymbol{\theta})$ is updated via back-propagation, the parameters of the teacher projection head $g(\cdot;\hat{\boldsymbol{\theta}})$ is updated using an exponential moving average, i.e., $\hat{\boldsymbol{\theta}}\leftarrow \lambda\hat{\boldsymbol{\theta}}+(1-\lambda)\boldsymbol{\theta}$ where $\lambda\in[0,1)$ is a momentum coefficient. In accordance with the original TEMI implementation, $L=50$ separate pairs of student and teacher projection heads are learned in parallel.

\textbf{PRO-DSC}~\citep{meng2025exploring} seeks to learn (1) a structured embedding space for the feature matrix $\mathbf{Z}$ via a projection head $g(\cdot;\boldsymbol{\theta}):\mathbb{R}^D\rightarrow\mathbb{R}^{d}$, and (2) a subspace-preserving self-expression coefficient matrix $\mathbf{C}\in\mathbb{R}^{M\times M}$ generated by another projection head $h(\cdot;\boldsymbol{\eta}):\mathbb{R}^D\rightarrow\mathbb{R}^{d}$ and a doubly stochastic projection $\Psi(\cdot)$, by minimizing the following objective function:
\begin{equation}
\mathcal{L}_{\text{PRO-DSC}}
=-\frac{1}{2}\log\det\left(\mathbf{I}_{d}+\epsilon\hat{\mathbf{Z}}\hat{\mathbf{Z}}^\top\right)+\frac{\lambda}{2}\|\hat{\mathbf{Z}}-\hat{\mathbf{Z}}\mathbf{C}\|_F^2+\beta\cdot\Omega(\mathbf{C}),
\end{equation}
where $\epsilon,\lambda,\beta>0$ are three positive hyper-parameters, $\hat{\mathbf{Z}}=\left[\frac{g(\mathbf{z}_1;\boldsymbol{\theta})}{\|g(\mathbf{z}_1;\boldsymbol{\theta})\|^2},\ldots,\frac{g(\mathbf{z}_M;\boldsymbol{\theta})}{\|g(\mathbf{z}_M;\boldsymbol{\theta})\|^2}\right]\in\mathbb{R}^{d\times M}$, $\mathbf{C}=\Psi(\hat{\mathbf{R}}^\top\hat{\mathbf{R}})$ with $\hat{\mathbf{R}}=\left[\frac{h(\mathbf{z}_1;\boldsymbol{\eta})}{\|h(\mathbf{z}_1;\boldsymbol{\eta})\|_2},\ldots,\frac{h(\mathbf{z}_M;\boldsymbol{\eta})}{\|h(\mathbf{z}_M;\boldsymbol{\eta})\|_2}\right]\in\mathbb{R}^{d\times M}$, and $\Omega(\mathbf{C})$ is the sum of the $K$ smallest eigenvalues of the Laplacian matrix of the affinity $\mathbf{A}=\vert{}\mathbf{C}\vert{} + \vert{}\mathbf{C}^{\top}\vert{}$. In accordance with the original PRO-DSC implementation, $\Psi(\cdot)$ is given by solving the following optimization problem:
\begin{equation}
\Psi(\boldsymbol{{\Pi}})=\min_{\mathbf{C}\in\mathbb{R}_{\geq0}^{M\times M}}~\operatorname{Tr}\left(\boldsymbol{{\Pi}}^\top\mathbf{C}\right)+\gamma\sum_{i=1}^MH(\mathbf{c}_i),~~\text{s.t.}~~ \mathbf{C}^\top\mathbf{1}_{M}=\mathbf{C}\mathbf{1}_{M}=\mathbf{1}_{M},
\end{equation}
where $\gamma>0$ is a hyper-parameter.

Once the training process is done, the optimized projection head $h(\cdot;\boldsymbol{\eta}^*)$ is used to compute $\hat{\mathbf{V}}^*$, which subsequently determines the final cluster membership matrix $\mathbf{C}^*$. SC is then applied to $\mathbf{C}^*$ to partition the images into $K$ clusters.

\subsection{Language-assisted Image Clustering}
In this section, we further categorize LaIC methods into two paradigms: post-hoc and training-based approaches. Specifically, post-hoc methods obtain cluster assignments by applying off-the-shelf classical clustering techniques, such as $k$-means and SC, whereas training-based methods produce cluster assignments by learning parametric clustering heads.

\subsubsection{Post-hoc Methods}
\textbf{TAC}~\citep{li2024image} produces clustering assignment by simply applying $k$-means on the concentrated features $[\mathbf{z}_i,\hat{\mathbf{z}}_i]_{i=1}^M\in\mathbb{R}^{2D\times M}$ where $\hat{\mathbf{z}}_i$ is the text counterpart for the image feature $\mathbf{z}_i$.

To compute text counterparts $\left\{\hat{\mathbf{z}}_1,\ldots,\hat{\mathbf{z}}_M\right\}$ in accordance with the original TAC implementation, we first perform $k$-means clustering on the image features $\mathbf{Z}$ to obtain a set of $L$ cluster prototypes $\left\{\hat{\boldsymbol{\mu}}_1,\dots,\hat{\boldsymbol{\mu}}_L\right\}$. Next, a high-confidence subset of text features, denoted as $\hat{\mathcal{V}}_{\text{TAC}}$, by filtering out ambiguous text features based on their softmax-normalized similarities to the cluster prototypes:
\begin{equation}
\label{eq15}
\hat{\mathcal{V}}_{\text{TAC}}=\bigcup_{k=1}^{L}\left\{\mathbf{v}\in\mathcal{V}\,\middle|\, \underset{j=1,\ldots,L}{\arg\max}~\mathbf{v}^\top\hat{\boldsymbol{\mu}}_j=k \text{ and } \frac{\exp\left(\mathbf{v}^\top\hat{\boldsymbol{\mu}}_k/\tau\right)}{\sum_{j=1}^L\exp\left[\mathbf{v}^\top\hat{\boldsymbol{\mu}}_j/\tau\right]}\geq\hat{\gamma}_k\right\},
\end{equation}
where $\tau$ is a temperature hyper-parameter and the confidence threshold $\hat{\gamma}_k$ is determined by the $q$-th highest softmax probability among the nouns assigned to $\hat{\boldsymbol{\mu}}_k$.

Finally, the text counterpart $\hat{\mathbf{z}}_i$ for each image feature $\mathbf{z}_i$ is computed via an attention-like similarity-weighted aggregation:
\begin{equation}
\label{eq16}
\hat{\mathbf{z}}_i=\sum_{\hat{\mathbf{v}}_j\in\hat{\mathcal{V}}_{\text{TAC}}}\left(\frac{\exp(\mathbf{z}_i^\top\hat{\mathbf{v}}_j/\tau)}{\sum_{\hat{\mathbf{v}}_k\in\hat{\mathcal{V}}_{\text{TAC}}}\exp(\mathbf{z}_i^\top\hat{\mathbf{v}}_k/\tau)}\cdot\hat{\mathbf v}_j\right),
\end{equation}
where $\tau$ is a hyper-parameter controlling the sharpness of the similarity distribution.

\textbf{GradNorm}~\citep{peng2025provable} produces clustering assignment by simply applying $k$-means on the concentrated features $[\mathbf{z}_i,\hat{\mathbf{z}}_i]_{i=1}^M\in\mathbb{R}^{2D\times M}$ where $\hat{\mathbf{z}}_i$ is the text counterpart for the image feature $\mathbf{z}_i$.

To compute the text counterparts $\left\{\hat{\mathbf{z}}_1,\ldots,\hat{\mathbf{z}}_M\right\}$ in accordance with the original GradNorm implementation, we first perform $k$-means clustering on the image features $\mathbf{Z}$ to obtain the pseudo-label $\tilde{y}_i\in\{1,\ldots,L\}$ for each image feature $\mathbf{z}_i$. Using these pseudo-labels, we learns a linear classifier parameterized by weights $\mathbf{W} = [\mathbf{w}_1, \ldots, \mathbf{w}_L]$ via solving the following optimization problem: 
\begin{equation}
\mathbf{W}^\star=\arg\min_{\mathbf{W}\in\mathcal{W}}-\frac{1}{N}\sum_{i=1}^N\log\frac{\exp(\mathbf{z}_i^\top\mathbf{w}_{\tilde{y}_i}/\tau)}{\sum_{j=1}^L\exp(\mathbf{z}_i^\top\mathbf{w}_{j}/\tau)},
\end{equation}
where $\tau>0$ is a hyper-parameter and $\mathcal{W}$ is a parameter space.

After training the classifier, we leverage the optimal weights $\mathbf{W}^\star$ to extract a positive subset of text features from $\mathcal{V}$. Specifically, we construct a refined vocabulary of highly confident features, denoted by $\hat{\mathcal{V}}_{\text{GradNorm}}$, by evaluating both the predicted pseudo-class assignment and the gradient norm magnitude of each candidate feature $\mathbf{v} \in \mathcal{V}$, i.e., 
\begin{equation}
\small
\hat{\mathcal{V}}_{\text{GradNorm}}=\bigcup_{k=1}^{L}\left\{\mathbf{v}\in\mathcal{V}\,\middle|\, \underset{j=1,\ldots,L}{\arg\max}~\mathbf{v}^\top\mathbf{w}^*_j=k \text{ and } \left\lVert \nabla_{\mathbf{W}^\star}\left(\frac{\exp\left(\mathbf{v}^\top\mathbf{w}_k^*/\tau\right)}{\sum_{j=1}^L\exp\left(\mathbf{v}^\top\mathbf{w}^*_j/\tau\right)}\right)\right\rVert_F^2\leq\hat{\gamma}_k\right\},
\end{equation}
where $\hat{\gamma}_k$ is a confidence threshold and the confidence threshold $\hat{\gamma}_k$ is determined by the $q$-th smallest gradient magnitude among the nouns assigned to $\hat{\boldsymbol{\mu}}_k$.

Finally, the text counterpart $\hat{\mathbf{z}}_i$ for each image feature $\mathbf{z}_i$ is computed via an attention-like similarity-weighted aggregation:
\begin{equation}
\hat{\mathbf{z}}_i=\sum_{\hat{\mathbf{v}}_j\in\hat{\mathcal{V}}_{\text{GradNorm}}}\left(\frac{\exp(\mathbf{z}_i^\top\hat{\mathbf{v}}_j/\kappa)}{\sum_{\hat{\mathbf{v}}_k\in\hat{\mathcal{V}}_{\text{GradNorm}}}\exp(\mathbf{z}_i^\top\hat{\mathbf{v}}_k/\kappa)}\cdot\hat{\mathbf v}_j\right),
\end{equation}
where $\kappa$ is a hyper-parameter controlling the sharpness of the similarity distribution.

\textbf{NTK-SC}~\citep{peng2026delving} extends SC by replacing the traditional RBF-based affinity matrix $\mathbf{A}_{\text{RBF}}$ 
with one derived from the Neural Tangent Kernel (NTK). Given a proxy network $g(\cdot;\boldsymbol{\theta}):\mathbb{R}^D\rightarrow\mathbb{R}$ differentiable w.r.t. parameters $\boldsymbol{\theta}\in\mathbb{R}^P$ (stretched into a single vector), the NTK-based affinity matrix is constructed as follows:
\begin{equation}
\mathbf{A}_{\text{NTK}}[i,j]=
\begin{cases}
\mathcal{K}_{\boldsymbol{\theta}}\left(\mathbf{z}_i,\mathbf{z}_j\right)&\text{ if } i\in\mathcal{N}_q(\mathbf{z}_j,\mathcal{K}_{\boldsymbol{\theta}},\mathbf{Z})\wedge j\in\mathcal{N}_q(\mathbf{z}_i,\mathcal{K}_{\boldsymbol{\theta}},\mathbf{Z}),\\
  0 &\text{  otherwise, } 
\end{cases}
\end{equation}
where
\begin{equation}
    \mathcal{K}_{\boldsymbol{\theta}}(\mathbf{z}_i,\mathbf{z}_j)= \left \langle\frac{\partial g_{\boldsymbol{\theta}}(\mathbf{z}_i)}{\partial \boldsymbol{\theta}},\frac{\partial g_{\boldsymbol{\theta}}(\mathbf{z}_j)}{\partial \boldsymbol{\theta}}  \right \rangle.
\end{equation}
Following the original NTK-SC implementation, we take $\boldsymbol{\theta}=\text{vec}\left(\hat{\mathcal{V}}_{\text{TAC}}\right)$ and 
\begin{equation}
g_{\boldsymbol{\theta}}\left(\mathbf{z}_i\right)=\log\sum_{\hat{\mathbf{v}}_j\in\hat{\mathcal{V}}_{\text{TAC}}}\exp\left({\mathbf{v}_j^\top \mathbf{z}_i/\tau}\right),
\end{equation}
where $\tau>0$ is a temperature hyper-parameter.

Same as TAC~\citep{li2024image} and GradNorm~\citep{peng2025provable}, NTK-SC can be applied to scenarios involving multiple prompt templates. Formally, given a $B$-sized pool of prompt templates $\{\Delta^{(1)},\ldots,\Delta^{(B)}\}$, let $\mathbf{A}^{(b)}_{\text{NTK}}$ be a prompt-specific affinity matrix for each $\Delta^{(b)}$ with $b\in[B]$. NTK-SC learns the ensembled affinity matrix $\hat{\mathbf{A}}$ by solving the following optimization problem:
\begin{equation}
\begin{split}
\boldsymbol{\beta}^*,\hat{\mathbf{A}}_{\text{NTK}}^* = \underset{\boldsymbol{\beta},\hat{\mathbf{A}}_{\text{NTK}}}{\arg\min}&~\sum_{b=1}^B\boldsymbol{\beta}[b]\cdot\ell(\hat{\mathbf{A}}_{\text{NTK}},\mathbf{A}^{(b)}_{\text{NTK}})+\lambda\|\hat{\mathbf{A}}_{\text{NTK}}-\mathbf{I}_{M}\|_F^2+\frac{\gamma}{2}\|\boldsymbol{\beta}\|^2_2 \\ \text{s.t.} & \quad \boldsymbol{\beta} \in [0,1]^B, \quad \sum_{b=1}^B \boldsymbol{\beta}[k] = 1,
\end{split}
\end{equation}
where $\lambda,\gamma>0$ are two hyper-parameters and 
\begin{equation}
    \ell(\hat{\mathbf{A}},\mathbf{A}^{(b)}_{\text{NTK}})=\frac{1}{2}\sum_{i,j,k,l=1}^Ma_{ij}^{(b)}a_{kl}^{(b)}\left(\frac{\hat{\mathbf{A}}[k,i]}{\sqrt{d_i^{(b)}d_k^{(b)}}}-\frac{\hat{\mathbf{A}}[l,j]}{\sqrt{d_j^{(b)}d_l^{(b)}}}\right)
\end{equation}
with $a_{ij}^{(b)}=\mathbf{A}^{(b)}_{\text{NTK}}[i,j]$ and $d_{i}^{(b)}=\sum_{j=1}^M\mathbf{A}^{(b)}_{\text{NTK}}[i,j]$.

Finally, NTK-SC produces clustering assignment by simply applying SC on $\hat{\mathbf{A}}_{\text{NTK}}^*$.

\subsubsection{Training-based Methods}
\textbf{SIC}~\citep{cai2023semantic} learns a clustering head $g(\cdot;\boldsymbol{\theta}):\mathbb{R}^D\rightarrow\mathbb{R}^K$ by minimizing the following objective function:
\begin{equation}
\mathcal{L}_{\text{SIC}}=-\frac{1}{M}\sum_{i=1}^M\left[\log\mathbf{p}_i^{\top}\mathbf{p}_{t_i}\right]-\frac{\lambda}{M}\sum_{i=1}^{M}\log\mathbf{p}_i[\tilde{y}_i]+\beta H(\bar{\mathbf{p}}),
\end{equation}
where $\mathbf{p}_{i}=\sigma\left(g(\mathbf{z}_i;\boldsymbol{\theta})\right)$, $\lambda, \beta$ are hyper-parameters, $t_i$ is randomly sampled from $\mathcal{N}_q(\hat{\mathbf{z}}_i, \Phi,\hat{\mathbf{Z}})$, $\tilde{y}_i$ is the pseudo-label of $\mathbf{x}_i$, and $\bar{\mathbf{p}}=\sum_{i=1}^M\mathbf{p}_i$. 

To compute pseudo-labels in accordance with the original SIC implementation, we first perform $k$-means clustering on the image features $\mathbf{Z}$ to obtain a set of $K$ cluster prototypes $\left\{\hat{\boldsymbol{\mu}}_1,\dots,\hat{\boldsymbol{\mu}}_L\right\}$. A high-confidence subset of text features, denoted as $\hat{\mathcal{V}}$, is then constructed by filtering out ambiguous samples based on their distances to the global mean $\bar{\mathbf{v}}$ and to the prototypes, i.e., 
\begin{equation}
\hat{\mathcal{V}}=\bigcup_{k=1}^{L}\left\{\mathbf{v}\in\mathcal{V}\mid 1-\mathbf{v}^\top\bar{\mathbf{v}}\geq\alpha \text{ and } \underset{j=1,\ldots,L}{\arg\max}~\mathbf{v}^\top\hat{\boldsymbol{\mu}}_j=k \text{ and } \mathbf{v}^\top\hat{\boldsymbol{\mu}}_k\geq\hat{\gamma}_k\right\},
\end{equation}
where $\bar{\mathbf{v}}=\ell_2\left(\sum_{i=1}^N\mathbf{v}_i\right)$, $\alpha$ is a hyper-parameter, and $\hat{\gamma}_k$ is a threshold set to the $q_1$-th largest cosine similarity to $\hat{\boldsymbol{\mu}}_k$ among the nouns assigned to that prototype.

Next, a second $k$-means clustering step is applied to $\mathbf{Z}$ to produce a new set of $K$ cluster prototypes $\left\{\tilde{\boldsymbol{\mu}}_1,\dots,\tilde{\boldsymbol{\mu}}_K\right\}$. For each cluster $j$, the refined set of features is defined as:
\begin{equation}
\tilde{\mathcal{V}}_j=\left\{\hat{\mathbf{v}}\in\hat{\mathcal{V}}\mid\hat{\mathbf{v}}^\top\tilde{\boldsymbol{\mu}}_j\geq\tilde{\gamma}_j \right\},
\end{equation}
where $\tilde{\gamma}_j$ is a threshold set  to the $q_2$-th largest cosine similarity to $\tilde{\boldsymbol{\mu}}_j$ among all the nouns.
Finally, we assign the pseudo-label $\tilde{y}_i$ to the sample $\mathbf{x}_i$ by computing $\tilde{y}_i=\arg\max_{j=1,\ldots,K}\mathbf{z}_i^\top\mathbf{h}_j$ where $\mathbf{h}_j=\ell_2\left(\sum_{\tilde{\mathbf{v}}\in\tilde{\mathcal{V}}_j}\tilde{\mathbf{v}}\right)$.

\textbf{TAC++}~\citep{li2024image} extends \textbf{TAC}~\citep{li2024image} by jointly learning two distinct clustering heads, $g(\cdot;\boldsymbol{\theta}):\mathbb{R}^D\rightarrow\mathbb{R}^{K}$ and $h(\cdot;\boldsymbol{\eta}):\mathbb{R}^D\rightarrow\mathbb{R}^{K}$, designed to process the image features and their corresponding text counterparts, respectively. Specifically, the two heads are jointly optimized by minimizing a cross-modal contrastive loss function:
\begin{equation}
\begin{split}
\mathcal{L}_{\text{TAC}}=
&-\sum_{i=1}^M\log\frac{\exp\left(\mathbf{p}_i^\top\mathbf{q}_{t_i}/\xi\right)}{\sum_{j=1}^K\exp\left(\mathbf{p}_i^\top\mathbf{q}_{t_j}/\xi\right)+\sum_{j\neq i}^K\exp\left(\mathbf{p}_i^\top\mathbf{p}_{j}/\xi\right)}\\
&-\sum_{i=1}^M\log\frac{\exp\left(\mathbf{q}_i^\top\mathbf{p}_{r_i}/\xi\right)}{\sum_{j=1}^K\exp\left(\mathbf{q}_i^\top\mathbf{p}_{r_j}/\xi\right)+\sum_{j\neq i}^K\exp\left(\mathbf{q}_i^\top\mathbf{q}_{j}/\xi\right)}\\
&-\log\sum_{i=1}^M\mathbf{p}_i^\top\mathbf{q}_i+\alpha\left[H(\bar{\mathbf{p}})+H(\bar{\mathbf{q}})\right],
\end{split}
\end{equation}
where $\mathbf{p}_{i}=\sigma\left(g(\mathbf{z}_i;\boldsymbol{\theta})\right)$, $\mathbf{q}_{i}=\sigma\left(h(\hat{\mathbf{z}}_i;\boldsymbol{\eta})\right)$, $t_i$ is randomly sampled from $\mathcal{N}_q(\hat{\mathbf{z}}_i, \Phi,\hat{\mathbf{Z}})$, $r_i$ is randomly sampled from $\mathcal{N}_q(\mathbf{z}_{i}, \Phi,\mathbf{Z})$, $\bar{\mathbf{p}}=M^{-1}\sum_{i=1}^M\mathbf{p}_i$, $\bar{\mathbf{q}}=M^{-1}\sum_{i=1}^M\mathbf{q}_i$, and $\alpha>0$ is a hyper-parameter.

\textbf{SEIC}~\citep{li2026self} jointly learns two distinct clustering heads to process the image features $\left\{\mathbf{z}_i\right\}_{i=1}^M$ and their corresponding textual counterparts $\left\{\hat{\mathbf{z}}_i\right\}_{i=1}^M$. Specifically, it optimizes an image head $g_1(\cdot;\boldsymbol{\theta}_1)\circ g_2(\cdot;\boldsymbol{\theta}_2):\mathbb{R}^D\rightarrow\mathbb{R}^d\rightarrow\mathbb{R}^{K}$ and a text head $h_1(\cdot;\boldsymbol{\eta}_1)\circ h_2(\cdot;\boldsymbol{\eta}_2):\mathbb{R}^D\rightarrow\mathbb{R}^d\rightarrow\mathbb{R}^{K}$ by minimizing the following objective function:
\begin{equation}
\small
\begin{split}
\mathcal{L}_{\text{SEIC}}=
&-\frac{\alpha}{2M}\sum_{i=1}^M\left[\log\frac{\exp\left(\mathbf{e}_i^\top\hat{\mathbf{e}}_i/\tau\right)}{\sum_{j=1}^M\exp\left(\mathbf{e}_i^\top\hat{\mathbf{e}}_j/\tau\right)}+\log\frac{\exp\left(\hat{\mathbf{e}}_i^\top\mathbf{e}_i/\tau\right)}{\sum_{j=1}^M\exp\left(\hat{\mathbf{e}}_i^\top\mathbf{e}_j/\tau\right)}\right]\\
&-\frac{\beta}{2K}\sum_{i=1}^K\left[\log\frac{\exp\left(\sum_{c=1}^Mp_c[i]q_c[i]/\hat{\tau}\right)}{\sum_{j=1}^K\exp\left(\sum_{c=1}^Mp_c[i]q_c[i]/\hat{\tau}\right)}+\log\frac{\exp\left(\sum_{c=1}^Mq_c[i]p_c[i]/\hat{\tau}\right)}{\sum_{j=1}^K\exp\left(\sum_{c=1}^Mq_c[i]p_c[i]/\hat{\tau}\right)}\right]\\
&-\frac{\lambda}{2K}\sum_{i=1}^K\left[\log\frac{\exp\left(\boldsymbol{\mu}_i^\top\hat{\boldsymbol{\mu}}_i/\kappa\right)}{\sum_{j=1}^K\exp\left(\boldsymbol{\mu}_i^\top\hat{\boldsymbol{\mu}}_j/\kappa\right)}+\log\frac{\exp\left(\hat{\boldsymbol{\mu}}_i^\top\boldsymbol{\mu}_i/\kappa\right)}{\sum_{j=1}^K\exp\left(\hat{\boldsymbol{\mu}}_i^\top\boldsymbol{\mu}_j/\kappa\right)}\right]\\
&-\gamma\sum_{i=1}^{K}\frac{\bar{p}[i]\log\bar{p}[i]+\bar{q}[i]\log\bar{q}[i]}{s_i},
\end{split}
\end{equation}
where $\mathbf{p}_{i}=\sigma\left(g_2(\mathbf{e}_i;\boldsymbol{\theta}_2)\right)\in\mathbb{R}^K$ with $\mathbf{e}_i=g_1(\mathbf{z}_i;\boldsymbol{\theta}_1)\in\mathbb{R}^d$, $\mathbf{q}_{i}=\sigma\left(h_2(\hat{\mathbf{e}}_i;\boldsymbol{\eta}_2)\right)\in\mathbb{R}^K$ with $\hat{\mathbf{e}}_i=h_1(\hat{\mathbf{z}}_i;\boldsymbol{\eta}_1)\in\mathbb{R}^d$, $\bar{\mathbf{p}}=M^{-1}\sum_{i=1}^M\mathbf{p}_i$, $\bar{\mathbf{q}}=M^{-1}\sum_{i=1}^M\mathbf{q}_i$, $\tau,\hat{\tau},\kappa,\alpha,\beta,\lambda,\gamma>0$ are hyper-parameters, $s_i=M^{-1}\sum_{c=1}^M\mathds{1}\left\{\max_{j=1,\ldots,K}p_c[j]=i\right\}$ and
\begin{equation}
\begin{split}
\boldsymbol{\mu}_k&=\sum_{i:\arg\max_{j}p_i[j]=k}\left(\frac{p_i[k]}{\sum_{j=1}^Mp_j[k]}\right)\cdot\mathbf{e}_i,\\
\hat{\boldsymbol{\mu}}_k&=\sum_{i:\arg\max_{j}q_i[j]=k}\left(\frac{q_i[k]}{\sum_{j=1}^Mq_j[k]}\right)\cdot\hat{\mathbf{e}}_i.
\end{split}
\end{equation}

Following the original SEIC implementation, the textual counterpart $\hat{\mathbf{z}}_i$ for a given image feature $\mathbf{z}_i$ is computed via Eq. (\ref{eq16}) while fixing $L=K$ in Eq. (\ref{eq15}).

\textbf{SAC}~\citep{zhang2026semantic}
jointly learns two distinct clustering heads, $g(\cdot;\boldsymbol{\theta}):\mathbb{R}^D\rightarrow\mathbb{R}^{K}$ and $h(\cdot;\boldsymbol{\eta}):\mathbb{R}^D\rightarrow\mathbb{R}^{K}$, designed to process the image features $\left\{\mathbf{z}_i\right\}_{i=1}^M$ and their corresponding text counterparts $\left\{\hat{\mathbf{z}}_i\right\}_{i=1}^M$, respectively, by minimizing the following objective function:
\begin{equation}
\begin{split}
\mathcal{L}_{\text{SAC}}=
&-\sum_{i=1}^M\alpha_i\log\frac{\exp\left(\mathbf{p}_i^\top\tilde{\mathbf{p}}_{i}/\tau\right)}{\sum_{j=1}^M\exp\left(\mathbf{p}_i^\top\tilde{\mathbf{p}}_{j}/\tau\right)}\\
&-\sum_{i=1}^M\beta_i\log\frac{\exp\left(\mathbf{p}_i^\top\tilde{\mathbf{q}}_{i}/\tau\right)}{\sum_{j=1}^M\exp\left(\mathbf{p}_i^\top\tilde{\mathbf{q}}_{j}/\tau\right)}
-\sum_{i=1}^M\beta_i\log\frac{\exp\left(\mathbf{q}_i^\top\tilde{\mathbf{p}}_{i}/\tau\right)}{\sum_{j=1}^M\exp\left(\mathbf{q}_i^\top\tilde{\mathbf{p}}_{j}/\tau\right)}\\
&-\frac{\lambda}{K}\sum_{i=1}^K\sum_{j=1}^M\mathbf{p}_j[i]\log\mathbf{q}_j[i]-\frac{\lambda}{M}\sum_{i=1}^M\sum_{j=1}^K\mathbf{p}_i[j]\log\mathbf{q}_i[j]\\
&+\gamma\left[H(\bar{\mathbf{p}})+H(\bar{\mathbf{q}})\right],
\end{split}
\end{equation}
where $\mathbf{p}_{i}=\sigma\left(g(\mathbf{z}_i;\boldsymbol{\theta})\right)$, $\mathbf{q}_{i}=\sigma\left(h(\hat{\mathbf{z}}_i;\boldsymbol{\eta})\right)$, $\tilde{\mathbf{p}}_{i}=\mathbb{E}_{j\in\mathcal{N}_q(\mathbf{z}_i, \Phi,\mathbf{Z})}\left[\mathbf{p}_{j}\right]$, $\tilde{\mathbf{q}}_{i}=\mathbb{E}_{j\in\mathcal{N}_q(\hat{\mathbf{z}}_i, \Phi,\hat{\mathbf{Z}})}\left[\mathbf{q}_{j}\right]$, $\bar{\mathbf{p}}=M^{-1}\sum_{i=1}^M\mathbf{p}_i$, $\bar{\mathbf{q}}=M^{-1}\sum_{i=1}^M\mathbf{q}_i$, $\lambda,\gamma>0$ are two hyper-parameters, and 
\begin{equation}
\begin{split}
\alpha_i&=\frac{\zeta}{1+\exp\left(-\mathbf{p}_i^\top\tilde{\mathbf{p}}_{i}/\kappa\right)}+\frac{1-\zeta}{1+\exp\left(-\mathbf{q}_i^\top\tilde{\mathbf{q}}_{i}/\kappa\right)},\\
\beta_i&=\frac{\zeta}{1+\exp\left(-\mathbf{p}_i^\top\tilde{\mathbf{q}}_{i}/\kappa\right)}+\frac{1-\zeta}{1+\exp\left(-\mathbf{q}_i^\top\tilde{\mathbf{p}}_{i}/\kappa\right)},
\end{split}
\end{equation}
with $\xi\in[0,1]$ and $\kappa>0$ as another two hyper-parameters. 

Unlike the original SAC implementation, which relies on MLLMs to generate the textual counterpart $\hat{\mathbf{z}}_i$ from the caption of image $\mathbf{x}_i$, this paper evaluates SAC using the TAC-style textual counterparts defined in Eq. (\ref{eq16}) to ensure a fair comparison with training-based LaIC methods.

\textbf{MAGIC}~\citep{zhangmagic} jointly learns two distinct clustering heads to process the image features $\left\{\mathbf{z}_i\right\}_{i=1}^M$ and their corresponding textual counterparts $\left\{\hat{\mathbf{z}}_i\right\}_{i=1}^M$. Specifically, it optimizes an image head $g_1(\cdot;\boldsymbol{\theta}_1)\circ g_2(\cdot;\boldsymbol{\theta}_2):\mathbb{R}^D\rightarrow\mathbb{R}^D\rightarrow\mathbb{R}^{K}$ and a text head $h_1(\cdot;\boldsymbol{\eta}_1)\circ h_2(\cdot;\boldsymbol{\eta}_2):\mathbb{R}^D\rightarrow\mathbb{R}^D\rightarrow\mathbb{R}^{K}$ by minimizing the following objective function:
\begin{equation}
\footnotesize
\begin{split}
\mathcal{L}_{\text{MAGIC}}=
&-\sum_{i=1}^K\left[\log\frac{\exp\left(\sum_{c=1}^Mp_c[i]q_{t_c}[i]/\tau\right)}{\sum_{j=1}^K\exp\left(\sum_{c=1}^Mp_c[j]q_{t_c}[j]/\tau\right)}+\log\frac{\exp\left(\sum_{c=1}^Mq_c[i]p_{t_c}[i]/\tau\right)}{\sum_{j=1}^K\exp\left(\sum_{c=1}^Mq_c[j]p_{t_c}[j]/\tau\right)}\right]\\
&-\sum_{i=1}^K\left[\log\frac{\exp\left(\sum_{c=1}^Mp_c[i]p_{r_c}[i]/\tau\right)}{\sum_{j=1}^K\exp\left(\sum_{c=1}^Mp_c[j]p_{r_c}[j]/\tau\right)}+\log\frac{\exp\left(\sum_{c=1}^Mq_c[i]q_{r_c}[i]/\tau\right)}{\sum_{j=1}^K\exp\left(\sum_{c=1}^Mq_c[j]q_{r_c}[j]/\tau\right)}\right]\\
&-\lambda\sum_{j=1}^K\sum_{i=1}^Mp_i[j]\log q_i[j]+\beta\left[H(\bar{\mathbf{p}})+H(\bar{\mathbf{q}})\right],
\end{split}
\end{equation}
where $\mathbf{p}_{i}=\sigma\left(g_2(\mathbf{e}_i;\boldsymbol{\theta}_2)\right)\in\mathbb{R}^K$ with $\mathbf{e}_i=g_1(\mathbf{z}_i;\boldsymbol{\theta}_1)\in\mathbb{R}^D$, $\mathbf{q}_{i}=\sigma\left(h_2(\hat{\mathbf{e}}_i;\boldsymbol{\eta}_2)\right)\in\mathbb{R}^K$ with $\hat{\mathbf{e}}_i=h_1(\hat{\mathbf{z}}_i;\boldsymbol{\eta}_1)\in\mathbb{R}^D$, $\left\{t_c\right\}=\mathcal{N}_1(\mathbf{z}_i, \Phi,\mathbf{Z})$, $\left\{r_c\right\}=\mathcal{N}_1(\hat{\mathbf{z}}_i, \Phi,\hat{\mathbf{Z}})$, $\tau, \lambda,\beta>0$ are three hyper-parameters, $\bar{\mathbf{p}}=M^{-1}\sum_{i=1}^M\mathbf{p}_i$, and $\bar{\mathbf{q}}=M^{-1}\sum_{i=1}^M\mathbf{q}_i$.

Following the original MAGIC implementation, we define
\begin{equation}
\begin{split}
g_1(\mathbf{z}_i;\boldsymbol{\theta}_1)&=\text{LayerNorm}\left(\text{Dropout}\left(\tilde{\mathbf{W}}_u^\top\text{GELU}\left(\tilde{\mathbf{W}}_d^\top\mathbf{z}_i\right)\right)\right)\\
h_1(\hat{\mathbf{z}}_i;\boldsymbol{\eta}_1)&=\text{LayerNorm}\left(\text{Dropout}\left(\hat{\mathbf{W}}_u^\top\text{GELU}\left(\hat{\mathbf{W}}_d^\top\hat{\mathbf{z}}_i\right)\right)\right),
\end{split}
\end{equation}
where $\tilde{\mathbf{W}}_u,\hat{\mathbf{W}}_u\in\mathbb{R}^{\frac{D}{4}\times D}$ denote the up-projection matrices, and $\tilde{\mathbf{W}}_d,\hat{\mathbf{W}}_d\in\mathbb{R}^{D\times \frac{D}{4}}$ denote the down-projection matrices.

Unlike the original MAGIC implementation, which relies on MLLMs to generate the textual counterpart $\hat{\mathbf{z}}_i$ from the caption of image $\mathbf{x}_i$, this paper evaluates MAGIC using the TAC-style textual counterparts defined in Eq. (\ref{eq16}) to ensure a fair comparison with training-based LaIC methods.

\section{Dataset Details}
\label{appendix_b}
\subsection{Classical Datasets}
\textbf{CIFAR-10}~\citep{alex2009learning} is a classical  dataset for visual recognition, consisting of 60,000 images across 10 classes. It is split into 50,000 training and 10,000 test images, with 6,000 images per class.

\textbf{CIFAR-20}~\citep{alex2009learning} refers to this coarse-label version of CIFAR-100, where each image is labeled only with its superclass. It is not a separate dataset but a re-labeling of CIFAR-100 that collapses the 100 categories into 20 superclasses.

\textbf{STL-10}~\citep{coates2011analysis} includes $96\times96$ images from the same 10 classes as CIFAR-10, with 5,000 images for training and 8,000 images for test.

\textbf{ImageNet-10}~\citep{chang2017deep} is a small-scale subset of the ImageNet-1K, including images belonging to the following 10 classes: n02056570, n02085936, n02128757, n02690373, n02692877, n03095699, n04254680, n04285008, n04467665, n07747607.

\textbf{ImageNet-Dogs}~\citep{chang2017deep} is a fine-grained subset of ImageNet-1K that focuses specifically on dog breeds. It is comprised of images from the following 15 dog categories: n02085936, n02086646, n02088238, n02091467, n02097130, n02099601, n02101388, n02101556, n02102177, n02105056, n02105412, n02105855, n02107142, n02110958, n02112137.
\subsection{Challenging Datasets}
\textbf{DTD}~\citep{cimpoi2014describing} is a dataset for the visual task that focuses on texture attributes. It contains 5,640 in-the-wild images annotated with 47 texture classes, with 120 images per class. The data is split in three equal parts, in train, validation and test, 40 images per class, for each split.

\textbf{UCF-101}~\citep{soomro2012ucf101} is a dataset for human action recognition in videos. It contains 13,320 video clips spanning 101 action classes, collected from realistic and unconstrained YouTube videos. 
We use the middle frame of each video clip as the input image.

\textbf{CIFAR-100}~\citep{alex2009learning} contains 60,000 images of size $32\times32$, split into 50,000 training and 10,000 test images. Unlike CIFAR-10, it includes 100 classes, each with 600 images.

\subsection{Fine-grained Datasets}
\textbf{Aircraft}~\citep{maji2013fine} is a fine-grained dataset for distinguishing visually similar aircraft types. It contains 10,000 images across 100 aircraft variants, with 100 images per class. The dataset uses a standard split of 3,334 training, 3333 validating, and 3,333 test images.

\textbf{Food}~\citep{bossard2014food} is a fine-grained dataset of 101 food categories with 101,000 images. For each class, 250 manually reviewed test images are provided as well as 750 training images.

\textbf{Pets}~\citep{pets} is a fine-grained dataset of 37 categories (including 25 dog breeds and 12 cat breeds) with 7,349 images. The dataset provides a standard split of about 3,680 training and 3,669 test images.

\textbf{Cars}~\citep{krause20133d} is a fine-grained dataset for distinguishing different car models. It contains 16,185 images across 196 classes (car makes, models, and years), with 8,144 training and 8,041 test images.

\textbf{Flowers}~\citep{nilsback2008automated} is a fine-grained image classification dataset containing 8,189 images across 102 flower categories. The dataset is split into 1,020 training, 1,020 validation, and 6,149 test images, with each class having between 40 and 258 images.
\subsection{Large-scale Datasets}

\textbf{Places365}~\citep{zhou2017places} is a large-scale dataset for scene recognition. It contains approximately 1.8 million training images across 365 scene categories, along with 36,500 validation and 328,500 test images.

\textbf{ImageNet-1K}~\citep{ImageNet} is a large-scale benchmark dataset for image classification, consisting of approximately 1.28 million training images and 50,000 validation images across 1,000 object categories.

\subsection{Out-of-distribution Datasets}

\textbf{ImageNet-C}~\citep{hendrycks2019benchmarking} is derived from ImageNet-1K by applying Gaussian noise, 
to the 50,000 validation images. The dataset has no training split and is used solely for evaluating generalization under distribution shifts.

\textbf{ImageNet-V2}~\citep{recht2019imagenet} is a reproduction benchmark of ImageNet-1K. It contains 10,000 images across the same 1,000 classes as ImageNet-1K, collected using a similar data sourcing and annotation pipeline but independently of the original validation set. The dataset has no training split and is used solely for evaluating generalization under distribution shifts.

\textbf{ImageNet-S}~\citep{wang2019learning} contains 50,000 black-and-white sketch images spanning 1,000 classes, matching the label space of ImageNet-1K. The dataset has no training split and is used solely for evaluating generalization under distribution shifts.

\textbf{ImageNet-A}~\citep{hendrycks2021natural} contains 7,500 images across 200 classes from ImageNet-1K, each of which are naturally occurring hard examples that standard models often misclassify. The dataset has no training split and is used solely for evaluating generalization under distribution shifts.

\textbf{ImageNet-R}~\citep{hendrycks2021many} contains 30,000 images spanning 200 classes from ImageNet-1K, where images are drawn from artistic domains such as paintings, sketches, cartoons, and sculptures. The dataset has no training split and is used solely for evaluating generalization under distribution shifts.

\begin{table*}[t]
\centering
\caption{A summary of datasets used for evaluation.}
\begin{tabular}{l|ccccc}
\textbf{Dataset} & \textbf{Training Split} & \textbf{Test Split} & \textbf{\# of Training} & \textbf{\# of Test} & \textbf{\# of Classes} \\
\multicolumn{6}{c}{\cellcolor{gray!40}\textbf{Classical Datasets}} \\
STL-10 & Train & Test & 5000 & 8000 & 10 \\
CIFAR-10 & Train & Test & 50000 & 10000 & 10 \\
CIFAR-20 & Train & Test & 50000 & 10000 & 20 \\
ImageNet-10 & Train & Test & 13000 & 500 & 10 \\
ImageNet-Dogs & Train & Test & 19500 & 750 & 15 \\
\multicolumn{6}{c}{\cellcolor{gray!40}\textbf{Challenging Datasets}} \\
DTD & Train+Val & Test & 3760 & 1880 & 47 \\
UCF-101 & Train & Test & 9537 & 3783 & 101 \\
CIFAR-100 & Train & Test & 50000 & 10000 & 100 \\
\multicolumn{6}{c}{\cellcolor{gray!40}\textbf{Fine-grained Datasets}} \\
Aircraft & Train & Test & 3334 & 3333 & 100 \\
Food & Train & Test & 75750 & 25250 & 101 \\
Pets & Train & Test & 3680 & 3669 & 37 \\
Cars & Train & Test & 8144 & 8041 & 196 \\
Flowers & Train & Test & 1020 & 6149 & 102 \\
\multicolumn{6}{c}{\cellcolor{gray!40}\textbf{Large-scale Datasets}} \\
Place365 & Train & Val & 1803460 & 36500 & 365 \\
ImageNet-1K & Train & Val & 1281167 & 50000 & 1000 \\
\multicolumn{6}{c}{\cellcolor{gray!40}\textbf{Out-of-distribution Datasets}} \\
ImageNet-C & -- & Val & -- & 50000 & 1000 \\
ImageNet-V2 & -- & Val & -- & 10000 & 1000 \\
ImageNet-S & -- & Val & -- & 50000 & 1000 \\
ImageNet-A & -- & Val & -- & 7500 & 200 \\
ImageNet-R & -- & Val & -- & 30000 & 200 \\
\end{tabular}
\label{tab6}
\end{table*}

\section{Metric Details}
\label{appendix_c}

\textbf{Clustering Accuracy (ACC)} quantifies the agreement between predicted cluster assignments and ground-truth labels. It is formally defined as:
$$
\mathrm{ACC}(\mathcal{G}) = \mathrm{E}_{i}\left[\mathds{1}\left\{y_i = \mathrm{Hungarian}(\hat{y}_i)\right\}\right],
$$
where $y_i$ represents the ground-truth label of the sample $\mathbf{x}_i$, and $\hat{y}_i$ is the cluster assignment generated by the clustering method $\mathcal{G}$. The function $\mathrm{Hungarian}(\hat{y}_i)$ denotes the mapped label obtained by optimally aligning the predicted clusters with the ground-truth labels using the Hungarian algorithm. 

\textbf{Normalized Mutual Information (NMI)} measures the degree of information sharing between the distributions of clustering labels and ground-truth labels, reflecting the amount of ground-truth information inferable from clustering labels. Ranging from 0 to 1, where 0 indicates complete irrelevance and 1 signifies perfect consistency, it is frequently employed in scenarios where the number of clusters differs from that of ground-truth categories. The calculation is as follows:
$$
\mathrm{NMI}(\mathcal{G})=\frac{2\cdot \sum_{i}\sum_{j} P(y_i, \hat{y}_j) \log \left( \frac{P(y_i, \hat{y}_j)}{P(y_i)P(\hat{y}_j)} \right)}{-\sum_{i} P(y_i) \log P(y_i)+-\sum_{j} P(\hat{y}_j) \log P(\hat{y}_j)},
$$
where $y_i$ is the class label of the sample $\mathbf{x}_i$, $\hat{y}_i$ is the cluster assignment generated by the clustering method $\mathcal{G}$, $P(y, \hat{y})$ is the joint probability of a sample belonging to true class $y$ and predicted cluster $\hat{y}$, and $P(y)$ and $P(\hat{y})$ are the marginal probabilities of the true classes and predicted clusters, respectively.

\textbf{Adjusted Rand Index (ARI)} evaluates the similarity between predicted cluster assignments and ground-truth labels by computing the agreement across all pairs of samples. It corrects the standard Rand Index for chance, ensuring that a random clustering yields an ARI close to $0$, while a perfect clustering evaluates to $1$. Formally, it is computed using the values from a contingency table mapping the true classes to the predicted clusters:
$$
\mathrm{ARI}(\mathcal{G}) = \frac{\sum_{i,j} \binom{n_{ij}}{2} - \left[ \sum_i \binom{a_i}{2} \sum_j \binom{b_j}{2} \right] \big/ \binom{M}{2}}{\frac{1}{2} \left[ \sum_i \binom{a_i}{2} + \sum_j \binom{b_j}{2} \right] - \left[ \sum_i \binom{a_i}{2} \sum_j \binom{b_j}{2} \right] \big/ \binom{M}{2}},
$$
where $n_{lc}=\sum_{i}\mathds{1}\left\{y_i=l \text{ and } \hat{y}_i=c\right\}$ is the number of samples simultaneously belonging to the $l$-th class and the $c$-th cluster generated by the clustering method $\mathcal{G}$, $a_i = \sum_{j=1}^K n_{ij}$, and $b_j = \sum_{i=1}^K n_{ij}$.

\textbf{Effective Robustness (ER)} evaluates a machine learning model's ability to generalize to out-of-distribution (OOD) data, beyond what would be predicted solely by its ID performance. Introduced in the context of distribution shifts, it quantifies a model's actual OOD accuracy relative to a baseline trend established by standard models. 
Suppose there are $L$ clustering methods $\mathcal{G}_1,\ldots,\mathcal{G}_L$. A baseline function $\Gamma(\mathcal{G}_i)$ is constructed to predict the OOD clustering performance of a given clustering method $\mathcal{G}_i$, i.e., $\Lambda_{\text{OOD}}(\mathcal{G}_i)$  where $\Lambda\in\{\mathrm{ACC},\mathrm{NMI},\mathrm{ARI}\}$. Given the single ID clustering performance of the clustering method $\mathcal{G}_i$, i.e., $\Lambda_{\text{ID}}(\mathcal{G}_i)$, the baseline function is instantiated as:
$$
\Gamma(\mathcal{G}_i)=\mathrm{Logit}^{-1}\left(w\cdot\mathrm{Logit}\left(\Lambda_{\text{ID}}(\mathcal{G}_i)\right)+b\right),
$$
where $w$ and $b$ are parameters, $\mathrm{p}\left(x\right)=\ln{\frac{p}{1-p}}$, and $\mathrm{Logit}^{-1}$ is the inverse function of $\mathrm{Logit}$. Since $\mathrm{Logit}(p)=w\cdot\mathrm{Logit}\left(p\right)+b$, the baseline function is essentially a linear function after applying a logit transformation on the accuracies. In othe words, we can obtain the optimal parameters $w^*$ and $b^*$ by solving a linear regression. Then the ER of a clustering method $\mathcal{G}_i$ with regard to the metric $\Lambda$ is evaluated as follows:
$$
\mathrm{ER}\left(\mathcal{G}_i;\Lambda\right)=\Lambda_{\text{OOD}}(\mathcal{G}_i)-\mathrm{Logit}^{-1}\left(w^*\cdot\mathrm{Logit}\left(\Lambda_{\text{ID}}(\mathcal{G}_i)\right)+b^*\right).
$$

\textbf{Silhouette Coefficient (SC)} evaluates the quality of the clusters generated by the clustering method $\mathcal{G}$ by measuring how well each sample is matched to its assigned cluster relative to the nearest alternative cluster. It captures both intra-cluster cohesion and inter-cluster separation, with values ranging from $-1$ to $1$. A value close to $1$ indicates that a sample is well matched to its assigned cluster and well separated from neighboring clusters, a value close to $0$ indicates that the sample lies near a cluster boundary, and a negative value suggests that the sample may have been assigned to an incorrect cluster. Formally, 
The calculation is as follows:
$$
\mathrm{SC}(\mathcal{G})=\frac{1}{M}\sum_{i=1}^{M}\frac{b(i)-a(i)}{\max\{a(i),b(i)\}},
$$
where, with $\hat{y}_i$ as the cluster assignment of the sample $\mathbf{x}_i$ generated by the clustering method $\mathcal{G}$,
$$
a(i)
=
\mathbb{E}_{\substack{j:\,\hat{y}_j=\hat{y}_i\\ j\neq i}}
\left[
\Phi\!\left(\mathbf{z}_i,\mathbf{z}_j\right)
\right],
$$
$$
b(i)=
\min_{c\neq \hat{y}_i}
\mathbb{E}_{j:\hat{y}_j=c}
\left[
\Phi\!\left(\mathbf{z}_i,\mathbf{z}_j\right)
\right].
$$

\textbf{Davies-Bouldin Index (DBI)} evaluates the internal quality of predicted clusters by computing the worst-case similarity ratio between each cluster and its most similar counterpart. It relies on both intra-cluster dispersion and inter-cluster separation, ensuring that tightly packed and highly separated clusters yield a DBI close to $0$, while overlapping or diffuse clusters evaluate to higher positive values. Formally, it is computed using the average of the maximum similarity measures across all clusters:
$$\mathrm{DBI}(\mathcal{G}) = \frac{1}{K} \sum_{i=1}^{K} \max_{j \neq i} \left( \frac{\mathbb{E}_{c:\hat{y}_c=i}
\left[
\Phi\!\left(\mathbf{z}_c,\boldsymbol{\mu}_i\right)
\right] + \mathbb{E}_{c:\hat{y}_c=j}
\left[
\Phi\!\left(\mathbf{z}_c,\boldsymbol{\mu}_j\right)
\right]}{\Phi\!\left(\boldsymbol{\mu}_i,\boldsymbol{\mu}_j\right)} \right),$$
where $\boldsymbol{\mu}_i=\mathbb{E}_{c:\hat{y}_c=i}
\left[\mathbf{z}_c\right]$ with $\hat{y}_i$ as the cluster assignment of $\mathbf{x}_i$ produced by the clustering method $\mathcal{G}$.

\textbf{Calinski-Harabasz Index (CHI)} evaluates the internal quality of predicted clusters by computing the ratio of the between-cluster dispersion to the within-cluster dispersion. It assesses both cluster cohesiveness and separation, ensuring that dense and well-separated clusters yield a higher CHI score, while overlapping or poorly defined clusters evaluate to lower values. Formally, it is computed using the traces of the between-cluster and within-cluster scatter matrices, scaled by their respective degrees of freedom:
$$\mathrm{CHI}(\mathcal{G}) = \frac{\sum_{k=1}^K n_k\cdot \Phi\!\left(\boldsymbol{\mu}_k,\boldsymbol{\mu}\right)}{\sum_{k=1}^K \sum_{i:\hat{y}_i=k} \Phi\!\left(\mathbf{z}_i,\boldsymbol{\mu}_k\right)} \times \frac{M - K}{K - 1},$$
where $\boldsymbol{\mu}_k=\mathbb{E}_{i:\hat{y}_i=k}
\left[\mathbf{z}_i\right]$ with $\hat{y}_i$ as the cluster assignment of $\mathbf{x}_i$ produced by the clustering method $\mathcal{G}$, $\boldsymbol{\mu}_k=\mathbb{E}_{i}
\left[\mathbf{z}_i\right]$, and $n_{k}=\sum_{i}\mathds{1}\left\{\hat{y}_i=k\right\}$.

\section{Experimental Setup}
\label{appendix_d}
Our experiments were primarily conducted on a Linux server equipped with an AMD EPYC 9254 24-core CPU, 188GB of RAM, and an NVIDIA RTX 2080 Ti GPU (12GB VRAM). The code was implemented using Python 3.8 and PyTorch 2.1.0.
We endeavor to follow the original implementations of the various image clustering methods described in their corresponding papers or source code, with only slight modifications detailed in Appendix~\ref{appendix_a} to either accommodate pre-trained VLMs or ensure a fair comparison with other methods. For classical image clustering methods, all of which are transductive, we train and evaluate them on the test split of each dataset. For deep image clustering methods and LaIC methods, we fit and evaluate them on the train and test split of each dataset, respectively. To mitigate the effects of randomness, we report the average results of each method on each dataset across 10 independent runs, each initialized with a different random seed. In consistent with most LaIC methods, we use the following $7$ templates to construct prompts for nouns from WordNet: \textit{itap of a $\left\{\right\}$, a bad photo of the $\left\{\right\}$, a origami $\left\{\right\}$, a photo of the large $\left\{\right\}$, a $\left\{\right\}$ in a video game, art of the $\left\{\right\}$, a photo of the small $\left\{\right\}$}. All images are resized to a resolution of 224$\times$224. Regarding hyper-parameters, we employed a grid search strategy across the search space detailed in Table~\ref{tab:hyperparameter_settings} to identify optimal hyper-parameters. This exhaustive tuning maximizes each method's performance, guaranteeing an accurate and unbiased comparison.



\newcommand{\hpset}[1]{#1}
\begin{longtable}{
  @{}
  >{\raggedright\arraybackslash}p{0.20\linewidth}
  >{\raggedright\arraybackslash}p{0.18\linewidth}
  >{\raggedright\arraybackslash}p{0.62\linewidth}
  @{}
}
\caption{Hyperparameter search spaces for different methods. For details on the meaning of these hyper-parameters, please refer to Appendix~\ref{appendix_a}.}
\label{tab:hyperparameter_settings}
\\

\textbf{Method}
& \textbf{Hyperparameter}
& \textbf{Values} \\
\endfirsthead

\multicolumn{3}{@{}l}{
  \tablename~\thetable\ (continued)
}\\
\textbf{Method}
& \textbf{Hyperparameter}
& \textbf{Searching Space} \\
\endhead

\multicolumn{3}{r@{}}{
  \footnotesize Continued on the next page
}\\
\endfoot

\endlastfoot

\hdashline 
General Settings
& $K$
& \# of classes\\
& $M$
& \# of training sample\\
& optimizer 
& \hpset{Adam, AdamW, SGD} \\
& learning rate
& \hpset{$1e-5, 5e-5, 1e-4, 5e-4, 1e-3, 5e-3, 1e-2$} \\
& weight decay
& \hpset{$0, 1e-4, 1e-3, 2e-3, 3e-3,\ldots,1e-2$} \\
& epochs
& \hpset{10, 20, 50, 100, 200, 500, 800} \\
& batch size
& \hpset{32, 64, 128, 256, 512, 1024, 2048, 4096, 8192} \\
\hdashline 
$k$-means
& N/A
& N/A  \\
\hdashline 
SC
& $\tau$
& \hpset{0.001, 0.005, 0.01, 0.05, 0.1, 0.5 1.0, 2.0, 5.0}  \\
\hdashline

SSC-OMP
& $\epsilon$
& \hpset{1, 5, 10, 50, 100, 200, 500, 1000} \\*
\hdashline
EnSC-ORGEN
& $\gamma$
& \hpset{0.01, 0.1, 1, 10, 20, 100, 200, 500} \\*
& $\lambda$
& \hpset{0, 0.1, 0.3, 0.5, 0.7, 0.9} \\*
\hdashline


IDC
& $L$
& \hpset{100, 500, 1000, 2000, 3000, 5000}\\
& $\epsilon$
& \hpset{0, 0.1, 0.3, 0.5, 0.7, 0.9, 0.99, 0.999}\\
\hdashline

SCAN
& $q$
& \hpset{1, 5, 10, 20, 50, 100} \\*
& $\lambda$
& \hpset{0.01, 0.5, 0.1, 1, 2, 5, 10} \\*
\hdashline

CPP
& $\gamma$
& $0.005,0.006,0.007,\ldots,0.2$\\*
& $\epsilon$
& \hpset{0.001, 0.005, 0.01, 0.05, 0.1, 0.5, 1.0} \\
\hdashline
TEMI
& $q$
& \hpset{1, 5, 10, 20, 25, 50, 100} 
\\
& $\tau$
& \hpset{0.01, 0.02, 0.03, 0.05, 0.8, 0.10, 0.15} 
\\
& $\beta$
& \hpset{0.55, 0.6, 0.65, 0.7} 
\\
& $\lambda$
& \hpset{0.9, 0.95, 0.99, 0.995, 0.999, 0.9999} 
\\
\hdashline

PRO-DSC
& $\epsilon$
& \hpset{0.001, 0.005, 0.01, 0.05, 0.1, 0.5, 1.0}
\\
& $\gamma$
& $0.005,0.006,0.007,\ldots,0.2$\\
& $\lambda$
& \hpset{100, 200, 300, 400, 500, 600, 800, 1000} 
\\
& $\beta$
& \hpset{$0.05, 0.1,0.2,\ldots,1.0,2.0,5.0$} 
\\
\hdashline


SIC
& $\beta$
& \hpset{
    0.01, 0.05, 0.1, 0.2, 0.5, 0.8, 1, 2, 5
} 
\\
& $\lambda$
& \hpset{
    0.01, 0.05, 0.1, 0.2, 0.5, 0.8, 1, 2, 5
} 
\\
& $\alpha$
& \hpset{
    $0.05, 0.10, 0.15, \ldots, 0.30$
}
\\
& $k$
& \hpset{
1, 5, 10, 20, 30, 50, 100, 500
}
\\
& $q_1$
& \hpset{
    1, 10, 50, 100, 200, 300, 500, 800, 1000
}
\\
& $q_2$
& \hpset{
1, 5, 10, 20, 30, 50, 100, 500
}
\\

\hdashline

TAC
& $\tau$
& \hpset{
    0.008, 0.01, 0.02, 0.03, 0.04,
      0.05, 0.06, 0.07, 0.08, 0.09, 0.10
} \\
& $q_1$
& \hpset{
    $1, 2, 3, \ldots, 10$
} \\
\hdashline

TAC++
& $\xi$
& \hpset{
    $0.01, 0.1, 0.5, 1, 2, 5, 10$
} \\
& $\alpha$
& \hpset{
    0.1, 0.5, 1, 2, 5, 8, 10
} \\
& $q_2$
& \hpset{
    1, 5, 10, 20, 30, 50, 100, 500
} \\
\hdashline

NTK-SC
& $\tau$
& $0.01, 0.02, 0.03,\ldots,0.10$
\\
& $q$
& \hpset{
    1, 5, 10, 20, 30, 50, 100, 500
} \\
& $\lambda$
& \hpset{
    0.1
}
\\*
& $\gamma$
& \hpset{
    10
}
\\*

\hdashline

GradNorm

& $\tau$
&
\hpset{
    0.001, 0.003, 0.005, 0.008, 0.01, 0.02, 0.03, 0.05, 0.06, 0.10
}\\*

& $\kappa$
& \hpset{
    0.001, 0.005, 0.008, 0.01, 0.02, 0.03, 0.04, 0.05, 0.06, 0.10
} \\*

\hdashline

SEIC
& $\tau$
& \hpset{
    0.001, 0.002, 0.005, 0.008, 0.01, 0.02, 0.03, 0.04, 0.05
} \\*

& $\hat{\tau}$
& \hpset{
    0.01, 0.05, 0.1, 0.2, 0.4, 0.5, 1.0
} 
\\*

& $\kappa$
& $0.1, 0.2, 0.3,\ldots,1.0$
\\*

& $\alpha$
& $0.1, 0.2, 0.3,\ldots,1.0, 2.0, 5.0$
\\*

& $\beta$
& $0.1, 0.2, 0.3,\ldots,1.0, 2.0, 5.0$
\\*

& $\lambda$
& $0.1, 0.2, 0.3,\ldots,1.0, 2.0, 5.0$
\\*

& $\gamma$
& $0.1, 0.2, 0.3,\ldots,1.0, 2.0, 5.0$
\\*

\hdashline

SAC
& $\tau$
& $0.01, 0.05, 0.1, 0.5, 1.0, 2.0, 5.0$ 
\\
& $\lambda$
& $0.1, 0.2, 0.3,\ldots,1.0, 2.0, 5.0$
\\
& $\gamma$
& $0.1, 0.2, 0.3,\ldots,1.0, 2.0, 5.0$
\\
& $\xi$
& $0.1, 0.2, 0.3,\ldots,1.0, 2.0, 5.0$
\\
& $\kappa$
& $0.1, 0.2, 0.3,\ldots,1.0, 2.0, 5.0$
\\
& $q$
& \hpset{
    1, 5, 10, 20, 30, 50, 100, 500
}
\\
\hdashline

MAGIC
& $\tau$
& $0.01, 0.02, 0.03, \ldots, 0.1$
\\
& $\lambda$
& $0.1, 0.2, 0.3,\ldots,1.0, 2.0, 5.0$
\\
& $\beta$
& $0.1, 0.2, 0.3,\ldots,1.0, 2.0, 5.0, 10.0$

\\

\end{longtable}

\section{Additional Results on Effective Analysis}
\label{appendix_e}

The classical, challenging, large-scale, and fine-grained clustering results using CLIP ViT-B/16 pre-trained on LAION-400M are shown in Tables~\ref{tab:classical_b16}, \ref{tab:challenging_b16}, \ref{tab:large_scale_b16}, and \ref{tab:fine_grained_b16}, respectively.

The classical, challenging, large-scale, and fine-grained clustering results using CLIP ViT-L/14 pre-trained on LAION-400M are shown in Tables~\ref{tab:classical_l14}, \ref{tab:challenging_l14}, \ref{tab:large_scale_l14}, and \ref{tab:fine_grained_l14}, respectively.

The classical, challenging, large-scale, and fine-grained clustering results using CLIP ViT-B/32 pre-trained on LAION-2B are shown in Tables~\ref{tab:laion2b_b32_classical}, \ref{tab:laion2b_b32_challenging}, \ref{tab:laion2b_b32_large_scale}, and \ref{tab:laion2b_b32_fine_grained}, respectively.

The classical, challenging, large-scale, and fine-grained clustering results using SigLip ViT-B/16 pre-trained on WebLI are shown in Tables~\ref{tab:siglip_b16_classical}, \ref{tab:siglip_b16_challenging}, \ref{tab:siglip_b16_largescale}, and \ref{tab:siglip_b16_finegrained}, respectively.

\begin{table*}[tb]
\centering
\caption{Clustering results on classical datasets using CLIP ViT-B/16
pre-trained on LAION-400M. The best result in each metric column is
highlighted in \colorbox[HTML]{FFF2CC}{\textbf{bold}}.}
\label{tab:classical_b16}

\resizebox{\textwidth}{!}{%
\begin{tabular}{l|ccc|ccc|ccc|ccc|ccc|ccc}
\toprule
Dataset
& \multicolumn{3}{c|}{STL-10}
& \multicolumn{3}{c|}{CIFAR-10}
& \multicolumn{3}{c|}{CIFAR-20}
& \multicolumn{3}{c|}{ImageNet-10}
& \multicolumn{3}{c|}{ImageNet-Dogs}
& \multicolumn{3}{c}{Average} \\
Metric
& NMI & ACC & ARI
& NMI & ACC & ARI
& NMI & ACC & ARI
& NMI & ACC & ARI
& NMI & ACC & ARI
& NMI & ACC & ARI \\
\midrule


\multicolumn{19}{c}{%
  \cellcolor{gray!40}\textbf{Classical Image Clustering}} \\

$k$-means
& 95.6 & 98.2 & 96.0
& 78.7 & 80.8 & 71.3
& 61.6 & 55.7 & 40.2
& 97.4 & 98.4 & 96.5
& 64.8 & 61.2 & 47.2
& 79.6 & 78.9 & 70.2 \\

SC
& 95.3 & 98.1 & 95.9
& 77.8 & 81.5 & 72.2
& 54.1 & 53.5 & 39.2
& 96.6 & 98.0 & 95.7
& 46.7 & 45.1 & 30.0
& 74.1 & 75.2 & 66.6 \\

SSC-OMP
& 82.4 & 84.1 & 76.9
& 73.9 & 76.7 & 66.2
& 58.0 & 50.8 & 37.6
& 86.7 & 90.0 & 81.8
& 34.0 & 36.9 & 18.4
& 67.0 & 67.7 & 56.2 \\

EnSC-ORGEN
& 83.4 & 88.3 & 78.8
& 78.6 & 83.5 & 73.8
& 62.3 & 53.3 & 39.3
& 79.9 & 89.2 & 77.9
& 49.4 & 53.7 & 34.5
& 70.7 & 73.6 & 60.9 \\

\midrule
\multicolumn{19}{c}{%
  \cellcolor{gray!40}\textbf{Deep Image Clustering}} \\

IDC
& 96.2 & 98.5 & 96.7
& 80.7 & 87.4 & 73.8
& 60.1 & 55.4 & 39.8
& 98.2 & 99.0 & 97.8
& 67.6 & 69.7 & 51.0
& 80.6 & 82.0 & 71.8 \\

SCAN
& \best{96.5} & \best{98.6} & \best{97.0}
& 88.4 & 94.6 & 88.6
& 63.4 & 59.9 & 45.7
& 98.7 & 99.4 & 98.7
& 68.2 & 64.8 & 52.0
& 83.0 & 83.5 & 76.4 \\

CPP
& 95.0 & 97.9 & 95.5
& 87.6 & 94.1 & 87.5
& 64.5 & 60.2 & 43.2
& 98.7 & 99.4 & 98.7
& 70.2 & 72.7 & 54.5
& 83.2 & 84.8 & 75.9 \\

TEMI
& 96.4 & 98.5 & 96.8
& 88.3 & 94.3 & 87.9
& 63.3 & 59.5 & 44.2
& 98.6 & 99.0 & 98.6
& 71.3 & 73.3 & 57.0
& 83.6 & 84.9 & 76.9 \\

PRO-DSC
& 96.1 & 98.5 & 96.6
& 86.2 & 93.2 & 85.4
& 65.8 & 63.8 & 49.2
& 96.1 & 97.8 & 95.2
& 72.3 & 73.2 & 58.5
& 83.3 & 85.3 & 77.0 \\

\midrule
\multicolumn{19}{c}{%
  \cellcolor{gray!40}\textbf{Language-assisted Image Clustering}} \\

SIC
& 96.3 & \best{98.6} & 96.8
& \best{89.2} & 95.0 & 89.4
& 59.7 & 58.8 & 44.0
& \best{99.2} & \best{99.6} & \best{99.1}
& 66.5 & 64.1 & 52.5
& 82.2 & 83.2 & 76.4 \\

TAC
& 95.9 & 98.4 & 96.4
& 87.6 & 94.2 & 87.8
& 67.9 & 61.8 & 49.0
& 98.6 & 99.2 & 98.2
& 77.2 & 77.3 & 66.0
& 85.4 & 86.2 & 79.5 \\

TAC++
& 96.1 & 98.5 & 96.6
& 89.0 & \best{95.1} & 89.5
& 65.0 & 65.0 & 49.4
& 98.4 & 99.2 & 98.2
& 79.7 & 83.9 & 71.5
& 85.7 & 88.3 & 81.0 \\

SAC
& 96.0 & 98.4 & 96.5
& 87.6 & 94.3 & 87.9
& 63.4 & 63.4 & 48.5
& 97.1 & 98.4 & 96.5
& 76.7 & 79.5 & 67.8
& 84.2 & 86.8 & 79.4 \\

GradNorm
& 95.6 & 98.2 & 96.0
& 87.1 & 94.0 & 87.3
& 64.0 & 63.0 & 47.4
& 97.9 & 98.8 & 97.4
& 77.4 & 79.2 & 67.0
& 84.4 & 86.6 & 79.0 \\

NTK-SC
& 96.1 & 98.4 & 96.5
& 88.9 & 95.0 & 89.4
& 68.5 & 57.9 & 47.3
& 98.8 & 99.4 & 98.7
& 83.4 & 85.2 & 75.3
& 87.1 & 87.2 & 81.4 \\

SEIC
& 96.3 & 98.5 & 96.8
& 87.0 & 95.0 & 87.5
& 62.1 & 61.7 & 47.0
& 97.9 & 99.0 & 97.8
& 73.3 & 73.2 & 62.0
& 83.3 & 85.5 & 78.2 \\

MAGIC
& 96.0 & 98.4 & 96.6
& 89.0 & \best{95.1} & \best{89.6}
& \best{70.7} & \best{73.5} & \best{58.9}
& 98.7 & 99.4 & 98.7
& \best{83.8} & \best{85.7} & \best{76.5}
& \best{87.7} & \best{90.4} & \best{84.0} \\

\bottomrule
\end{tabular}%
}
\end{table*}

\begin{table*}[tb]
\centering
\caption{Clustering results on challenging datasets with CLIP ViT-B/16
pre-trained on LAION-400M. The best result in each metric column is
highlighted in \colorbox[HTML]{FFF2CC}{\textbf{bold}}.}
\label{tab:challenging_b16}

\resizebox{0.8\textwidth}{!}{%
\begin{tabular}{l|ccc|ccc|ccc|ccc}
\toprule
Dataset
& \multicolumn{3}{c|}{CIFAR-100}
& \multicolumn{3}{c|}{DTD}
& \multicolumn{3}{c|}{UCF101}
& \multicolumn{3}{c}{Average} \\
Metric
& NMI & ACC & ARI
& NMI & ACC & ARI
& NMI & ACC & ARI
& NMI & ACC & ARI \\
\midrule

zero-shot
& 75.2 & 71.4 & 53.4
& 62.7 & 50.9 & 35.3
& 80.6 & 67.0 & 53.1
& 72.8 & 63.1 & 47.3 \\
\midrule

\multicolumn{13}{c}{%
  \cellcolor{gray!40}\textbf{Classical Image Clustering}} \\

$k$-means
& 70.9 & 55.2 & 42.5
& 67.5 & 54.7 & 39.7
& 83.3 & 62.8 & 55.8
& 73.9 & 57.6 & 46.0 \\

SC
& 69.2 & 55.4 & 44.9
& 66.3 & 56.6 & 42.8
& 82.9 & 64.9 & 56.5
& 72.8 & 59.0 & 48.0 \\

SSC-OMP
& 62.4 & 46.4 & 31.9
& 58.9 & 47.3 & 31.3
& 66.4 & 40.0 & 28.0
& 62.6 & 44.6 & 30.4 \\

EnSC-ORGEN
& 69.5 & 55.0 & 40.3
& 64.4 & 54.5 & 36.6
& 77.6 & 51.0 & 42.1
& 70.5 & 53.5 & 39.7 \\

\midrule
\multicolumn{13}{c}{%
  \cellcolor{gray!40}\textbf{Deep Image Clustering}} \\

IDC
& 70.6 & 58.2 & 44.0
& 70.1 & \best{62.4} & \best{47.3}
& 80.8 & 66.2 & 52.9
& 73.8 & 62.3 & 48.0 \\

SCAN
& 69.5 & 46.0 & 38.5
& 67.2 & 55.9 & 41.4
& 74.2 & 40.1 & 35.4
& 70.3 & 47.3 & 38.4 \\

CPP
& 76.2 & 63.4 & 43.4
& 66.5 & 57.9 & 40.8
& 81.5 & 62.4 & 54.0
& 74.7 & 61.2 & 46.0 \\

TEMI
& 76.8 & \best{69.6} & \best{56.0}
& 68.9 & 58.6 & 44.4
& 82.4 & 68.3 & 59.0
& 76.0 & 65.5 & 53.1 \\

PRO-DSC
& 76.6 & 68.1 & 53.8
& 67.9 & 57.3 & 40.9
& 84.6 & 68.9 & 62.0
& 76.3 & 64.8 & 52.2 \\

\midrule
\multicolumn{13}{c}{%
  \cellcolor{gray!40}\textbf{Language-assisted Image Clustering}} \\

SIC
& 71.4 & 56.2 & 44.0
& 70.1 & 59.3 & 45.5
& 81.2 & 66.2 & 55.6
& 74.2 & 60.6 & 48.4 \\

TAC
& 74.2 & 60.0 & 46.8
& 68.5 & 58.3 & 40.1
& 83.6 & 65.6 & 56.0
& 75.4 & 61.3 & 47.6 \\

TAC++
& 74.7 & 66.4 & 52.3
& 68.4 & 59.3 & 43.0
& 82.5 & 69.8 & 60.4
& 75.2 & 65.2 & 51.9 \\

SAC
& 74.9 & 65.7 & 52.2
& 68.5 & 58.1 & 43.8
& 81.9 & 68.6 & 59.7
& 75.1 & 64.1 & 51.9 \\

GradNorm
& 73.1 & 61.3 & 45.4
& 70.2 & 58.3 & 44.2
& 84.2 & 68.4 & 56.6
& 75.8 & 62.7 & 48.7 \\

NTK-SC
& \best{77.9} & 69.2 & 53.7
& \best{70.4} & 61.1 & 45.8
& \best{86.4} & \best{74.6} & \best{66.1}
& \best{78.2} & \best{68.3} & \best{55.2} \\

SEIC
& 74.7 & 69.1 & 53.7
& 67.9 & 57.7 & 43.8
& 80.5 & 66.8 & 56.3
& 74.4 & 64.5 & 51.3 \\

MAGIC
& 74.1 & 62.2 & 49.4
& 63.7 & 49.8 & 36.0
& 75.9 & 52.6 & 44.0
& 71.3 & 54.9 & 43.1 \\

\bottomrule
\end{tabular}%
}
\end{table*}
\begin{table*}[tb]
\centering
\caption{Clustering results on large-scale datasets with CLIP ViT-B/16
pre-trained on LAION-400M. The best result in each metric column is
highlighted in \colorbox[HTML]{FFF2CC}{\textbf{bold}}.}
\label{tab:large_scale_b16}

\resizebox{0.7\textwidth}{!}{%
\begin{tabular}{l|ccc|ccc|ccc}
\toprule
Dataset
& \multicolumn{3}{c|}{Places}
& \multicolumn{3}{c|}{ImageNet}
& \multicolumn{3}{c}{Average} \\
Metric
& NMI & ACC & ARI
& NMI & ACC & ARI
& NMI & ACC & ARI \\
\midrule


\multicolumn{10}{c}{%
  \cellcolor{gray!40}\textbf{Classical Image Clustering}} \\

$k$-means
& 58.8 & 30.9 & 18.8
& 77.0 & 46.4 & 35.3
& 67.9 & 38.7 & 27.1 \\

SC
& 57.4 & 30.8 & 19.0
& 75.4 & 43.3 & 33.1
& 66.4 & 37.0 & 26.0 \\

SSC-OMP
& 52.0 & 24.3 & 11.2
& 66.1 & 31.7 & 13.7
& 59.1 & 28.0 & 12.5 \\

EnSC-ORGEN
& 61.2 & 32.8 & 20.7
& 75.4 & 47.1 & 27.7
& 68.3 & 40.0 & 24.2 \\

\midrule
\multicolumn{10}{c}{%
  \cellcolor{gray!40}\textbf{Deep Image Clustering}} \\

IDC
& 58.7 & 32.2 & 18.5
& 76.6 & 47.8 & 34.5
& 67.7 & 40.0 & 26.5 \\

SCAN
& 59.6 & 32.8 & 19.8
& 78.2 & 50.4 & 38.4
& 68.9 & 41.6 & 29.1 \\

CPP
& 54.5 & 27.9 & 13.0
& 73.2 & 43.9 & 14.2
& 63.9 & 35.9 & 13.6 \\

TEMI
& 60.8 & 35.2 & 21.2
& \best{81.0} & \best{58.2} & \best{44.6}
& \best{70.9} & \best{46.7} & \best{32.9} \\

PRO-DSC
& 54.8 & 26.7 & 13.6
& 74.6 & 45.8 & 26.6
& 64.7 & 36.3 & 20.1 \\

\midrule
\multicolumn{10}{c}{%
  \cellcolor{gray!40}\textbf{Language-assisted Image Clustering}} \\

SIC
& 60.7 & 31.0 & 20.9
& 79.9 & 54.1 & 41.4
& 70.3 & 42.6 & 31.2 \\

TAC
& 59.9 & 33.8 & 20.0
& 79.6 & 51.4 & 39.8
& 69.8 & 42.6 & 29.9 \\

TAC++
& 59.6 & 34.4 & 20.3
& 79.9 & 57.7 & 43.4
& 69.8 & 46.0 & 31.9 \\

SAC
& 59.2 & 34.8 & 21.0
& 80.2 & 58.1 & 44.4
& 69.7 & 46.5 & 32.7 \\

GradNorm
& 60.2 & 33.3 & 20.1
& 79.8 & 52.0 & 40.3
& 70.0 & 42.7 & 30.2 \\

NTK-SC
& \best{61.5} & \best{35.7} & \best{22.7}
& 79.6 & 56.8 & 41.4
& 70.6 & 46.2 & 32.0 \\

SEIC
& 59.8 & 32.0 & 18.5
& 77.8 & 48.9 & 34.1
& 68.8 & 40.4 & 26.3 \\

MAGIC
& 60.5 & 34.7 & 21.0
& 77.8 & 46.3 & 35.0
& 69.1 & 40.5 & 28.0 \\

\bottomrule
\end{tabular}%
}
\end{table*}

\begin{table*}[tb]
\centering
\caption{Clustering results on fine-grained datasets with CLIP ViT-B/16
pre-trained on LAION-400M. The best result in each metric column is
highlighted in \textbf{bold}.}
\label{tab:fine_grained_b16}

\resizebox{\textwidth}{!}{%
\begin{tabular}{l|ccc|ccc|ccc|ccc|ccc|ccc}
\toprule
Dataset
& \multicolumn{3}{c|}{Aircraft}
& \multicolumn{3}{c|}{Flowers}
& \multicolumn{3}{c|}{Food}
& \multicolumn{3}{c|}{Cars}
& \multicolumn{3}{c|}{Pets}
& \multicolumn{3}{c}{Average} \\
Metric
& NMI & ACC & ARI
& NMI & ACC & ARI
& NMI & ACC & ARI
& NMI & ACC & ARI
& NMI & ACC & ARI
& NMI & ACC & ARI \\
\midrule


\multicolumn{19}{c}{%
  \cellcolor{gray!40}\textbf{Classical Image Clustering}} \\

$k$-means
& 52.2 & 25.7 & 15.4
& 91.5 & 76.6 & 75.9
& 80.9 & 64.6 & 63.4
& 84.0 & 58.8 & 51.7
& 81.4 & 71.4 & 63.1
& 78.0 & 59.4 & 53.9 \\

SC
& 47.4 & 22.8 & 13.3
& 88.2 & 77.4 & 75.0
& 74.2 & 63.0 & 54.4
& 81.1 & 54.4 & 49.4
& 63.0 & 50.7 & 43.3
& 70.8 & 53.7 & 47.1 \\

SSC-OMP
& 43.2 & 19.4 & 8.0
& 51.1 & 30.7 & 19.4
& 54.0 & 41.5 & 28.3
& 66.2 & 40.5 & 25.6
& 44.9 & 32.2 & 19.1
& 51.9 & 32.9 & 20.1 \\

EnSC-ORGEN
& 49.1 & 25.0 & 12.7
& 87.9 & 73.8 & 65.9
& 79.1 & 73.8 & 61.1
& 80.0 & 61.5 & 47.1
& 64.3 & 54.5 & 39.2
& 72.1 & 57.7 & 45.2 \\

\midrule
\multicolumn{19}{c}{%
  \cellcolor{gray!40}\textbf{Deep Image Clustering}} \\

IDC
& 50.8 & 26.9 & 14.6
& 72.9 & 47.9 & 39.9
& 77.9 & 70.5 & 55.6
& 82.5 & 61.4 & 51.1
& 83.7 & 76.9 & 68.0
& 73.5 & 56.7 & 45.8 \\

SCAN
& 44.4 & 15.3 & 8.3
& 80.4 & 55.2 & 53.3
& 74.3 & 52.5 & 46.4
& 78.2 & 30.3 & 32.7
& 81.0 & 70.4 & 62.8
& 71.7 & 44.8 & 40.7 \\

CPP
& 48.9 & 25.5 & 13.4
& 87.1 & 71.5 & 66.5
& 80.8 & 73.0 & 58.0
& 80.8 & 59.3 & 48.7
& 82.9 & 72.0 & 64.7
& 76.1 & 60.2 & 50.3 \\

TEMI
& 51.5 & 26.5 & 15.2
& 74.0 & 51.7 & 41.7
& \best{84.8} & \best{83.0} & \best{72.3}
& \best{88.1} & \best{71.6} & \best{64.5}
& 82.0 & 72.2 & 63.7
& 76.1 & 61.0 & 51.5 \\

PRO-DSC
& 51.1 & 25.7 & 15.8
& 87.1 & 71.6 & 65.9
& 77.2 & 65.4 & 47.3
& 86.4 & 68.0 & 56.6
& 85.7 & 81.9 & 73.1
& 77.5 & 62.5 & 51.8 \\

\midrule
\multicolumn{19}{c}{%
  \cellcolor{gray!40}\textbf{Language-assisted Image Clustering}} \\

SIC
& 50.7 & 25.2 & 14.4
& 84.9 & 70.3 & 66.0
& 81.8 & 75.1 & 66.0
& 83.9 & 60.0 & 54.1
& 80.1 & 70.2 & 61.7
& 76.3 & 60.2 & 52.4 \\

TAC
& 52.5 & 26.2 & 15.9
& 91.2 & 76.5 & 75.0
& 75.8 & 67.4 & 52.6
& 84.6 & 60.8 & 54.2
& 87.2 & 77.5 & 72.6
& 78.2 & 61.7 & 54.1 \\

TAC++
& 50.6 & 26.2 & 14.3
& 88.7 & 82.0 & \best{78.0}
& 76.9 & 70.9 & 58.3
& 82.4 & 59.7 & 50.7
& \best{87.3} & \best{83.3} & \best{75.6}
& 77.2 & 64.4 & 55.4 \\

SAC
& 50.6 & 26.0 & 15.0
& 82.8 & 67.5 & 64.5
& 81.0 & 78.3 & 66.1
& 83.0 & 61.2 & 53.5
& 84.7 & 80.9 & 72.8
& 76.4 & 62.8 & 54.4 \\

GradNorm
& 52.1 & 25.8 & 15.5
& 91.1 & 76.5 & 74.7
& 79.8 & 73.8 & 60.9
& 84.6 & 60.2 & 54.3
& 86.4 & 78.4 & 72.2
& 78.8 & 62.9 & 55.5 \\

NTK-SC
& \best{53.6} & \best{28.8} & \best{17.6}
& \best{91.9} & \best{83.2} & 77.9
& 82.0 & 76.8 & 63.8
& 86.6 & 69.8 & 61.3
& 85.8 & 82.0 & 72.9
& \best{80.0} & \best{68.1} & \best{58.7} \\

SEIC
& 46.5 & 20.8 & 11.1
& 76.1 & 58.2 & 51.7
& 78.0 & 73.6 & 60.9
& 74.9 & 50.3 & 37.8
& 72.5 & 54.5 & 44.2
& 69.6 & 51.5 & 41.1 \\

MAGIC
& 45.6 & 18.2 & 9.5
& 61.8 & 35.2 & 27.6
& 78.8 & 69.6 & 58.9
& 65.1 & 27.0 & 19.4
& 82.3 & 68.3 & 62.6
& 66.7 & 43.7 & 35.6 \\

\bottomrule
\end{tabular}%
}
\end{table*}

\begin{table*}[tb]
\centering
\caption{Clustering results on classical datasets with CLIP ViT-L/14 pre-trained on LAION-400M. The best result in each metric column is highlighted in \colorbox[HTML]{FFF2CC}{\textbf{bold}}.}
\label{tab:classical_l14}
\resizebox{\textwidth}{!}{%
\begin{tabular}{l|ccc|ccc|ccc|ccc|ccc|ccc}
\toprule
Dataset & \multicolumn{3}{c|}{STL-10} & \multicolumn{3}{c|}{CIFAR-10} & \multicolumn{3}{c|}{CIFAR-20} & \multicolumn{3}{c|}{ImageNet-10} & \multicolumn{3}{c|}{ImageNet-Dogs} & \multicolumn{3}{c}{Average} \\
Metric & NMI & ACC & ARI & NMI & ACC & ARI & NMI & ACC & ARI & NMI & ACC & ARI & NMI & ACC & ARI & NMI & ACC & ARI \\
\midrule


\multicolumn{19}{c}{\cellcolor{gray!40}\textbf{Classical Image Clustering}} \\
$k$-means & 97.4 & 99.0 & 97.8 & 83.7 & 86.5 & 78.5 & 64.2 & 56.6 & 42.9 & 98.6 & 99.2 & 98.2 & 74.2 & 68.5 & 58.0 & 83.6 & 81.9 & 75.1 \\
SC & 97.0 & 98.8 & 97.3 & 82.5 & 82.9 & 75.8 & 59.2 & 58.5 & 44.2 & 98.7 & 99.4 & 98.7 & 56.8 & 56.0 & 42.3 & 78.8 & 79.1 & 71.7 \\
SSC-OMP & 89.3 & 93.8 & 87.6 & 84.4 & 85.9 & 79.5 & 62.8 & 56.0 & 42.2 & 96.2 & 97.8 & 95.2 & 40.3 & 38.8 & 21.7 & 74.6 & 74.5 & 65.3 \\
EnSC-ORGEN & 86.0 & 90.3 & 81.0 & 78.5 & 84.0 & 77.7 & 60.1 & 53.8 & 40.5 & 91.1 & 94.2 & 88.0 & 52.0 & 57.9 & 36.2 & 73.5 & 76.0 & 64.7 \\
\midrule
\multicolumn{19}{c}{\cellcolor{gray!40}\textbf{Deep Image Clustering}} \\
IDC & \best{98.3} & \best{99.4} & \best{98.6} & 87.4 & 87.9 & 82.9 & 66.4 & 63.6 & 48.5 & 98.5 & 99.6 & 97.9 & 76.2 & 75.6 & 63.6 & 85.3 & 85.2 & 78.3 \\
SCAN & 98.2 & 99.3 & 98.4 & 92.8 & 97.0 & 93.5 & 69.5 & \best{67.1} & 53.7 & \best{99.2} & 99.6 & \best{99.1} & 72.5 & 71.1 & 58.2 & 86.4 & 86.8 & 80.6 \\
CPP & 96.1 & 98.4 & 96.5 & 91.9 & 97.1 & 92.0 & 69.0 & 65.8 & 49.7 & \best{99.2} & 99.6 & \best{99.1} & 78.6 & 77.6 & 66.2 & 87.0 & 87.7 & 80.7 \\
TEMI & 97.8 & 99.1 & 98.1 & 92.8 & 96.8 & 93.1 & 66.3 & 60.4 & 46.9 & \best{99.2} & 99.6 & \best{99.1} & 78.2 & 79.2 & 66.7 & 86.8 & 87.0 & 80.8 \\
PRO-DSC & 96.2 & 98.5 & 96.7 & 92.0 & 96.6 & 92.6 & 70.0 & 66.9 & 53.4 & 97.0 & 98.4 & 96.5 & 78.6 & 77.7 & 67.4 & 86.8 & 87.6 & 81.3 \\
\midrule
\multicolumn{19}{c}{\cellcolor{gray!40}\textbf{Language-assisted Image Clustering}} \\
SIC & 97.9 & 99.2 & 98.2 & 93.0 & 97.0 & 93.5 & 64.3 & 60.2 & 48.7 & \best{99.2} & \best{99.8} & \best{99.1} & 74.1 & 70.8 & 60.2 & 85.7 & 85.4 & 80.0 \\
TAC & 97.9 & 99.2 & 98.3 & 92.3 & 96.7 & 92.8 & 67.8 & 63.2 & 48.6 & 98.7 & 99.4 & 98.7 & 81.3 & 79.9 & 68.4 & 87.6 & 87.7 & 81.3 \\
TAC++ & \best{98.3} & \best{99.4} & \best{98.6} & \best{93.4} & \best{97.2} & \best{93.9} & 69.6 & 66.2 & 53.2 & \best{99.2} & 99.6 & \best{99.1} & 83.7 & 85.6 & 76.0 & 88.8 & 89.6 & 84.2 \\
SAC & 98.1 & 99.3 & 98.4 & 91.3 & 96.3 & 92.0 & 69.0 & 65.4 & 53.0 & 98.7 & 99.4 & 98.7 & 79.1 & 84.8 & 71.8 & 87.3 & 89.0 & 82.8 \\
GradNorm & 97.6 & 99.1 & 97.9 & 91.2 & 96.0 & 91.4 & 68.1 & 65.0 & 51.5 & 97.7 & 99.0 & 98.6 & 83.9 & 82.9 & 75.3 & 87.7 & 88.4 & 83.0 \\
NTK-SC & 97.6 & 99.1 & 98.0 & 93.1 & 97.1 & 93.7 & 68.7 & 61.5 & 49.6 & \best{99.2} & 99.6 & \best{99.1} & 86.0 & 90.9 & 81.8 & 88.9 & 89.6 & 84.4 \\
SEIC & 97.8 & 99.2 & 98.2 & 91.2 & 96.2 & 91.9 & 64.8 & 63.5 & 47.9 & 98.3 & 99.2 & 98.2 & 75.2 & 78.7 & 66.6 & 85.5 & 87.3 & 80.6 \\
MAGIC & \best{98.3} & \best{99.4} & \best{98.6} & 92.7 & 96.9 & 93.3 & \best{70.5} & \best{67.1} & \best{54.9} & 98.9 & 99.6 & 98.7 & \best{88.0} & \best{92.3} & \best{84.2} & \best{89.7} & \best{91.0} & \best{86.0} \\
\bottomrule
\end{tabular}
}
\end{table*}

\begin{table*}[tb]
\centering
\caption{Clustering results on challenging datasets with CLIP ViT-L/14 pre-trained on LAION-400M. The best result in each metric column is highlighted in \textbf{bold}.}
\label{tab:challenging_l14}
\resizebox{0.8\textwidth}{!}{%
\begin{tabular}{l|ccc|ccc|ccc|ccc}
\toprule
Dataset & \multicolumn{3}{c|}{CIFAR-100} & \multicolumn{3}{c|}{DTD} & \multicolumn{3}{c|}{UCF101} & \multicolumn{3}{c}{Average} \\
Metric & NMI & ACC & ARI & NMI & ACC & ARI & NMI & ACC & ARI & NMI & ACC & ARI \\
\midrule


\multicolumn{13}{c}{\cellcolor{gray!40}\textbf{Classical Image Clustering}} \\
$k$-means & 75.3 & 58.4 & 48.9 & 67.9 & 57.4 & 42.5 & 84.7 & 65.2 & 57.6 & 76.0 & 60.3 & 49.7 \\
SC & 74.7 & 59.9 & 51.9 & 69.8 & 61.4 & 45.8 & 85.8 & 69.4 & 62.6 & 76.7 & 63.6 & 53.4 \\
SSC-OMP & 68.7 & 54.6 & 40.7 & 63.0 & 51.4 & 35.1 & 70.0 & 43.5 & 33.5 & 67.2 & 49.8 & 36.5 \\
EnSC-ORGEN & 73.8 & 58.3 & 43.0 & 66.5 & 56.4 & 38.9 & 79.1 & 53.6 & 44.2 & 73.1 & 56.1 & 42.0 \\
\midrule
\multicolumn{13}{c}{\cellcolor{gray!40}\textbf{Deep Image Clustering}} \\
IDC & 74.9 & 63.1 & 47.9 & \best{72.2} & \best{65.4} & \best{49.6} & 85.6 & 73.3 & 60.7 & 77.6 & 67.3 & 52.7 \\
SCAN & 75.1 & 52.4 & 46.4 & 68.3 & 56.7 & 42.6 & 78.7 & 45.2 & 41.9 & 74.0 & 51.5 & 43.7 \\
CPP & 79.1 & 65.9 & 42.6 & 67.3 & 58.3 & 43.1 & 82.9 & 63.8 & 53.8 & 76.4 & 62.7 & 46.5 \\
TEMI & 81.4 & \best{76.1} & \best{63.5} & 68.8 & 57.8 & 42.1 & 85.7 & 73.8 & 65.7 & 78.6 & 69.2 & 57.1 \\
PRO-DSC & \best{82.0} & 72.2 & 61.5 & 69.3 & 59.3 & 43.8 & 85.4 & 72.8 & 63.5 & 78.9 & 68.1 & 56.3 \\
\midrule
\multicolumn{13}{c}{\cellcolor{gray!40}\textbf{Language-assisted Image Clustering}} \\
SIC & 78.0 & 62.9 & 54.2 & 70.0 & 60.4 & 46.5 & 86.1 & 71.2 & 63.7 & 78.0 & 64.8 & 54.8 \\
TAC & 79.8 & 66.5 & 54.7 & 72.1 & 58.2 & 46.4 & 86.6 & 68.4 & 62.5 & 79.5 & 64.4 & 54.5 \\
TAC++ & 79.9 & 74.2 & 62.0 & 71.7 & 63.1 & 47.4 & 85.6 & 75.5 & 66.9 & 79.1 & 70.9 & 58.8 \\
SAC & 79.1 & 73.2 & 61.0 & 70.8 & 60.4 & 46.9 & 84.5 & 72.2 & 63.8 & 78.1 & 68.6 & 57.2 \\
GradNorm & 78.2 & 65.9 & 51.2 & 70.0 & 59.0 & 43.5 & 87.4 & 71.1 & 64.4 & 78.5 & 65.4 & 53.0 \\
NTK-SC & 81.3 & 75.0 & 61.2 & 72.0 & 62.5 & 48.4 & \best{88.6} & \best{79.1} & \best{72.0} & \best{80.7} & \best{72.2} & \best{60.6} \\
SEIC & 79.3 & 74.6 & 61.5 & 68.6 & 60.0 & 43.3 & 84.1 & 72.5 & 63.3 & 77.3 & 69.0 & 56.0 \\
MAGIC & 81.0 & 74.3 & 63.0 & 66.4 & 52.5 & 38.8 & 80.8 & 63.0 & 55.2 & 76.1 & 63.3 & 52.3 \\
\bottomrule
\end{tabular}
}
\end{table*}

\begin{table*}[tb]
\centering
\caption{Clustering results on large-scale datasets with CLIP ViT-L/14 pre-trained on LAION-400M. The best result in each metric column is highlighted in \colorbox[HTML]{FFF2CC}{\textbf{bold}}.}
\label{tab:large_scale_l14}
\resizebox{0.7\textwidth}{!}{%
\begin{tabular}{l|ccc|ccc|ccc}
\toprule
Dataset & \multicolumn{3}{c|}{Places} & \multicolumn{3}{c|}{ImageNet} & \multicolumn{3}{c}{Average} \\
Metric & NMI & ACC & ARI & NMI & ACC & ARI & NMI & ACC & ARI \\
\midrule


\multicolumn{10}{c}{\cellcolor{gray!40}\textbf{Classical Image Clustering}} \\
$k$-means & 59.3 & 32.1 & 19.8 & 79.8 & 50.5 & 40.6 & 69.5 & 41.3 & 30.2 \\
SC & 57.0 & 31.0 & 19.2 & 79.0 & 47.6 & 38.8 & 68.0 & 39.3 & 29.0 \\
SSC-OMP & 53.4 & 26.3 & 13.0 & 70.5 & 38.1 & 17.7 & 62.0 & 32.2 & 15.3 \\
EnSC-ORGEN & \best{64.1} & 34.6 & \best{24.3} & 77.8 & 50.1 & 31.0 & 71.0 & 42.4 & 27.6 \\
\midrule
\multicolumn{10}{c}{\cellcolor{gray!40}\textbf{Deep Image Clustering}} \\
IDC & 59.1 & 32.7 & 19.6 & 80.1 & 54.2 & 41.8 & 69.6 & 43.4 & 30.7 \\
SCAN & 59.5 & 32.8 & 20.0 & 80.3 & 53.7 & 42.4 & 69.9 & 43.2 & 31.2 \\
CPP & 54.4 & 28.2 & 13.1 & 76.4 & 48.9 & 19.0 & 65.4 & 38.5 & 16.1 \\
TEMI & 62.6 & \best{36.7} & 22.9 & \best{83.6} & 63.0 & \best{50.4} & \best{73.1} & 49.9 & \best{36.7} \\
PRO-DSC & 55.0 & 27.3 & 13.7 & 77.0 & 49.4 & 20.8 & 66.0 & 38.3 & 17.2 \\
\midrule
\multicolumn{10}{c}{\cellcolor{gray!40}\textbf{Language-assisted Image Clustering}} \\
SIC & 61.8 & 31.9 & 21.6 & 83.0 & 60.0 & 48.9 & 72.4 & 45.9 & 35.2 \\
TAC & 60.5 & 34.0 & 20.6 & 82.0 & 55.3 & 44.8 & 71.2 & 44.6 & 32.7 \\
TAC++ & 60.6 & 35.7 & 21.4 & 82.7 & 63.2 & 49.7 & 71.7 & 49.4 & 35.5 \\
SAC & 59.5 & 35.8 & 21.9 & 82.7 & 62.9 & 50.0 & 71.1 & 49.4 & 36.0 \\
GradNorm & 60.6 & 33.9 & 20.5 & 81.9 & 55.3 & 44.5 & 71.2 & 44.6 & 32.5 \\
NTK-SC & 62.6 & 35.9 & 23.6 & 81.7 & \best{64.7} & 49.3 & 72.2 & \best{50.3} & 36.4 \\
SEIC & 60.1 & 33.3 & 19.5 & 80.8 & 53.6 & 39.6 & 70.5 & 43.4 & 29.5 \\
MAGIC & 60.7 & 35.8 & 21.7 & 81.7 & 54.0 & 43.3 & 71.2 & 44.9 & 32.5 \\
\bottomrule
\end{tabular}
}
\end{table*}

\begin{table*}[tb]
\centering
\caption{Clustering results on fine-grained datasets with CLIP ViT-L/14 pre-trained on LAION-400M. The best result in each metric column is highlighted in \colorbox[HTML]{FFF2CC}{\textbf{bold}}.}
\label{tab:fine_grained_l14}
\resizebox{\textwidth}{!}{%
\begin{tabular}{l|ccc|ccc|ccc|ccc|ccc|ccc}
\toprule
Dataset & \multicolumn{3}{c|}{Aircraft} & \multicolumn{3}{c|}{Flowers} & \multicolumn{3}{c|}{Food} & \multicolumn{3}{c|}{Cars} & \multicolumn{3}{c|}{Pets} & \multicolumn{3}{c}{Average} \\
Metric & NMI & ACC & ARI & NMI & ACC & ARI & NMI & ACC & ARI & NMI & ACC & ARI & NMI & ACC & ARI & NMI & ACC & ARI \\
\midrule


\multicolumn{19}{c}{\cellcolor{gray!40}\textbf{Classical Image Clustering}} \\
$k$-means & 56.5 & 30.6 & 19.8 & 93.2 & 77.6 & 77.6 & 85.3 & 78.3 & 70.3 & 89.0 & 67.5 & 64.0 & 85.9 & 74.9 & 70.6 & 82.0 & 65.8 & 60.4 \\
SC & 52.6 & 28.3 & 19.7 & 92.7 & 83.9 & 83.3 & 82.0 & 70.6 & 64.4 & 86.8 & 62.2 & 60.0 & 71.2 & 57.0 & 52.7 & 77.1 & 60.4 & 56.0 \\
SSC-OMP & 47.9 & 23.8 & 11.9 & 62.0 & 41.8 & 34.8 & 68.0 & 54.5 & 42.1 & 75.6 & 53.6 & 41.2 & 54.9 & 46.1 & 31.8 & 61.7 & 44.0 & 32.4 \\
EnSC-ORGEN & 52.9 & 27.8 & 17.0 & 91.0 & 77.6 & 72.9 & 82.5 & 76.3 & 63.8 & 85.6 & 66.7 & 55.8 & 71.3 & 62.6 & 48.5 & 76.7 & 62.2 & 51.6 \\
\midrule
\multicolumn{19}{c}{\cellcolor{gray!40}\textbf{Deep Image Clustering}} \\
IDC & \best{60.4} & 35.9 & 24.1 & 83.0 & 67.4 & 63.2 & 82.2 & 75.5 & 63.9 & 87.7 & 70.3 & 61.0 & 86.4 & 80.7 & 71.4 & 79.9 & 66.0 & 56.7 \\
SCAN & 48.8 & 15.5 & 9.2 & 85.8 & 62.4 & 62.5 & 81.7 & 60.3 & 57.3 & 83.6 & 33.1 & 39.5 & 87.7 & 82.8 & 76.3 & 77.5 & 50.8 & 49.0 \\
CPP & 55.5 & 32.2 & 20.2 & 90.3 & 74.3 & 70.5 & 86.1 & 77.0 & 65.2 & 86.7 & 67.7 & 60.6 & 85.4 & 77.3 & 68.8 & 80.8 & 65.7 & 57.0 \\
TEMI & 57.8 & 32.1 & 23.7 & 79.5 & 59.1 & 52.2 & \best{88.8} & \best{88.0} & \best{79.7} & 92.4 & \best{81.1} & 75.6 & 88.0 & 80.4 & 75.3 & 81.3 & 68.2 & 61.3 \\
PRO-DSC & 55.4 & 33.0 & 22.9 & 91.4 & 76.7 & 73.6 & 87.0 & 80.6 & 72.8 & \best{93.0} & 79.8 & \best{75.7} & 88.1 & 83.9 & 77.3 & 83.0 & 70.8 & 64.5 \\
\midrule
\multicolumn{19}{c}{\cellcolor{gray!40}\textbf{Language-assisted Image Clustering}} \\
SIC & 57.6 & 32.6 & 22.6 & 87.2 & 73.5 & 68.7 & 85.8 & 80.4 & 71.9 & 89.6 & 68.7 & 65.9 & 85.5 & 79.2 & 71.8 & 81.2 & 66.9 & 60.2 \\
TAC & 58.5 & 33.5 & 23.5 & 93.0 & 77.1 & 77.6 & 85.8 & 79.3 & 70.3 & 89.5 & 67.2 & 64.6 & 89.4 & 81.2 & 74.7 & 83.2 & 67.7 & 62.2 \\
TAC++ & 57.4 & 34.6 & 23.0 & 93.8 & \best{89.7} & \best{86.1} & 86.0 & 83.2 & 74.2 & 87.5 & 70.1 & 63.1 & 89.2 & 84.3 & \best{78.4} & 82.8 & 72.4 & 65.0 \\
SAC & 56.9 & 32.2 & 23.2 & 92.1 & 81.8 & 80.5 & 82.5 & 79.5 & 67.6 & 88.3 & 72.8 & 66.5 & 86.6 & 82.8 & 74.9 & 81.3 & 69.8 & 62.5 \\
GradNorm & 59.2 & 34.1 & 24.5 & 93.7 & 80.2 & 79.9 & 86.2 & 80.2 & 70.8 & 89.7 & 68.1 & 64.1 & \best{89.5} & 81.4 & 76.4 & 83.7 & 68.8 & 63.1 \\
NTK-SC & 59.3 & \best{36.4} & \best{26.2} & \best{94.3} & 87.3 & 83.7 & 86.6 & 83.4 & 73.1 & 91.5 & 80.0 & 74.1 & \best{89.5} & \best{85.2} & 77.8 & \best{84.2} & \best{74.5} & \best{67.0} \\
SEIC & 50.5 & 25.3 & 14.0 & 80.4 & 64.7 & 59.7 & 84.6 & 82.9 & 72.5 & 77.5 & 53.6 & 42.6 & 77.4 & 60.5 & 49.3 & 74.1 & 57.4 & 47.6 \\
MAGIC & 50.9 & 22.3 & 13.2 & 66.9 & 39.0 & 34.5 & 85.7 & 80.9 & 71.9 & 71.2 & 33.8 & 28.4 & 87.5 & 78.8 & 73.1 & 72.4 & 51.0 & 44.2 \\
\bottomrule
\end{tabular}
}
\end{table*}

\begin{table*}[tb]
\centering
\caption{Clustering results on classical datasets with CLIP ViT-B/32 pre-trained on LAION-2B. The best result in each metric column is highlighted in \colorbox[HTML]{FFF2CC}{\textbf{bold}}.}
\label{tab:laion2b_b32_classical}
\resizebox{\textwidth}{!}{%
\begin{tabular}{l|ccc|ccc|ccc|ccc|ccc|ccc}
\toprule
Dataset & \multicolumn{3}{c|}{STL-10} & \multicolumn{3}{c|}{CIFAR-10} & \multicolumn{3}{c|}{CIFAR-20} & \multicolumn{3}{c|}{ImageNet-10} & \multicolumn{3}{c|}{ImageNet-Dogs} & \multicolumn{3}{c}{Average} \\
Metric & NMI & ACC & ARI & NMI & ACC & ARI & NMI & ACC & ARI & NMI & ACC & ARI & NMI & ACC & ARI & NMI & ACC & ARI \\
\midrule


\multicolumn{19}{c}{\cellcolor{gray!40}\textbf{Classical Image Clustering}} \\
$k$-means & 95.1 & 98.0 & 95.6 & 80.9 & 86.7 & 75.0 & 63.4 & 55.2 & 42.3 & 95.6 & 96.2 & 92.6 & 67.1 & 66.9 & 51.8 & 80.4 & 80.6 & 71.5 \\
SC & 94.0 & 97.5 & 94.5 & 79.3 & 82.7 & 74.6 & 56.5 & 55.2 & 42.3 & 95.6 & 97.2 & 94.1 & 50.4 & 49.6 & 34.0 & 75.1 & 76.4 & 67.9 \\
SSC-OMP & 79.5 & 86.1 & 74.5 & 77.8 & 83.0 & 73.2 & 59.5 & 52.6 & 39.8 & 90.7 & 94.4 & 88.4 & 39.3 & 47.6 & 24.2 & 69.4 & 72.7 & 60.0 \\
EnSC & 82.6 & 84.0 & 76.0 & 78.9 & 83.0 & 74.0 & 62.4 & 58.9 & 42.4 & 83.8 & 88.6 & 78.8 & 51.0 & 52.3 & 34.2 & 71.7 & 73.3 & 61.1 \\
\midrule
\multicolumn{19}{c}{\cellcolor{gray!40}\textbf{Deep Image Clustering}} \\
IDC & 95.4 & 98.0 & 95.7 & 80.4 & 80.0 & 70.9 & 61.8 & 60.3 & 44.1 & 96.3 & 98.0 & 95.6 & 69.2 & 69.9 & 52.9 & 80.6 & 81.2 & 71.8 \\
SCAN & 94.8 & 97.8 & 95.3 & 87.4 & 93.0 & 86.0 & 64.7 & 59.2 & 46.7 & 97.7 & 98.8 & 97.4 & 68.6 & 68.5 & 54.7 & 82.6 & 83.5 & 76.0 \\
CPP & 93.2 & 97.1 & 93.7 & 87.7 & 93.9 & 87.2 & 68.3 & 62.5 & 46.9 & 97.4 & 98.4 & 96.6 & 71.3 & 68.4 & 53.2 & 83.6 & 84.0 & 75.5 \\
TEMI & 95.7 & 98.3 & 96.2 & 89.4 & 95.0 & 89.2 & 67.0 & 65.1 & 49.8 & 97.5 & 98.6 & 97.0 & 70.1 & 68.8 & 55.8 & 84.0 & 85.2 & 77.6 \\
PRO-DSC & 95.6 & 98.3 & 96.2 & 88.4 & 94.3 & 88.0 & 67.7 & 63.4 & 50.0 & 96.0 & 97.8 & 95.2 & 67.7 & 67.5 & 50.8 & 83.1 & 84.3 & 76.0 \\
\midrule
\multicolumn{19}{c}{\cellcolor{gray!40}\textbf{Language-assisted Image Clustering}} \\
SIC & 95.3 & 98.1 & 95.9 & \best{90.8} & \best{96.0} & \best{91.4} & 63.2 & 60.5 & 46.5 & 97.9 & 98.8 & 97.4 & 73.0 & 73.6 & 61.7 & 84.0 & 85.4 & 78.6 \\
TAC & 95.0 & 97.9 & 95.4 & 88.9 & 95.0 & 89.4 & 68.5 & 62.2 & 49.6 & 97.7 & 98.6 & 97.0 & 79.7 & 79.3 & 69.0 & 86.0 & 86.6 & 80.1 \\
TAC$^*$ & \best{95.9} & \best{98.4} & 96.4 & 89.3 & 95.2 & 89.7 & 64.6 & 63.7 & 48.6 & 98.0 & 99.0 & 97.8 & 79.0 & 84.5 & 70.9 & 85.3 & 88.2 & 80.7 \\
SAC & 95.6 & 98.2 & 96.1 & 89.4 & 95.2 & 89.9 & \best{68.6} & \best{68.0} & \best{54.0} & 98.2 & 99.0 & 97.8 & 75.5 & 78.8 & 66.0 & 85.5 & 87.8 & 80.8 \\
GradNorm & 94.4 & 97.7 & 94.9 & 88.5 & 94.8 & 89.0 & 67.9 & 62.2 & 50.1 & 97.5 & 98.0 & 95.9 & 79.8 & 80.0 & 69.9 & 85.6 & 86.5 & 80.0 \\
NTK-SC & 95.1 & 97.9 & 95.5 & 89.0 & 94.8 & 89.1 & 61.8 & 56.1 & 43.5 & \best{98.4} & \best{99.2} & \best{98.2} & 80.3 & 86.7 & 74.1 & 84.9 & 86.9 & 80.1 \\
SEIC & 95.6 & 98.2 & 96.2 & 89.4 & 95.3 & 90.0 & 64.0 & 62.4 & 48.7 & 97.4 & 98.8 & 97.3 & 75.6 & 78.3 & 65.3 & 84.4 & 86.6 & 79.5 \\
MAGIC & \best{95.9} & \best{98.4} & \best{96.5} & 90.7 & 95.9 & 91.3 & 67.7 & 64.1 & 51.1 & 98.0 & 99.0 & 97.8 & \best{82.9} & \best{88.1} & \best{76.9} & \best{87.0} & \best{89.1} & \best{82.7} \\
\bottomrule
\end{tabular}
}
\end{table*}

\begin{table*}[tb]
\centering
\caption{Clustering results on challenging datasets with CLIP ViT-B/32 pre-trained on LAION-2B. The best result in each metric column is highlighted in \colorbox[HTML]{FFF2CC}{\textbf{bold}}.}
\label{tab:laion2b_b32_challenging}
\resizebox{0.8\textwidth}{!}{%
\begin{tabular}{l|ccc|ccc|ccc|ccc}
\toprule
Dataset & \multicolumn{3}{c|}{CIFAR-100} & \multicolumn{3}{c|}{DTD} & \multicolumn{3}{c|}{UCF101} & \multicolumn{3}{c}{Average} \\
Metric & NMI & ACC & ARI & NMI & ACC & ARI & NMI & ACC & ARI & NMI & ACC & ARI \\
\midrule


\multicolumn{13}{c}{\cellcolor{gray!40}\textbf{Classical Image Clustering}} \\
$k$-means & 73.1 & 59.4 & 45.7 & 66.7 & 57.1 & 40.7 & 80.9 & 59.8 & 51.8 & 73.5 & 58.8 & 46.0 \\
SC & 72.2 & 59.6 & 49.5 & 66.4 & 55.1 & 42.8 & 81.8 & 63.2 & 55.6 & 73.4 & 59.3 & 49.3 \\
SSC-OMP & 63.9 & 49.3 & 34.8 & 59.4 & 48.1 & 31.6 & 63.5 & 37.7 & 25.1 & 62.3 & 45.0 & 30.5 \\
EnSC & 71.9 & 58.4 & 43.4 & 64.8 & 56.0 & 36.9 & 77.2 & 51.9 & 41.0 & 71.3 & 55.4 & 40.4 \\
\midrule
\multicolumn{13}{c}{\cellcolor{gray!40}\textbf{Deep Image Clustering}} \\
IDC & 72.4 & 61.8 & 45.7 & 69.1 & \best{61.4} & 44.9 & 79.6 & 64.3 & 53.1 & 73.7 & 62.5 & 47.9 \\
SCAN & 71.2 & 47.6 & 39.9 & 66.1 & 54.2 & 40.5 & 74.5 & 40.9 & 35.9 & 70.6 & 47.5 & 38.8 \\
CPP & 76.2 & 63.4 & 39.1 & 66.0 & 58.0 & 40.5 & 80.1 & 61.3 & 50.5 & 74.1 & 60.9 & 43.4 \\
TEMI & 78.3 & 71.0 & 58.5 & 68.8 & 59.3 & 43.7 & 80.9 & 66.7 & 57.2 & 76.0 & 65.7 & 53.1 \\
PRO-DSC & 78.5 & 67.7 & 56.3 & 67.4 & 58.6 & 40.7 & 82.8 & 66.0 & 56.7 & 76.2 & 64.1 & 51.3 \\
\midrule
\multicolumn{13}{c}{\cellcolor{gray!40}\textbf{Language-assisted Image Clustering}} \\
SIC & 76.4 & 63.7 & 52.6 & 68.5 & 58.7 & 43.3 & 81.8 & 66.0 & 56.7 & 75.6 & 62.8 & 50.9 \\
TAC & 76.7 & 64.4 & 53.3 & 69.0 & 59.8 & 42.6 & 81.9 & 61.8 & 53.1 & 75.9 & 62.0 & 49.7 \\
TAC$^*$ & 77.0 & 71.1 & 57.4 & 69.5 & 59.7 & 44.2 & 81.1 & 66.7 & 56.8 & 75.9 & 65.9 & 52.8 \\
SAC & 78.0 & \best{73.1} & \best{59.5} & 69.6 & 59.7 & 45.6 & 81.8 & 69.3 & 59.0 & 76.5 & 67.4 & \best{54.7} \\
GradNorm & 75.8 & 63.0 & 50.5 & 68.5 & 57.8 & 40.9 & 82.4 & 63.0 & 52.9 & 75.6 & 61.3 & 48.1 \\
NTK-SC & \best{79.6} & 69.8 & 55.0 & \best{70.1} & 61.3 & \best{45.9} & \best{84.3} & \best{72.0} & \best{62.5} & \best{78.0} & \best{67.7} & 54.5 \\
SEIC & 76.0 & 69.6 & 55.2 & 65.2 & 55.6 & 38.3 & 79.3 & 64.7 & 54.1 & 73.5 & 63.3 & 49.2 \\
MAGIC & 76.9 & 65.8 & 54.2 & 62.6 & 48.7 & 34.6 & 74.1 & 51.0 & 41.6 & 71.2 & 55.2 & 43.5 \\
\bottomrule
\end{tabular}
}
\end{table*}

\begin{table*}[tb]
\centering
\caption{Clustering results on large-scale datasets with CLIP ViT-B/32 pre-trained on LAION-2B. The best result in each metric column is highlighted in \colorbox[HTML]{FFF2CC}{\textbf{bold}}.}
\label{tab:laion2b_b32_large_scale}
\resizebox{0.7\textwidth}{!}{%
\begin{tabular}{l|ccc|ccc|ccc}
\toprule
Dataset & \multicolumn{3}{c|}{Places-365} & \multicolumn{3}{c|}{ImageNet-1K} & \multicolumn{3}{c}{Average} \\
Metric & NMI & ACC & ARI & NMI & ACC & ARI & NMI & ACC & ARI \\
\midrule


\multicolumn{10}{c}{\cellcolor{gray!40}\textbf{Classical Image Clustering}} \\
$k$-means & 58.4 & 31.2 & 18.3 & 75.1 & 43.3 & 31.7 & 66.7 & 37.2 & 25.0 \\
SC & 56.6 & 29.9 & 18.1 & 74.1 & 40.4 & 30.5 & 65.4 & 35.1 & 24.3 \\
SSC-OMP & 52.3 & 25.0 & 12.0 & 65.9 & 31.1 & 12.7 & 59.1 & 28.1 & 12.4 \\
EnSC & 57.0 & 30.2 & 15.2 & 74.9 & 46.1 & 26.9 & 65.9 & 38.1 & 21.0 \\
\midrule
\multicolumn{10}{c}{\cellcolor{gray!40}\textbf{Deep Image Clustering}} \\
IDC & 58.0 & 31.4 & 17.7 & 74.8 & 45.1 & 30.8 & 66.4 & 38.3 & 24.2 \\
SCAN & 59.1 & 32.6 & 19.4 & 76.7 & 47.5 & 35.2 & 67.9 & 40.0 & 27.3 \\
CPP & 54.4 & 28.2 & 12.8 & 72.0 & 42.1 & 16.4 & 63.2 & 35.2 & 14.6 \\
TEMI & 60.2 & 34.1 & 20.3 & \best{79.5} & 55.3 & 41.3 & 69.8 & 44.7 & 30.8 \\
PRO-DSC & 54.4 & 25.6 & 13.0 & 73.2 & 43.6 & 21.7 & 63.8 & 34.6 & 17.3 \\
\midrule
\multicolumn{10}{c}{\cellcolor{gray!40}\textbf{Language-assisted Image Clustering}} \\
SIC & 59.0 & 30.7 & 18.8 & 79.2 & 52.9 & 40.1 & 69.1 & 41.8 & 29.4 \\
TAC & 59.2 & 32.2 & 19.1 & 78.8 & 51.2 & 38.7 & 69.0 & 41.7 & 28.9 \\
TAC$^*$ & 59.7 & 34.1 & 20.1 & 78.9 & \best{56.8} & 41.7 & 69.3 & 45.4 & 30.9 \\
SAC & 59.5 & \best{35.3} & 21.4 & 79.2 & 56.7 & \best{42.5} & 69.3 & \best{46.0} & \best{32.0} \\
GradNorm & 59.9 & 33.7 & 20.0 & 79.4 & 52.1 & 40.1 & 69.7 & 42.9 & 30.0 \\
NTK-SC & \best{62.4} & 34.7 & \best{21.9} & 79.4 & 54.9 & 41.0 & \best{70.9} & 44.8 & 31.5 \\
SEIC & 59.7 & 32.1 & 18.3 & 76.9 & 47.8 & 32.2 & 68.3 & 39.9 & 25.3 \\
MAGIC & 60.4 & 34.2 & 20.6 & 76.6 & 44.4 & 33.0 & 68.5 & 39.3 & 26.8 \\
\bottomrule
\end{tabular}
}
\end{table*}

\begin{table*}[tb]
\centering
\caption{Clustering results on fine-grained datasets with CLIP ViT-B/32 pre-trained on LAION-2B. The best result in each metric column is highlighted in \colorbox[HTML]{FFF2CC}{\textbf{bold}}.}
\label{tab:laion2b_b32_fine_grained}
\resizebox{\textwidth}{!}{%
\begin{tabular}{l|ccc|ccc|ccc|ccc|ccc|ccc}
\toprule
Dataset & \multicolumn{3}{c|}{Aircraft} & \multicolumn{3}{c|}{Flowers} & \multicolumn{3}{c|}{Food} & \multicolumn{3}{c|}{Cars} & \multicolumn{3}{c|}{Pets} & \multicolumn{3}{c}{Average} \\
Metric & NMI & ACC & ARI & NMI & ACC & ARI & NMI & ACC & ARI & NMI & ACC & ARI & NMI & ACC & ARI & NMI & ACC & ARI \\
\midrule


\multicolumn{19}{c}{\cellcolor{gray!40}\textbf{Classical Image Clustering}} \\
$k$-means & 55.3 & 27.2 & 17.9 & 89.2 & 72.2 & 70.2 & 75.5 & 68.5 & 54.5 & 81.0 & 55.2 & 47.8 & 81.5 & 70.4 & 63.6 & 76.5 & 58.7 & 50.8 \\
SC & 49.6 & 24.8 & 15.7 & 85.6 & 74.5 & \best{74.0} & 69.0 & 58.8 & 47.8 & 77.4 & 49.4 & 44.1 & 74.4 & 68.8 & 63.5 & 71.2 & 55.3 & 49.0 \\
SSC-OMP & 44.4 & 20.3 & 8.7 & 54.5 & 34.6 & 23.1 & 50.9 & 36.4 & 24.3 & 66.6 & 41.5 & 27.3 & 50.6 & 37.5 & 23.8 & 53.4 & 34.1 & 21.4 \\
EnSC & 50.4 & 25.7 & 13.3 & 86.5 & 76.9 & 66.2 & 73.7 & 65.9 & 51.4 & 78.8 & 59.1 & 45.8 & 67.1 & 57.9 & 44.8 & 71.3 & 57.1 & 44.3 \\
\midrule
\multicolumn{19}{c}{\cellcolor{gray!40}\textbf{Deep Image Clustering}} \\
IDC & \best{55.6} & \best{30.7} & 18.4 & 64.1 & 35.2 & 28.0 & 73.5 & 69.6 & 52.6 & 80.6 & 58.8 & 47.6 & 81.6 & 71.5 & 62.2 & 71.1 & 53.2 & 41.8 \\
SCAN & 47.5 & 16.0 & 9.2 & 78.8 & 53.6 & 49.3 & 68.7 & 46.6 & 38.8 & 74.8 & 27.4 & 28.8 & 82.5 & 73.0 & 66.8 & 70.4 & 43.3 & 38.6 \\
CPP & 50.7 & 26.4 & 14.7 & 85.8 & 71.4 & 66.0 & 73.3 & 63.5 & 37.7 & 78.1 & 57.1 & 44.0 & 73.9 & 67.0 & 52.8 & 72.3 & 57.1 & 43.0 \\
TEMI & 54.7 & 29.4 & 18.9 & 72.8 & 52.5 & 45.2 & \best{80.3} & \best{78.2} & \best{65.6} & \best{85.4} & \best{66.4} & \best{58.2} & 82.2 & 72.3 & 63.9 & 75.1 & 59.8 & 50.3 \\
PRO-DSC & 52.3 & 27.6 & 16.3 & 85.6 & 68.3 & 63.1 & 75.6 & 66.2 & 53.0 & 84.1 & 62.6 & 54.2 & 82.9 & 78.1 & 68.2 & 76.1 & 60.6 & 51.0 \\
\midrule
\multicolumn{19}{c}{\cellcolor{gray!40}\textbf{Language-assisted Image Clustering}} \\
SIC & 54.5 & 28.0 & 17.5 & 81.6 & 66.7 & 59.3 & 76.2 & 69.7 & 56.8 & 80.4 & 56.1 & 47.8 & 82.3 & 70.8 & 65.2 & 75.0 & 58.3 & 49.3 \\
TAC & 55.0 & 28.1 & 17.3 & 90.0 & 74.4 & 72.0 & 76.3 & 70.2 & 56.1 & 81.3 & 55.9 & 47.0 & 85.6 & 76.5 & 69.3 & 77.7 & 61.0 & 52.4 \\
TAC$^*$ & 53.4 & 28.7 & 16.4 & 86.0 & 77.2 & 72.6 & 75.6 & 70.2 & 56.8 & 80.1 & 56.2 & 46.5 & 85.6 & 79.4 & 72.0 & 76.2 & 62.4 & 52.9 \\
SAC & 53.3 & 28.0 & 17.5 & 81.8 & 64.0 & 61.1 & 76.7 & 72.6 & 59.7 & 82.8 & 61.9 & 53.8 & 84.8 & \best{81.4} & 72.1 & 75.9 & 61.6 & 52.8 \\
GradNorm & 55.3 & 28.4 & 17.6 & \best{90.6} & 76.9 & 73.7 & 75.7 & 69.2 & 54.2 & 81.8 & 56.7 & 49.3 & \best{86.7} & 78.1 & 71.4 & 78.0 & 61.8 & 53.2 \\
NTK-SC & \best{55.6} & 29.4 & \best{19.0} & 88.9 & \best{81.2} & 71.5 & 78.5 & 73.4 & 58.6 & 84.2 & 66.2 & 56.7 & 86.2 & \best{81.4} & \best{72.9} & \best{78.7} & \best{66.3} & \best{55.7} \\
SEIC & 47.4 & 21.6 & 11.0 & 76.3 & 60.9 & 53.0 & 73.3 & 69.3 & 54.7 & 72.7 & 46.2 & 33.8 & 76.1 & 56.6 & 46.2 & 69.1 & 50.9 & 39.7 \\
MAGIC & 48.3 & 19.4 & 9.7 & 56.2 & 29.0 & 20.1 & 74.4 & 65.6 & 53.0 & 63.9 & 25.4 & 18.3 & 81.5 & 68.9 & 61.2 & 64.8 & 41.7 & 32.4 \\
\bottomrule
\end{tabular}
}
\end{table*}

\begin{table*}[tb]
\centering
\caption{Clustering results on classical datasets with SigLIP ViT-B/16. The best result in each metric column is highlighted in \colorbox[HTML]{FFF2CC}{\textbf{bold}}.}
\label{tab:siglip_b16_classical}
\resizebox{\textwidth}{!}{%
\begin{tabular}{l|ccc|ccc|ccc|ccc|ccc|ccc}
\toprule
Dataset
 & \multicolumn{3}{c|}{STL-10}
 & \multicolumn{3}{c|}{CIFAR-10}
 & \multicolumn{3}{c|}{CIFAR-20}
 & \multicolumn{3}{c|}{ImageNet-10}
 & \multicolumn{3}{c|}{ImageNet-Dogs}
 & \multicolumn{3}{c}{Average} \\
Metric & NMI & ACC & ARI
 & NMI & ACC & ARI
 & NMI & ACC & ARI
 & NMI & ACC & ARI
 & NMI & ACC & ARI
 & NMI & ACC & ARI \\
\midrule

\multicolumn{19}{c}{\cellcolor{gray!40}\textbf{Classical Image Clustering}} \\
$k$-means & 92.2 & 94.2 & 88.6 & 77.7 & 78.8 & 68.8 & 58.1 & 51.1 & 36.8 & 98.6 & 99.0 & 98.2 & 77.0 & 74.5 & 63.6 & 80.7 & 79.5 & 71.2 \\
SC & 88.1 & 91.2 & 84.7 & 74.2 & 78.1 & 67.9 & 51.9 & 49.5 & 36.6 & 98.2 & 99.0 & 97.8 & 61.7 & 63.1 & 48.5 & 74.8 & 76.2 & 67.1 \\
SSC-OMP & 81.7 & 86.2 & 76.6 & 74.2 & 84.5 & 70.4 & 55.8 & 50.8 & 37.3 & 95.8 & 97.8 & 95.2 & 44.3 & 50.3 & 31.6 & 70.4 & 73.9 & 62.2 \\
EnSC-ORGEN & 87.9 & 85.9 & 80.5 & 83.0 & 90.8 & 80.9 & 59.9 & 49.5 & 38.2 & 98.4 & 99.2 & 98.2 & 66.9 & 73.2 & 55.5 & 79.2 & 79.7 & 70.7 \\
\midrule
\multicolumn{19}{c}{\cellcolor{gray!40}\textbf{Deep Image Clustering}} \\
IDC & \best{97.3} & \best{99.0} & \best{97.8} & 78.6 & 81.7 & 71.5 & 60.8 & 58.8 & 43.1 & 96.9 & 98.0 & 95.7 & 78.2 & 79.6 & 66.0 & 82.4 & 83.4 & 74.8 \\
SCAN & 97.0 & 98.9 & 97.5 & 78.4 & 82.8 & 72.1 & 60.9 & 58.8 & 45.2 & 98.5 & 99.2 & 98.2 & 82.2 & 80.5 & 70.3 & 83.4 & 84.0 & 76.7 \\
CPP & 88.0 & 95.0 & 87.6 & 85.3 & 92.8 & 84.7 & \best{65.6} & 59.4 & 42.6 & \best{99.3} & \best{99.6} & \best{99.1} & 76.2 & 76.0 & 50.6 & 82.9 & 84.6 & 72.9 \\
TEMI & 96.8 & 98.8 & 97.2 & 86.1 & 93.1 & 85.5 & 63.3 & 62.3 & 46.7 & 99.2 & \best{99.6} & \best{99.1} & 83.5 & 83.2 & 73.6 & 85.8 & 87.4 & 80.4 \\
PRO-DSC & 96.0 & 98.2 & 96.1 & 85.4 & 92.5 & 84.0 & 61.5 & 54.7 & 41.6 & 98.1 & 99.0 & 97.8 & 78.7 & 78.7 & 66.5 & 83.9 & 84.6 & 77.2 \\
\midrule
\multicolumn{19}{c}{\cellcolor{gray!40}\textbf{Language-assisted Image Clustering}} \\
SIC & 96.6 & 98.7 & 97.2 & \best{88.2} & 94.5 & 88.3 & 61.5 & 60.9 & 47.4 & 98.6 & 99.2 & 98.2 & 84.2 & 84.1 & 76.6 & 85.8 & 87.5 & 81.6 \\
TAC & 95.9 & 98.4 & 96.4 & 85.0 & 92.5 & 84.2 & 64.0 & 61.1 & 43.6 & 98.8 & 99.4 & 98.7 & 83.9 & 85.6 & 76.6 & 85.5 & 87.4 & 79.9 \\
TAC++ & 96.5 & 98.6 & 97.0 & 88.0 & \best{94.6} & \best{88.4} & 65.4 & \best{66.4} & \best{50.2} & 98.8 & \best{99.6} & 98.7 & 88.2 & 92.1 & 84.1 & \best{87.4} & \best{90.3} & \best{83.7} \\
SAC & 96.9 & 98.8 & 97.4 & 85.8 & 93.5 & 86.2 & 59.2 & 57.7 & 43.0 & 98.8 & 99.4 & 98.7 & 89.1 & \best{93.1} & 85.8 & 85.9 & 88.5 & 82.2 \\
GradNorm & 96.5 & 98.6 & 97.0 & 85.0 & 92.5 & 84.2 & 64.9 & 58.5 & 42.7 & 98.3 & 99.0 & 97.8 & 85.1 & 85.9 & 77.5 & 86.0 & 86.9 & 79.8 \\
NTK-SC & 96.6 & 98.7 & 97.1 & 86.2 & 93.5 & 86.2 & 63.2 & 52.0 & 37.4 & 98.4 & 99.2 & 98.2 & 89.1 & 92.9 & 85.7 & 86.7 & 87.3 & 80.9 \\
SEIC & 96.8 & 98.8 & 97.4 & 86.6 & 93.9 & 87.1 & 62.4 & 60.9 & 46.9 & 99.2 & \best{99.6} & \best{99.1} & 81.6 & 82.9 & 73.9 & 85.3 & 87.2 & 80.9 \\
MAGIC & 97.0 & 98.9 & 97.5 & 86.2 & 93.6 & 86.5 & 63.0 & 60.6 & 47.8 & 99.2 & \best{99.6} & \best{99.1} & \best{89.5} & \best{93.1} & \best{86.0} & 87.0 & 89.2 & 83.4 \\
\bottomrule
\end{tabular}
}
\end{table*}

\begin{table*}[tb]
\centering
\caption{Clustering results on challenging datasets with SigLIP ViT-B/16. The best result in each metric column is highlighted in \colorbox[HTML]{FFF2CC}{\textbf{bold}}.}
\label{tab:siglip_b16_challenging}
\resizebox{0.8\textwidth}{!}{%
\begin{tabular}{l|ccc|ccc|ccc|ccc}
\toprule
Dataset
 & \multicolumn{3}{c|}{CIFAR-100}
 & \multicolumn{3}{c|}{DTD}
 & \multicolumn{3}{c|}{UCF101}
 & \multicolumn{3}{c}{Average} \\
Metric & NMI & ACC & ARI
 & NMI & ACC & ARI
 & NMI & ACC & ARI
 & NMI & ACC & ARI \\
\midrule


\multicolumn{13}{c}{\cellcolor{gray!40}\textbf{Classical Image Clustering}} \\
$k$-means & 68.3 & 51.5 & 36.7 & 67.8 & 54.7 & 40.6 & 82.8 & 63.4 & 54.4 & 73.0 & 56.5 & 43.9 \\
SC & 70.4 & 57.6 & 46.4 & 68.5 & 57.2 & 45.2 & 85.2 & 68.8 & 62.2 & 74.7 & 61.2 & 51.3 \\
SSC-OMP & 63.3 & 46.0 & 34.0 & 61.1 & 45.9 & 32.3 & 69.4 & 43.6 & 32.3 & 64.6 & 45.1 & 32.9 \\
EnSC-ORGEN & 71.4 & 57.8 & 43.4 & 66.5 & 55.6 & 39.1 & 81.7 & 58.1 & 49.0 & 73.2 & 57.2 & 43.9 \\
\midrule
\multicolumn{13}{c}{\cellcolor{gray!40}\textbf{Deep Image Clustering}} \\
IDC & 68.3 & 56.1 & 37.3 & 71.4 & \best{63.9} & \best{47.2} & 82.7 & 70.0 & 56.2 & 74.1 & 63.3 & 46.9 \\
SCAN & 61.1 & 24.2 & 19.5 & 67.2 & 52.9 & 39.6 & 71.6 & 33.0 & 27.6 & 66.6 & 36.7 & 28.9 \\
CPP & 74.1 & 62.0 & 27.6 & 63.4 & 54.9 & 37.6 & 79.0 & 60.9 & 31.8 & 72.2 & 59.3 & 32.3 \\
TEMI & 74.7 & 65.1 & \best{52.5} & 70.2 & 58.9 & 45.6 & 83.7 & 71.4 & 62.6 & 76.2 & 65.1 & 53.6 \\
PRO-DSC & \best{75.9} & 64.7 & 51.6 & 68.2 & 57.8 & 42.5 & 85.1 & 70.6 & 63.0 & 76.4 & 64.4 & 52.3 \\
\midrule
\multicolumn{13}{c}{\cellcolor{gray!40}\textbf{Language-assisted Image Clustering}} \\
SIC & 72.6 & 60.8 & 47.9 & 68.6 & 56.3 & 42.0 & 83.0 & 65.7 & 57.4 & 74.7 & 60.9 & 49.1 \\
TAC & 72.7 & 58.7 & 44.0 & 69.3 & 56.1 & 41.6 & 83.1 & 65.2 & 55.0 & 75.0 & 60.0 & 46.9 \\
TAC++ & 72.6 & 64.3 & 50.0 & 69.8 & 60.0 & 45.0 & 83.2 & 72.2 & 62.5 & 75.2 & 65.5 & 52.5 \\
SAC & 72.4 & 63.2 & 49.4 & 70.3 & 59.8 & 45.3 & 83.8 & 72.5 & 64.2 & 75.5 & 65.2 & 53.0 \\
GradNorm & 71.9 & 58.6 & 43.5 & 69.5 & 57.3 & 41.4 & 85.0 & 68.3 & 59.7 & 75.5 & 61.4 & 48.2 \\
NTK-SC & 74.5 & \best{66.4} & 49.3 & \best{71.6} & 61.1 & \best{47.2} & \best{86.6} & \best{76.8} & \best{68.7} & \best{77.5} & \best{68.1} & \best{55.0} \\
SEIC & 71.7 & 64.0 & 49.0 & 66.6 & 54.9 & 39.6 & 82.3 & 69.8 & 60.2 & 73.5 & 62.9 & 49.6 \\
MAGIC & 73.0 & 61.3 & 47.9 & 66.0 & 52.4 & 39.5 & 78.5 & 58.6 & 50.4 & 72.5 & 57.4 & 45.9 \\
\bottomrule
\end{tabular}
}
\end{table*}

\begin{table*}[tb]
\centering
\caption{Clustering results on large-scale datasets with SigLIP ViT-B/16. The best result in each metric column is highlighted in \colorbox[HTML]{FFF2CC}{\textbf{bold}}.}
\label{tab:siglip_b16_largescale}
\resizebox{0.7\textwidth}{!}{%
\begin{tabular}{l|ccc|ccc|ccc}
\toprule
Dataset
 & \multicolumn{3}{c|}{Places}
 & \multicolumn{3}{c|}{ImageNet}
 & \multicolumn{3}{c}{Average} \\
Metric & NMI & ACC & ARI
 & NMI & ACC & ARI
 & NMI & ACC & ARI \\
\midrule


\multicolumn{10}{c}{\cellcolor{gray!40}\textbf{Classical Image Clustering}} \\
$k$-means & 60.0 & 32.8 & 20.2 & 80.9 & 52.5 & 42.6 & 70.4 & 42.6 & 31.4 \\
SC & 58.2 & 32.6 & 20.2 & 80.5 & 48.7 & 40.4 & 69.3 & 40.6 & 30.3 \\
SSC-OMP & 54.6 & 26.8 & 13.2 & 73.7 & 42.6 & 21.4 & 64.2 & 34.7 & 17.3 \\
EnSC-ORGEN & 58.9 & 32.7 & 17.2 & 80.9 & 55.2 & 35.7 & 69.9 & 44.0 & 26.5 \\
\midrule
\multicolumn{10}{c}{\cellcolor{gray!40}\textbf{Deep Image Clustering}} \\
IDC & 59.8 & 33.0 & 19.7 & 81.4 & 56.4 & 44.0 & 70.6 & 44.7 & 31.9 \\
SCAN & 60.9 & 35.0 & 21.5 & 81.8 & 55.3 & 44.8 & 71.3 & 45.2 & 33.2 \\
CPP & 54.7 & 29.7 & 10.4 & 77.5 & 51.4 & 11.2 & 66.1 & 40.6 & 10.8 \\
TEMI & 61.6 & 35.9 & \best{22.1} & 84.9 & 65.4 & 53.2 & \best{73.2} & 50.7 & 37.6 \\
PRO-DSC & 56.2 & 27.2 & 14.4 & 78.5 & 50.5 & 29.8 & 67.3 & 38.8 & 22.1 \\
\midrule
 \multicolumn{10}{c}{\cellcolor{gray!40}\textbf{Language-assisted Image Clustering}} \\
SIC & 60.0 & 33.0 & 19.8 & 84.9 & 63.9 & 52.8 & 72.4 & 48.4 & 36.3 \\
TAC & 60.9 & 34.7 & 21.0 & 84.4 & 59.7 & 49.2 & 72.6 & 47.2 & 35.1 \\
TAC++ & 61.4 & 35.6 & 21.0 & 84.9 & \best{68.3} & 55.1 & \best{73.2} & 52.0 & 38.0 \\
SAC & 59.8 & \best{36.1} & 21.9 & \best{85.0} & 68.1 & \best{55.5} & 72.4 & \best{52.1} & \best{38.7} \\
GradNorm & 60.9 & 34.7 & 21.4 & 84.4 & 59.6 & 49.1 & 72.7 & 47.1 & 35.2 \\
NTK-SC & 60.3 & 35.6 & \best{22.1} & 84.3 & 65.7 & 52.7 & 72.3 & 50.7 & 37.4 \\
SEIC & \best{61.8} & 35.0 & 20.3 & 82.6 & 57.6 & 44.1 & 72.2 & 46.3 & 32.2 \\
MAGIC & 60.9 & 35.3 & 21.3 & 83.5 & 58.0 & 47.4 & 72.2 & 46.7 & 34.4 \\
\bottomrule
\end{tabular}
}
\end{table*}

\begin{table*}[tb]
\centering
\caption{Clustering results on fine-grained datasets with SigLIP ViT-B/16. The best result in each metric column is highlighted in \colorbox[HTML]{FFF2CC}{\textbf{bold}}.}
\label{tab:siglip_b16_finegrained}
\resizebox{\textwidth}{!}{%
\begin{tabular}{l|ccc|ccc|ccc|ccc|ccc|ccc}
\toprule
Dataset
 & \multicolumn{3}{c|}{Aircraft}
 & \multicolumn{3}{c|}{Flowers}
 & \multicolumn{3}{c|}{Food}
 & \multicolumn{3}{c|}{Cars}
 & \multicolumn{3}{c|}{Pets}
 & \multicolumn{3}{c}{Average} \\
Metric & NMI & ACC & ARI
 & NMI & ACC & ARI
 & NMI & ACC & ARI
 & NMI & ACC & ARI
 & NMI & ACC & ARI
 & NMI & ACC & ARI \\
\midrule


\multicolumn{19}{c}{\cellcolor{gray!40}\textbf{Classical Image Clustering}} \\
$k$-means & 71.8 & 44.0 & 37.2 & 94.4 & 79.7 & 80.5 & 86.7 & 80.0 & 71.7 & 91.0 & 69.9 & 66.4 & 87.2 & 78.1 & 73.2 & 86.2 & 70.3 & 65.8 \\
SC & 63.6 & 39.2 & 33.2 & 94.4 & 87.3 & 87.9 & 83.7 & 75.4 & 68.7 & 88.4 & 63.7 & 62.9 & 72.3 & 57.5 & 53.3 & 80.5 & 64.6 & 61.2 \\
SSC-OMP & 55.2 & 34.1 & 21.1 & 69.9 & 47.3 & 41.6 & 73.2 & 60.7 & 49.2 & 79.8 & 58.2 & 47.2 & 63.9 & 56.0 & 42.6 & 68.4 & 51.2 & 40.3 \\
EnSC-ORGEN & 63.9 & 39.0 & 28.9 & 90.9 & 71.1 & 70.3 & 85.7 & 78.3 & 69.6 & 91.1 & 74.1 & 69.8 & 85.1 & 78.0 & 69.8 & 83.3 & 68.1 & 61.7 \\
\midrule
\multicolumn{19}{c}{\cellcolor{gray!40}\textbf{Deep Image Clustering}} \\
IDC & \best{74.7} & \best{52.0} & 42.5 & 79.1 & 56.8 & 48.7 & 86.2 & 80.4 & 69.3 & 89.7 & 72.9 & 62.6 & 88.0 & 82.4 & 74.1 & 83.5 & 68.9 & 59.4 \\
SCAN & 58.2 & 20.4 & 14.9 & 86.1 & 59.7 & 57.3 & 78.3 & 48.8 & 46.7 & 79.9 & 24.7 & 27.4 & 89.2 & 81.8 & 77.4 & 78.3 & 47.1 & 44.7 \\
CPP & 66.6 & 42.2 & 32.1 & 92.5 & 77.7 & 76.8 & 88.0 & 78.6 & 68.5 & 88.3 & 69.9 & 63.5 & 88.9 & 82.0 & 76.5 & 84.9 & 70.1 & 63.5 \\
TEMI & 72.0 & 48.2 & 39.1 & 58.6 & 32.2 & 23.4 & \best{89.9} & \best{89.1} & \best{81.5} & 93.9 & \best{82.2} & \best{78.9} & 89.5 & 79.7 & 76.0 & 80.8 & 66.3 & 59.8 \\
PRO-DSC & 69.1 & 46.6 & 36.9 & 94.8 & 81.0 & 80.2 & 85.9 & 76.4 & 67.7 & \textbf{94.2} & 81.4 & 78.0 & 90.6 & 85.9 & 80.7 & 86.9 & 74.3 & 68.7 \\
\midrule
\multicolumn{19}{c}{\cellcolor{gray!40}\textbf{Language-assisted Image Clustering}} \\
SIC & 71.6 & 47.2 & 38.6 & 90.2 & 77.2 & 71.7 & 87.5 & 82.4 & 75.0 & 90.7 & 71.0 & 66.4 & 89.1 & 85.7 & 79.0 & 85.8 & 72.7 & 66.1 \\
TAC & 72.3 & 45.4 & 38.4 & 94.3 & 80.1 & 80.2 & 88.5 & 81.0 & 73.0 & 91.3 & 70.9 & 67.7 & 89.2 & 81.4 & 76.5 & 87.1 & 71.8 & 67.2 \\
TAC++ & 70.5 & 49.4 & 37.8 & \best{97.8} & \best{96.6} & \best{94.6} & 88.0 & 86.0 & 78.0 & 91.5 & 80.0 & 74.1 & 91.0 & 87.8 & 81.6 & 87.8 & \best{79.9} & \best{73.2} \\
SAC & 70.5 & 47.5 & 38.8 & 93.9 & 84.2 & 81.8 & 85.6 & 84.0 & 74.3 & 91.5 & 80.8 & 74.4 & 90.7 & 88.2 & 82.5 & 86.4 & 76.9 & 70.4 \\
GradNorm & 72.0 & 46.7 & 38.6 & 94.5 & 85.6 & 83.7 & 86.8 & 80.4 & 70.9 & 91.0 & 70.0 & 67.0 & 90.9 & 83.7 & 78.9 & 87.1 & 73.3 & 67.8 \\
NTK-SC & 72.8 & 51.6 & \best{43.5} & 95.5 & 88.0 & 85.1 & 88.0 & 84.2 & 74.3 & 92.0 & 79.8 & 74.8 & \best{91.6} & \best{89.0} & \best{83.5} & \best{88.0} & 78.5 & 72.3 \\
SEIC & 61.1 & 31.9 & 22.6 & 84.0 & 67.6 & 63.3 & 83.8 & 80.8 & 70.8 & 77.6 & 53.2 & 42.3 & 74.2 & 50.4 & 42.3 & 76.1 & 56.8 & 48.3 \\
MAGIC & 61.7 & 29.3 & 21.4 & 73.4 & 43.2 & 37.0 & 89.1 & 88.1 & 80.3 & 73.0 & 33.6 & 29.5 & 88.2 & 79.3 & 73.2 & 77.1 & 54.7 & 48.3 \\
\bottomrule
\end{tabular}
}
\end{table*}

\section{Detailed Results on Robustness Analysis}
\label{appendix_f}

Table~\ref{tab:anyattack-b32-classical-eps8}, Table~\ref{tab:anyattack-b32-challenging-eps8}, Table~\ref{tab:anyattack_eps8_large_scale}, Table~\ref{tab:anyattack-b32-finegrained-eps8} supplement Table~\ref{tab:autoattack_b32} by reporting post-attack clustering results on the classical, challenging, large-scale, and fine-grained datasets, respectively. 

\begin{table*}[tb]
\centering
\caption{Average pre-attack and post-attack clustering performance across
15 datasets using CLIP ViT-B/32 pretrained on LAION-400M. The best result
in each metric column is highlighted in \colorbox[HTML]{FFF2CC}{\textbf{bold}}.}
\label{tab:autoattack_b32}

\resizebox{0.8\textwidth}{!}{%
\begin{tabular}{l|ccc|ccc|ccc}
\toprule
Dataset
& \multicolumn{3}{c|}{Pre-attack}
& \multicolumn{3}{c|}{Post-attack}
& \multicolumn{3}{c}{Change (\%)} \\
Metric
& NMI & ACC & ARI
& NMI & ACC & ARI
& NMI & ACC & ARI \\
\midrule

\multicolumn{10}{c}{%
  \cellcolor{gray!40}\textbf{Classical Image Clustering}} \\

$k$-means
& 73.7 & 58.8 & 50.0
& 56.5 & 44.7 & 32.2
& 23.3 & 24.0 & 35.6 \\

SC
& 69.5 & 57.8 & 48.5
& 53.2 & 41.8 & 30.7
& 23.5 & 27.7 & 36.7 \\

SSC-OMP
& 58.0 & 44.5 & 31.8
& 45.0 & 33.4 & 20.4
& 22.4 & 24.9 & 35.8 \\

EnSC-ORGEN
& 68.3 & 56.9 & 43.7
& 54.0 & 43.7 & 28.8
& \best{20.9} & 23.2 & 34.1 \\

\midrule
\multicolumn{10}{c}{%
  \cellcolor{gray!40}\textbf{Deep Image Clustering}} \\

IDC
& 71.9 & 60.4 & 47.6
& 54.7 & 45.1 & 28.4
& 23.9 & 25.3 & 40.3 \\

SCAN
& 71.2 & 54.5 & 45.9
& 55.3 & 42.2 & 28.3
& 22.3 & \best{22.6} & 38.3 \\

CPP
& 73.0 & 62.2 & 48.1
& 54.0 & 43.5 & 28.5
& 26.0 & 30.1 & 40.7 \\

TEMI
& 75.3 & 64.7 & 54.2
& 56.5 & 47.3 & 32.2
& 25.0 & 26.9 & 40.6 \\

PRO-DSC
& 74.4 & 63.2 & 51.4
& 56.4 & 47.1 & 32.0
& 24.2 & 25.5 & 37.7 \\

\midrule
\multicolumn{10}{c}{%
  \cellcolor{gray!40}\textbf{Language-assisted Image Clustering}} \\

SIC
& 74.9 & 63.0 & 53.3
& 57.3 & 47.8 & 31.3
& 23.5 & 24.1 & 41.3 \\

TAC
& 76.3 & 64.0 & 54.0
& 57.9 & 47.3 & 34.5
& 24.1 & 26.1 & 36.1 \\

TAC++
& 76.2 & 67.0 & 56.4
& 56.8 & 48.7 & 32.5
& 25.5 & 27.3 & 42.4 \\

SAC
& 75.2 & 65.7 & 55.1
& 56.2 & 48.6 & 32.4
& 25.3 & 26.0 & 41.2 \\

GradNorm
& 76.5 & 65.3 & 54.7
& 58.3 & 47.0 & 34.4
& 23.8 & 28.0 & 37.1 \\

NTK-SC
& \best{78.2} & \best{68.5} & \best{57.7}
& \best{60.5} & \best{52.2} & \best{38.6}
& 22.6 & 23.8 & \best{33.1} \\

SEIC
& 72.9 & 61.3 & 50.2
& 54.4 & 45.2 & 29.6
& 25.4 & 26.3 & 41.0 \\

MAGIC
& 72.3 & 58.1 & 48.8
& 53.4 & 41.7 & 28.7
& 26.1 & 28.2 & 41.2 \\

\bottomrule
\end{tabular}%
}
\end{table*}

\begin{table*}[tb]
\centering
\caption{Post-attack clustering performance on classical datasets using CLIP ViT-B/32 pretrained on LAION-400M. The best result in each metric column is highlighted in \colorbox[HTML]{FFF2CC}{\textbf{bold}}.}
\label{tab:anyattack-b32-classical-eps8}
\resizebox{\textwidth}{!}{%
\begin{tabular}{l|ccc|ccc|ccc|ccc|ccc|ccc}
\toprule
Dataset
& \multicolumn{3}{c|}{CIFAR-10}
& \multicolumn{3}{c|}{CIFAR-20}
& \multicolumn{3}{c|}{STL-10}
& \multicolumn{3}{c|}{ImageNet-10}
& \multicolumn{3}{c|}{ImageNet-Dogs}
& \multicolumn{3}{c}{Average} \\
Metric & NMI & ACC & ARI & NMI & ACC & ARI & NMI & ACC & ARI
& NMI & ACC & ARI & NMI & ACC & ARI & NMI & ACC & ARI \\
\midrule
\multicolumn{19}{c}{\cellcolor{gray!40}\textbf{Classical Image Clustering}} \\
$k$-means & 11.8 & 23.4 & 7.0 & 14.8 & 20.9 & 6.3 & 71.2 & 76.4 & 61.3 & 88.6 & 91.8 & 83.5 & 51.5 & 49.7 & 31.4 & 47.6 & 52.4 & 37.9 \\
SC & 11.8 & 23.7 & 7.4 & 13.6 & 20.5 & 6.5 & 69.3 & 75.1 & 61.1 & 86.5 & 91.0 & 81.4 & 41.2 & 41.7 & 24.5 & 44.5 & 50.4 & 36.2 \\
SSC-OMP & 11.5 & 22.8 & 7.8 & 11.7 & 17.9 & 6.1 & 67.9 & 73.6 & 59.5 & 81.2 & 89.0 & 78.1 & 29.1 & 32.4 & 14.0 & 40.3 & 47.1 & 33.1 \\
EnSC-ORGEN & \best{14.1} & 25.7 & \best{9.2} & 17.2 & \best{21.9} & \best{7.8} & 61.8 & 68.9 & 49.0 & 83.2 & 90.0 & 78.4 & 41.0 & 44.5 & 23.7 & 43.5 & 50.2 & 33.6 \\
\midrule
\multicolumn{19}{c}{\cellcolor{gray!40}\textbf{Deep Image Clustering}} \\
IDC & 12.4 & 30.2 & 5.9 & 15.4 & 19.0 & 2.3 & 74.1 & 86.2 & 71.5 & 89.1 & 93.6 & 87.3 & 58.2 & 59.7 & 37.2 & 49.8 & 57.7 & 40.8 \\
SCAN & 11.4 & 25.7 & 2.9 & \best{17.4} & 21.4 & 2.6 & 73.8 & 85.4 & 70.9 & 90.5 & 94.6 & 88.7 & 59.1 & 58.1 & 41.4 & 50.4 & 57.0 & 41.3 \\
CPP & 10.9 & 27.0 & 6.2 & 12.2 & 18.7 & 5.2 & 67.7 & 72.2 & 55.7 & 90.7 & 94.0 & 88.1 & 54.7 & 54.5 & 33.7 & 47.2 & 53.3 & 37.8 \\
TEMI & 13.6 & 26.5 & 2.7 & 15.1 & 19.1 & 3.1 & 74.7 & 85.8 & 70.4 & 91.2 & 95.2 & 89.9 & 63.5 & 63.3 & 46.6 & 51.6 & 58.0 & 42.5 \\
PRO-DSC & 11.4 & \best{30.9} & 7.8 & 14.3 & 21.2 & 6.3 & 74.9 & 86.6 & 72.2 & 85.7 & 91.8 & 82.5 & 61.0 & 60.7 & 41.3 & 49.5 & 58.2 & 42.0 \\
\midrule
\multicolumn{19}{c}{\cellcolor{gray!40}\textbf{Language-assisted Image Clustering}} \\
SIC & 13.7 & 28.8 & 3.6 & 14.2 & 19.5 & 3.4 & 72.2 & 84.8 & 69.2 & 86.0 & 90.2 & 81.6 & 65.4 & 66.8 & 49.8 & 50.3 & 58.0 & 41.5 \\
TAC & 10.7 & 28.0 & 7.6 & 15.4 & 20.1 & 6.0 & 72.1 & 77.6 & 66.0 & 90.2 & 94.2 & 87.7 & 69.7 & 71.5 & 57.0 & 51.6 & 58.3 & 44.9 \\
TAC++ & 12.9 & 29.5 & 5.8 & 13.2 & 18.5 & 2.2 & 75.4 & 87.2 & 73.6 & 90.2 & 94.4 & 88.2 & 69.3 & 73.5 & 57.2 & 52.2 & 60.6 & 45.4 \\
SAC & 11.2 & 27.6 & 4.0 & 12.6 & 18.1 & 3.4 & 74.6 & 87.0 & 73.6 & 87.0 & 92.0 & 84.3 & 69.2 & 74.0 & 56.3 & 50.9 & 59.7 & 44.3 \\
GradNorm & 10.0 & 25.5 & 5.9 & 15.6 & 20.5 & 6.0 & 71.6 & 76.3 & 65.8 & 85.2 & 84.8 & 78.4 & 67.4 & 66.8 & 51.1 & 50.0 & 54.8 & 41.4 \\
NTK-SC & 12.8 & 30.1 & 8.7 & 13.6 & 17.6 & 2.5 & \best{77.4} & \best{88.5} & \best{76.3} & 90.0 & 94.4 & 88.7 & \best{71.4} & \best{75.6} & \best{60.7} & \best{53.1} & \best{61.2} & \best{47.4} \\
SEIC & 10.4 & 28.6 & 6.2 & 12.0 & 17.6 & 2.0 & 73.7 & 86.4 & 72.3 & 88.5 & 93.2 & 86.5 & 63.6 & 67.3 & 51.5 & 49.6 & 58.6 & 43.7 \\
MAGIC & 11.5 & 26.7 & 3.5 & 13.1 & 18.5 & 1.8 & 75.6 & 86.6 & 72.0 & \best{91.8} & \best{95.8} & \best{91.1} & 70.4 & 74.3 & 58.3 & 52.5 & 60.4 & 45.3 \\
\bottomrule
\end{tabular}%
}
\end{table*}

\begin{table*}[tb]
\centering
\caption{Post-attack clustering performance on challenging datasets using CLIP ViT-B/32 pretrained on LAION-400M. The best result in each metric column is highlighted in \colorbox[HTML]{FFF2CC}{\textbf{bold}}.}
\label{tab:anyattack-b32-challenging-eps8}
\resizebox{0.8\textwidth}{!}{%
\begin{tabular}{l|ccc|ccc|ccc|ccc}
\toprule
Dataset
& \multicolumn{3}{c|}{CIFAR-100}
& \multicolumn{3}{c|}{DTD}
& \multicolumn{3}{c|}{UCF-101}
& \multicolumn{3}{c}{Average} \\
Metric & NMI & ACC & ARI & NMI & ACC & ARI
& NMI & ACC & ARI & NMI & ACC & ARI \\
\midrule
\multicolumn{13}{c}{\cellcolor{gray!40}\textbf{Classical Image Clustering}} \\
$k$-means & 27.6 & 13.3 & 4.3 & 57.4 & 46.1 & 28.6 & 69.4 & 46.2 & 33.5 & 51.5 & 35.2 & 22.2 \\
SC & 27.1 & 13.5 & \best{5.7} & 58.2 & 48.0 & 33.1 & 68.8 & 45.5 & 35.2 & 51.3 & 35.7 & 24.7 \\
SSC-OMP & 23.1 & 10.4 & 3.8 & 49.5 & 36.0 & 20.6 & 60.2 & 33.9 & 20.4 & 44.3 & 26.8 & 14.9 \\
EnSC-ORGEN & 28.9 & 14.9 & 5.6 & 55.1 & 43.0 & 25.7 & 67.3 & 41.6 & 28.5 & 50.4 & 33.2 & 19.9 \\
\midrule
\multicolumn{13}{c}{\cellcolor{gray!40}\textbf{Deep Image Clustering}} \\
IDC & 18.0 & 8.5 & 0.8 & 58.7 & 49.0 & 26.2 & 64.7 & 47.8 & 27.1 & 47.1 & 35.1 & 18.0 \\
SCAN & 19.5 & 8.9 & 0.4 & 59.7 & 46.8 & 19.2 & 58.4 & 29.7 & 11.7 & 45.9 & 28.4 & 10.5 \\
CPP & 25.8 & 13.3 & 3.9 & 57.0 & 47.1 & 28.9 & 67.2 & 46.1 & 32.6 & 50.0 & 35.5 & 21.8 \\
TEMI & 19.2 & 10.0 & 0.5 & 62.0 & 50.2 & 26.7 & 68.8 & 51.0 & 33.3 & 50.0 & 37.1 & 20.2 \\
PRO-DSC & 26.6 & 15.7 & 4.5 & 58.5 & 46.8 & 29.0 & 72.2 & 54.5 & 41.6 & 52.4 & 39.0 & 25.1 \\
\midrule
\multicolumn{13}{c}{\cellcolor{gray!40}\textbf{Language-assisted Image Clustering}} \\
SIC & 22.5 & 12.7 & 1.4 & 59.8 & 47.5 & 27.0 & 66.5 & 45.9 & 18.9 & 49.6 & 35.4 & 15.8 \\
TAC & 25.1 & 13.2 & 3.6 & 57.8 & 45.5 & 28.1 & 67.8 & 47.7 & 32.6 & 50.2 & 35.4 & 21.4 \\
TAC++ & 21.0 & 12.0 & 0.6 & 60.3 & 47.9 & 28.4 & 63.9 & 48.6 & 21.3 & 48.4 & 36.2 & 16.8 \\
SAC & 20.1 & 11.6 & 0.9 & 57.9 & 48.2 & 26.2 & 64.9 & 49.1 & 25.1 & 47.6 & 36.3 & 17.4 \\
GradNorm & 28.0 & 13.2 & 4.4 & 60.8 & 48.6 & 30.6 & 69.7 & 47.3 & 33.9 & 52.8 & 36.4 & 23.0 \\
NTK-SC & \best{29.8} & \best{16.5} & 5.6 & \best{63.2} & \best{53.1} & \best{36.4} & \best{72.6} & \best{56.6} & \best{44.1} & \best{55.2} & \best{42.1} & \best{28.7} \\
SEIC & 20.0 & 11.5 & 0.3 & 57.1 & 46.6 & 26.3 & 64.9 & 49.4 & 23.7 & 47.4 & 35.8 & 16.8 \\
MAGIC & 21.4 & 11.5 & 1.5 & 54.2 & 38.4 & 23.1 & 60.0 & 37.5 & 23.0 & 45.2 & 29.1 & 15.8 \\
\bottomrule
\end{tabular}%
}
\end{table*}

\begin{table*}[tb]
\centering
\caption{Post-attack clustering performance on large-scale datasets using CLIP ViT-B/32 pretrained on LAION-400M. The best result in each metric column is highlighted in \colorbox[HTML]{FFF2CC}{\textbf{bold}}.}
\label{tab:anyattack_eps8_large_scale}
\resizebox{0.7\textwidth}{!}{%
\begin{tabular}{l|ccc|ccc|ccc}
\toprule
Dataset & \multicolumn{3}{c|}{Places365} & \multicolumn{3}{c|}{ImageNet-1K} & \multicolumn{3}{c}{Average} \\
Metric & NMI & ACC & ARI & NMI & ACC & ARI & NMI & ACC & ARI \\
\midrule
\multicolumn{10}{c}{\cellcolor{gray!40}\textbf{Classical Image Clustering}} \\
$k$-means & 54.1 & 25.8 & 13.5 & 68.1 & 32.7 & 20.4 & 61.1 & 29.2 & 16.9 \\
SC & 52.4 & 24.9 & 14.6 & 67.5 & 31.9 & 21.4 & 60.0 & 28.4 & 18.0 \\
SSC-OMP & 47.8 & 20.7 & 8.9 & 60.8 & 25.1 & 8.6 & 54.3 & 22.9 & 8.7 \\
EnSC-ORGEN & 52.9 & 26.6 & 11.7 & 69.1 & 38.0 & 19.7 & 61.0 & 32.3 & 15.7 \\
\midrule
\multicolumn{10}{c}{\cellcolor{gray!40}\textbf{Deep Image Clustering}} \\
IDC & 53.4 & 26.2 & 7.5 & 67.7 & 35.7 & 15.4 & 60.6 & 31.0 & 11.5 \\
SCAN & 55.8 & 28.8 & 14.8 & 70.3 & 39.8 & 15.5 & 63.0 & 34.3 & 15.2 \\
CPP & 49.0 & 21.8 & 10.0 & 64.5 & 32.5 & 10.8 & 56.8 & 27.2 & 10.4 \\
TEMI & \best{56.5} & \best{30.7} & \best{15.9} & \best{73.0} & \best{46.5} & 25.9 & \best{64.7} & \best{38.6} & \best{20.9} \\
PRO-DSC & 49.1 & 20.6 & 8.7 & 65.2 & 33.5 & 12.7 & 57.2 & 27.0 & 10.7 \\
\midrule
\multicolumn{10}{c}{\cellcolor{gray!40}\textbf{Language-assisted Image Clustering}} \\
SIC & 54.5 & 26.9 & 14.2 & 72.2 & 43.9 & 15.8 & 63.3 & 35.4 & 15.0 \\
TAC & 55.2 & 27.7 & 14.5 & 71.7 & 40.4 & 26.2 & 63.5 & 34.0 & 20.3 \\
TAC++ & 54.2 & 27.6 & 13.0 & 72.2 & 45.7 & \best{27.3} & 63.2 & 36.7 & 20.1 \\
SAC & 54.1 & 29.1 & 13.5 & 72.1 & 45.5 & 26.5 & 63.1 & 37.3 & 20.0 \\
GradNorm & 55.1 & 27.4 & 14.7 & 72.0 & 41.0 & 27.0 & 63.5 & 34.2 & 20.8 \\
NTK-SC & 54.7 & 27.8 & 13.7 & 72.1 & 43.7 & 26.9 & 63.4 & 35.8 & 20.3 \\
SEIC & 55.1 & 27.4 & 13.7 & 70.3 & 38.2 & 21.8 & 62.7 & 32.8 & 17.8 \\
MAGIC & 55.8 & 29.5 & 15.5 & 69.7 & 35.5 & 22.0 & 62.8 & 32.5 & 18.8 \\
\bottomrule
\end{tabular}%
}
\end{table*}

\begin{table*}[tb]
\centering
\caption{Post-attack clustering performance on fine-grained datasets using CLIP ViT-B/32 pretrained on LAION-400M. The best result in each metric column is highlighted in \colorbox[HTML]{FFF2CC}{\textbf{bold}}.}
\label{tab:anyattack-b32-finegrained-eps8}
\resizebox{\textwidth}{!}{%
\begin{tabular}{l|ccc|ccc|ccc|ccc|ccc|ccc}
\toprule
Dataset
& \multicolumn{3}{c|}{Aircraft}
& \multicolumn{3}{c|}{Cars}
& \multicolumn{3}{c|}{Flowers}
& \multicolumn{3}{c|}{Food}
& \multicolumn{3}{c|}{Pets}
& \multicolumn{3}{c}{Average} \\
Metric & NMI & ACC & ARI & NMI & ACC & ARI & NMI & ACC & ARI
& NMI & ACC & ARI & NMI & ACC & ARI & NMI & ACC & ARI \\
\midrule
\multicolumn{19}{c}{\cellcolor{gray!40}\textbf{Classical Image Clustering}} \\
$k$-means & 43.3 & 18.5 & 8.4 & 73.3 & 44.6 & 35.0 & 81.5 & 65.5 & 58.0 & 64.3 & 57.1 & 40.9 & 71.4 & 59.0 & 50.1 & 66.7 & 48.9 & 38.5 \\
SC & 40.6 & 17.0 & 7.7 & 69.2 & 38.8 & 31.7 & 80.5 & 66.8 & 61.8 & 57.1 & 46.5 & 34.2 & 54.9 & 42.6 & 34.8 & 60.5 & 42.4 & 34.1 \\
SSC-OMP & 39.2 & 15.9 & 5.2 & 57.8 & 30.6 & 16.6 & 53.5 & 32.2 & 22.0 & 41.8 & 31.2 & 17.8 & 40.0 & 29.0 & 15.9 & 46.5 & 27.8 & 15.5 \\
EnSC-ORGEN & 42.4 & 19.4 & 7.6 & 70.5 & 47.2 & 32.1 & 82.7 & 65.8 & 58.8 & 62.8 & 55.1 & 38.1 & 61.4 & 53.4 & 35.8 & 64.0 & 48.2 & 34.5 \\
\midrule
\multicolumn{19}{c}{\cellcolor{gray!40}\textbf{Deep Image Clustering}} \\
IDC & 40.6 & 17.6 & 6.9 & 71.6 & 47.3 & 30.8 & 60.7 & 35.9 & 24.0 & 61.9 & 53.6 & 29.1 & 74.6 & 66.5 & 53.5 & 61.9 & 44.2 & 28.9 \\
SCAN & 36.7 & 11.7 & 5.5 & 68.0 & 24.7 & 22.4 & 74.0 & 49.4 & 42.6 & 59.8 & 40.1 & 30.5 & 74.4 & 67.3 & 55.6 & 62.6 & 38.7 & 31.3 \\
CPP & 40.9 & 17.6 & 7.3 & 68.6 & 42.8 & 28.1 & 78.6 & 61.4 & 52.1 & 63.0 & 52.8 & 29.2 & 60.0 & 50.0 & 35.3 & 62.2 & 44.9 & 30.4 \\
TEMI & 42.7 & 19.3 & 8.5 & \best{78.5} & \best{56.2} & \best{43.6} & 45.1 & 22.6 & 12.3 & \best{70.3} & \best{67.2} & \best{50.0} & 73.8 & 66.5 & 53.8 & 62.1 & 46.4 & 33.6 \\
PRO-DSC & 41.4 & 17.6 & 7.1 & 74.4 & 48.6 & 37.9 & 78.4 & 60.5 & 51.3 & 55.8 & 42.4 & 15.0 & \best{77.5} & \best{74.8} & \best{61.4} & 65.5 & 48.8 & 34.6 \\
\midrule
\multicolumn{19}{c}{\cellcolor{gray!40}\textbf{Language-assisted Image Clustering}} \\
SIC & 43.2 & 19.2 & 8.1 & 73.7 & 46.2 & 35.1 & 77.3 & 58.9 & 50.1 & 66.2 & 60.6 & 39.3 & 72.5 & 64.6 & 51.7 & 66.6 & 49.9 & 36.9 \\
TAC & 43.1 & 18.7 & 8.1 & 73.3 & 45.1 & 35.1 & 82.6 & 65.7 & 59.3 & 59.8 & 49.6 & 32.1 & 73.8 & 64.6 & 53.3 & 66.5 & 48.8 & 37.6 \\
TAC++ & 38.0 & 16.3 & 4.6 & 69.7 & 42.8 & 27.5 & 75.2 & 63.5 & 47.1 & 60.7 & 53.8 & 35.1 & 75.6 & 69.1 & 55.8 & 63.8 & 49.1 & 34.0 \\
SAC & 39.0 & 16.7 & 6.2 & 71.9 & 48.4 & 32.4 & 74.1 & 55.5 & 45.4 & 62.4 & 58.7 & 39.2 & 71.2 & 67.0 & 49.0 & 63.7 & 49.3 & 34.4 \\
GradNorm & 43.0 & 18.2 & 7.9 & 71.7 & 42.2 & 32.1 & \best{84.4} & 69.8 & \textbf{62.2} & 63.9 & 55.1 & 38.0 & 76.8 & 67.8 & 58.3 & 68.0 & 50.6 & 39.7 \\
NTK-SC & \best{45.5} & \best{20.9} & \best{9.9} & 75.6 & 51.1 & 39.9 & 83.9 & \best{74.1} & 62.0 & 67.6 & 61.0 & 43.5 & 77.4 & 72.6 & 59.3 & \best{70.0} & \best{55.9} & \best{42.9} \\
SEIC & 38.2 & 15.9 & 5.4 & 63.3 & 35.8 & 17.8 & 69.7 & 51.2 & 40.7 & 61.5 & 56.2 & 37.8 & 67.6 & 51.8 & 38.3 & 60.1 & 42.2 & 28.0 \\
MAGIC & 37.1 & 13.9 & 4.0 & 55.5 & 20.3 & 12.6 & 51.8 & 25.8 & 17.3 & 62.3 & 52.9 & 37.7 & 70.6 & 58.4 & 46.9 & 55.5 & 34.3 & 23.7 \\
\bottomrule
\end{tabular}%
}
\end{table*}

\section{Discriminability Analysis}
In this section, we perform a discriminability analysis to assess whether the groups produced by an algorithm are meaningful and well separated based on the intrinsic structure of the data. This analysis complements the effectiveness analysis in Section.~\ref{effectiveness} by providing an unsupervised evaluation that does not require known class labels or external ground truth. Specifically, we consider the Silhouette Coefficient (SIL), Davies–Bouldin Index (DBI), and Calinski–Harabasz Index (CHI). Detailed definitions of these metrics are provided in Appendix~\ref{appendix_c}. The responding results on classical, challenging, large-scale and fine-grained datasets are present in Table~\ref{tab:laion400m_b32_label_free_classical}, Table~\ref{tab:laion400m_b32_label_free_challenging}, Table~\ref{tab:laion400m_b32_label_free_large_scale}, and Table~\ref{tab:laion400m_b32_label_free_fine_grained}.

\begin{table*}[tb]
\centering
\caption{Discriminability analysis on classical datasets using CLIP ViT-B/32 pretrained on LAION-400M.  $\uparrow$ indicates larger values are better and vice versa. The best value in each metric column is highlighted in \colorbox[HTML]{FFF2CC}{\textbf{bold}}.}
\label{tab:laion400m_b32_label_free_classical}
\resizebox{\textwidth}{!}{%
\begin{tabular}{l|ccc|ccc|ccc|ccc|ccc|ccc}
\toprule
Dataset & \multicolumn{3}{c|}{STL-10} & \multicolumn{3}{c|}{CIFAR-10} & \multicolumn{3}{c|}{CIFAR-20} & \multicolumn{3}{c|}{ImageNet-10} & \multicolumn{3}{c|}{ImageNet-Dogs} & \multicolumn{3}{c}{Average} \\
Metric & SIL$\uparrow$ & DBI$\downarrow$ & CHI$\uparrow$ & SIL$\uparrow$ & DBI$\downarrow$ & CHI$\uparrow$ & SIL$\uparrow$ & DBI$\downarrow$ & CHI$\uparrow$ & SIL$\uparrow$ & DBI$\downarrow$ & CHI$\uparrow$ & SIL$\uparrow$ & DBI$\downarrow$ & CHI$\uparrow$ & SIL$\uparrow$ & DBI$\downarrow$ & CHI$\uparrow$ \\
\midrule
\multicolumn{19}{c}{\cellcolor{gray!40}\textbf{Classical Image Clustering}} \\
$k$-means & \best{0.207} & 2.590 & \best{419.44} & 0.172 & \best{2.903} & \best{441.43} & \best{0.104} & \best{3.250} & \best{156.82} & 0.320 & 2.059 & 36.12 & \best{0.085} & \best{3.140} & \best{17.44} & \best{0.178} & \best{2.789} & \best{214.25} \\
SC & 0.203 & 2.622 & 415.40 & 0.164 & 2.974 & 430.53 & 0.071 & 3.583 & 142.61 & \best{0.325} & 2.020 & \best{37.51} & 0.045 & 3.214 & 14.78 & 0.162 & 2.883 & 208.17 \\
SSC-OMP & 0.160 & 3.405 & 367.07 & 0.120 & 3.777 & 383.19 & 0.056 & 4.154 & 128.14 & 0.297 & 2.113 & 34.72 & -0.030 & 5.255 & 8.38 & 0.120 & 3.741 & 184.30 \\
EnSC & 0.170 & 3.372 & 369.31 & 0.162 & 3.272 & 416.86 & 0.044 & 3.850 & 138.56 & 0.315 & 2.019 & 36.47 & 0.028 & 3.678 & 13.19 & 0.144 & 3.238 & 194.88 \\
\midrule
\multicolumn{19}{c}{\cellcolor{gray!40}\textbf{Deep Image Clustering}} \\
IDC & 0.207 & 2.592 & 417.86 & 0.171 & 2.950 & 439.51 & 0.067 & 3.683 & 142.25 & 0.318 & 2.101 & 36.67 & 0.076 & 3.283 & 16.83 & 0.168 & 2.922 & 210.62 \\
SCAN & 0.204 & 2.657 & 414.81 & 0.164 & 2.977 & 436.05 & 0.079 & 3.582 & 153.48 & 0.319 & 2.026 & 37.04 & 0.072 & 3.330 & 16.62 & 0.168 & 2.915 & 211.60 \\
CPP & 0.204 & 2.623 & 416.37 & 0.169 & 3.026 & 435.46 & 0.053 & 4.078 & 134.79 & 0.323 & 2.030 & 37.25 & 0.052 & 3.425 & 15.63 & 0.160 & 3.037 & 207.90 \\
TEMI & 0.207 & 2.591 & 418.18 & 0.172 & 2.955 & 438.13 & 0.056 & 3.888 & 134.76 & 0.322 & 2.037 & 37.29 & 0.078 & 3.230 & 16.60 & 0.167 & 2.940 & 209.00 \\
PRO-DSC & 0.206 & 2.597 & 418.85 & \best{0.172} & 2.960 & 440.31 & 0.070 & 3.625 & 149.91 & 0.321 & 2.017 & 37.08 & 0.053 & 3.374 & 15.70 & 0.165 & 2.915 & 212.37 \\
\midrule
\multicolumn{19}{c}{\cellcolor{gray!40}\textbf{Language-assisted Image Clustering}} \\
SIC & 0.190 & 2.691 & 393.04 & 0.171 & 2.972 & 438.73 & 0.062 & 4.072 & 137.08 & 0.281 & 2.190 & 33.37 & 0.078 & 3.247 & 16.97 & 0.156 & 3.034 & 203.84 \\
TAC & 0.204 & \best{2.584} & 414.64 & 0.166 & 3.009 & 432.43 & 0.088 & 3.496 & 148.17 & 0.323 & \best{1.998} & 37.28 & 0.063 & 3.569 & 15.86 & 0.169 & 2.931 & 209.67 \\
TAC$^\dagger$ & 0.206 & 2.592 & 417.95 & 0.168 & 2.999 & 435.21 & 0.060 & 4.106 & 129.50 & 0.322 & 2.037 & 37.29 & 0.079 & 3.561 & 16.22 & 0.167 & 3.059 & 207.24 \\
SAC & 0.205 & 2.591 & 416.67 & 0.165 & 3.032 & 430.21 & 0.057 & 4.318 & 127.98 & 0.321 & 2.045 & 37.17 & 0.072 & 3.715 & 15.59 & 0.164 & 3.140 & 205.52 \\
GradNorm & 0.204 & 2.626 & 415.03 & 0.167 & 2.997 & 433.86 & 0.091 & 3.501 & 152.26 & 0.318 & 2.041 & 36.97 & 0.080 & 3.557 & 16.45 & 0.172 & 2.944 & 210.92 \\
NTK-SC & 0.206 & 2.608 & 416.72 & 0.167 & 3.004 & 433.30 & 0.062 & 3.478 & 128.18 & 0.321 & 2.030 & 37.22 & 0.077 & 3.578 & 15.92 & 0.167 & 2.940 & 206.27 \\
SEIC & 0.206 & 2.598 & 417.02 & 0.166 & 3.015 & 432.34 & 0.064 & 3.905 & 136.58 & 0.319 & 2.022 & 37.03 & 0.071 & 3.569 & 15.93 & 0.165 & 3.022 & 207.78 \\
MAGIC & 0.206 & 2.590 & 418.00 & 0.167 & 3.011 & 433.21 & 0.073 & 3.992 & 140.76 & 0.322 & 2.023 & 37.21 & 0.079 & 3.533 & 16.30 & 0.169 & 3.030 & 209.09 \\
\bottomrule
\end{tabular}
}
\end{table*}

\begin{table*}[tb]
\centering
\caption{Discriminability analysis on challenging datasets using CLIP ViT-B/32 pretrained on LAION-400M. $\uparrow$ indicates larger values are better and vice versa. The best value in each metric column is highlighted in \colorbox[HTML]{FFF2CC}{\textbf{bold}}.}
\label{tab:laion400m_b32_label_free_challenging}
\resizebox{0.9\textwidth}{!}{%
\begin{tabular}{l|ccc|ccc|ccc|ccc}
\toprule
Dataset & \multicolumn{3}{c|}{CIFAR-100} & \multicolumn{3}{c|}{DTD} & \multicolumn{3}{c|}{UCF101} & \multicolumn{3}{c}{Average} \\
Metric & SIL$\uparrow$ & DBI$\downarrow$ & CHI$\uparrow$ & SIL$\uparrow$ & DBI$\downarrow$ & CHI$\uparrow$ & SIL$\uparrow$ & DBI$\downarrow$ & CHI$\uparrow$ & SIL$\uparrow$ & DBI$\downarrow$ & CHI$\uparrow$ \\
\midrule
\multicolumn{13}{c}{\cellcolor{gray!40}\textbf{Classical Image Clustering}} \\
$k$-means & \best{0.105} & \best{3.140} & 58.57 & \best{0.138} & 2.771 & \best{21.08} & \best{0.211} & \best{2.307} & 33.63 & \best{0.151} & \best{2.739} & 37.76 \\
SC & 0.084 & 3.222 & 51.19 & 0.122 & 2.874 & 18.92 & 0.205 & 2.310 & 32.33 & 0.137 & 2.802 & 34.15 \\
SSC-OMP & 0.011 & 4.827 & 40.01 & 0.045 & 3.613 & 14.70 & 0.017 & 3.670 & 16.38 & 0.024 & 4.037 & 23.70 \\
EnSC & 0.048 & 3.430 & 51.17 & 0.064 & 3.277 & 17.26 & 0.104 & 2.643 & 26.47 & 0.072 & 3.117 & 31.63 \\
\midrule
\multicolumn{13}{c}{\cellcolor{gray!40}\textbf{Deep Image Clustering}} \\
IDC & 0.072 & 3.201 & 55.68 & 0.083 & 2.873 & 18.89 & 0.136 & 2.626 & 32.01 & 0.097 & 2.900 & 35.53 \\
SCAN & 0.074 & 3.350 & \best{65.82} & 0.119 & \best{2.723} & 20.94 & 0.133 & 2.736 & \best{47.55} & 0.109 & 2.936 & \best{44.77} \\
CPP & 0.053 & 3.586 & 50.44 & 0.097 & 2.971 & 19.78 & 0.159 & 2.609 & 31.17 & 0.103 & 3.056 & 33.80 \\
TEMI & 0.082 & 3.331 & 55.69 & 0.100 & 2.996 & 19.15 & 0.147 & 3.075 & 30.50 & 0.110 & 3.134 & 35.11 \\
PRO-DSC & 0.072 & 3.463 & 53.85 & 0.090 & 3.247 & 18.36 & 0.168 & 2.609 & 32.05 & 0.110 & 3.106 & 34.75 \\
\midrule
\multicolumn{13}{c}{\cellcolor{gray!40}\textbf{Language-assisted Image Clustering}} \\
SIC & 0.058 & 3.811 & 50.14 & 0.105 & 2.967 & 19.51 & 0.146 & 2.718 & 30.63 & 0.103 & 3.165 & 33.43 \\
TAC & 0.071 & 3.576 & 51.83 & 0.106 & 3.000 & 18.94 & 0.169 & 2.635 & 30.52 & 0.115 & 3.070 & 33.77 \\
TAC$^*$ & 0.073 & 3.614 & 51.89 & 0.083 & 3.457 & 18.25 & 0.157 & 2.916 & 30.30 & 0.104 & 3.329 & 33.48 \\
SAC & 0.068 & 3.665 & 50.76 & 0.100 & 3.117 & 18.66 & 0.155 & 2.930 & 29.39 & 0.107 & 3.237 & 32.94 \\
GradNorm & 0.093 & 3.217 & 55.93 & 0.129 & 2.757 & 20.61 & 0.200 & 2.396 & 33.40 & 0.141 & 2.790 & 36.64 \\
NTK-SC & 0.079 & 3.354 & 53.24 & 0.137 & 2.761 & 20.39 & 0.206 & 2.428 & 33.08 & 0.140 & 2.848 & 35.57 \\
SEIC & 0.065 & 3.702 & 49.03 & 0.089 & 3.101 & 18.30 & 0.142 & 2.907 & 29.04 & 0.099 & 3.237 & 32.12 \\
MAGIC & 0.044 & 6.664 & 48.05 & 0.022 & 5.651 & 14.67 & 0.041 & 4.570 & 21.91 & 0.036 & 5.628 & 28.21 \\
\bottomrule
\end{tabular}
}
\end{table*}

\begin{table*}[tb]
\centering
\caption{Discriminability analysis on large-scale datasets using CLIP ViT-B/32 pretrained on LAION-400M. $\uparrow$ indicates larger values are better and vice versa. The best value in each metric column is highlighted in \colorbox[HTML]{FFF2CC}{\textbf{bold}}.}
\label{tab:laion400m_b32_label_free_large_scale}
\resizebox{0.8\textwidth}{!}{%
\begin{tabular}{l|ccc|ccc|ccc}
\toprule
Dataset & \multicolumn{3}{c|}{Places-365} & \multicolumn{3}{c|}{ImageNet-1K} & \multicolumn{3}{c}{Average} \\
Metric & SIL$\uparrow$ & DBI$\downarrow$ & CHI$\uparrow$ & SIL$\uparrow$ & DBI$\downarrow$ & CHI$\uparrow$ & SIL$\uparrow$ & DBI$\downarrow$ & CHI$\uparrow$ \\
\midrule
\multicolumn{10}{c}{\cellcolor{gray!40}\textbf{Classical Image Clustering}} \\
$k$-means & \best{0.070} & 3.303 & \best{58.90} & \best{0.047} & 3.162 & 39.10 & \best{0.058} & 3.232 & \best{49.00} \\
SC & 0.039 & \best{3.134} & 52.33 & 0.038 & \best{3.032} & \best{44.77} & 0.038 & \best{3.083} & 48.55 \\
SSC-OMP & -0.050 & 5.137 & 35.44 & -0.143 & 4.939 & 21.12 & -0.096 & 5.038 & 28.28 \\
EnSC & 0.013 & 3.536 & 50.94 & -0.021 & 3.390 & 34.26 & -0.004 & 3.463 & 42.60 \\
\midrule
\multicolumn{10}{c}{\cellcolor{gray!40}\textbf{Deep Image Clustering}} \\
IDC & 0.052 & 3.292 & 57.88 & 0.024 & 3.202 & 38.12 & 0.038 & 3.247 & 48.00 \\
SCAN & 0.044 & 3.383 & 57.17 & 0.022 & 3.332 & 40.37 & 0.033 & 3.358 & 48.77 \\
CPP & 0.001 & 3.500 & 51.00 & -0.056 & 3.491 & 27.54 & -0.028 & 3.495 & 39.27 \\
TEMI & 0.040 & 3.527 & 54.56 & 0.019 & 3.389 & 37.19 & 0.029 & 3.458 & 45.87 \\
PRO-DSC & -0.040 & 4.233 & 40.54 & -0.063 & 3.874 & 28.28 & -0.051 & 4.054 & 34.41 \\
\midrule
\multicolumn{10}{c}{\cellcolor{gray!40}\textbf{Language-assisted Image Clustering}} \\
SIC & 0.012 & 3.655 & 52.16 & 0.035 & 3.145 & 40.06 & 0.023 & 3.400 & 46.11 \\
TAC & 0.043 & 3.422 & 55.14 & 0.006 & 3.408 & 36.49 & 0.025 & 3.415 & 45.82 \\
TAC$^\dagger$ & 0.021 & 3.739 & 50.76 & -0.003 & 3.696 & 34.72 & 0.009 & 3.717 & 42.74 \\
SAC & 0.021 & 3.718 & 49.89 & 0.001 & 3.660 & 34.80 & 0.011 & 3.689 & 42.34 \\
GradNorm & 0.044 & 3.460 & 54.34 & 0.009 & 3.429 & 36.09 & 0.027 & 3.444 & 45.22 \\
NTK-SC & 0.049 & 3.492 & 56.65 & 0.006 & 3.469 & 36.98 & 0.028 & 3.480 & 46.82 \\
SEIC & 0.009 & 3.946 & 51.17 & -0.036 & 3.608 & 32.90 & -0.014 & 3.777 & 42.04 \\
MAGIC & 0.019 & 5.503 & 50.86 & -0.057 & 6.824 & 30.47 & -0.019 & 6.164 & 40.67 \\
\bottomrule
\end{tabular}
}
\end{table*}

\begin{table*}[tb]
\centering
\caption{Discriminability analysis on fine-grained datasets using CLIP ViT-B/32 pretrained on LAION-400M. $\uparrow$ indicates larger values are better and vice versa. The best value in each metric column is highlighted in \colorbox[HTML]{FFF2CC}{\textbf{bold}}.}
\label{tab:laion400m_b32_label_free_fine_grained}
\resizebox{\textwidth}{!}{%
\begin{tabular}{l|ccc|ccc|ccc|ccc|ccc|ccc}
\toprule
Dataset & \multicolumn{3}{c|}{Aircraft} & \multicolumn{3}{c|}{Flowers} & \multicolumn{3}{c|}{Food} & \multicolumn{3}{c|}{Cars} & \multicolumn{3}{c|}{Pets} & \multicolumn{3}{c}{Average} \\
Metric & SIL$\uparrow$ & DBI$\downarrow$ & CHI$\uparrow$ & SIL$\uparrow$ & DBI$\downarrow$ & CHI$\uparrow$ & SIL$\uparrow$ & DBI$\downarrow$ & CHI$\uparrow$ & SIL$\uparrow$ & DBI$\downarrow$ & CHI$\uparrow$ & SIL$\uparrow$ & DBI$\downarrow$ & CHI$\uparrow$ & SIL$\uparrow$ & DBI$\downarrow$ & CHI$\uparrow$ \\
\midrule
\multicolumn{19}{c}{\cellcolor{gray!40}\textbf{Classical Image Clustering}} \\
$k$-means & \best{0.059} & \best{3.136} & 20.02 & \best{0.216} & \best{2.361} & 69.61 & \best{0.123} & \best{3.034} & 148.57 & 0.124 & 2.790 & 35.86 & \best{0.117} & \best{3.228} & \best{64.55} & \best{0.128} & \best{2.910} & 67.72 \\
SC & 0.011 & 3.411 & 15.97 & 0.201 & 2.537 & 63.29 & 0.096 & 3.475 & 125.24 & 0.091 & 2.744 & 31.15 & 0.080 & 3.232 & 53.94 & 0.096 & 3.080 & 57.92 \\
SSC-OMP & -0.095 & 5.041 & 9.44 & -0.104 & 5.811 & 18.91 & -0.024 & 6.891 & 72.35 & -0.058 & 4.271 & 18.08 & -0.039 & 5.502 & 31.99 & -0.064 & 5.503 & 30.15 \\
EnSC & -0.038 & 3.717 & 15.19 & 0.137 & 2.625 & 60.39 & 0.075 & 3.350 & 133.45 & 0.074 & 2.906 & 32.10 & 0.037 & 3.329 & 52.94 & 0.057 & 3.185 & 58.82 \\
\midrule
\multicolumn{19}{c}{\cellcolor{gray!40}\textbf{Deep Image Clustering}} \\
IDC & -0.004 & 3.670 & \best{21.80} & 0.100 & 2.969 & \textbf{94.63} & 0.093 & 3.129 & 139.60 & 0.073 & 3.053 & 36.53 & 0.083 & 3.459 & 62.79 & 0.069 & 3.256 & 71.07 \\
SCAN & -0.067 & 4.110 & 15.57 & 0.157 & 2.492 & 87.20 & 0.106 & 3.189 & \best{187.25} & 0.107 & 2.987 & \best{77.83} & 0.104 & 3.252 & 62.60 & 0.081 & 3.206 & \best{86.09} \\
CPP & -0.017 & 3.839 & 16.34 & 0.169 & 2.518 & 65.11 & 0.070 & 4.030 & 126.02 & 0.094 & 2.843 & 34.63 & 0.065 & 3.357 & 58.17 & 0.076 & 3.317 & 60.05 \\
TEMI & 0.001 & 3.953 & 17.57 & -0.090 & 3.768 & 23.73 & 0.109 & 3.160 & 143.37 & 0.095 & 3.143 & 35.11 & 0.092 & 3.533 & 61.16 & 0.042 & 3.511 & 56.19 \\
PRO-DSC & -0.007 & 4.047 & 16.71 & 0.177 & 2.679 & 65.35 & 0.024 & 3.814 & 107.99 & 0.089 & 3.275 & 33.70 & 0.096 & 3.773 & 61.89 & 0.076 & 3.518 & 57.13 \\
\midrule
\multicolumn{19}{c}{\cellcolor{gray!40}\textbf{Language-assisted Image Clustering}} \\
SIC & 0.029 & 3.589 & 18.39 & 0.141 & 2.837 & 58.58 & 0.108 & 3.240 & 143.21 & 0.098 & 3.034 & 34.64 & 0.096 & 3.422 & 61.43 & 0.094 & 3.224 & 63.25 \\
TAC & 0.053 & 3.148 & 19.92 & 0.213 & 2.433 & 69.67 & 0.076 & 3.543 & 127.90 & 0.129 & 2.779 & 35.95 & 0.089 & 3.604 & 60.12 & 0.112 & 3.102 & 62.71 \\
TAC$^*$ & -0.029 & 4.286 & 15.91 & 0.175 & 2.758 & 61.35 & 0.080 & 3.551 & 128.02 & 0.056 & 3.730 & 30.62 & 0.091 & 4.058 & 60.87 & 0.075 & 3.677 & 59.35 \\
SAC & -0.006 & 4.052 & 16.82 & 0.149 & 2.817 & 54.69 & 0.088 & 3.408 & 131.37 & 0.083 & 3.390 & 32.91 & 0.092 & 3.600 & 60.55 & 0.081 & 3.453 & 59.27 \\
GradNorm & 0.048 & 3.165 & 19.33 & 0.214 & 2.444 & 68.47 & 0.107 & 3.232 & 142.89 & 0.119 & 2.793 & 35.52 & 0.108 & 3.242 & 63.81 & 0.119 & 2.975 & 66.00 \\
NTK-SC & 0.053 & 3.140 & 20.03 & 0.173 & 2.526 & 66.72 & 0.109 & 3.067 & 142.99 & \best{0.137} & \best{2.700} & 37.16 & 0.095 & 3.516 & 61.30 & 0.113 & 2.990 & 65.64 \\
SEIC & -0.036 & 4.229 & 15.81 & 0.074 & 3.051 & 51.83 & 0.079 & 3.510 & 129.80 & 0.035 & 3.616 & 29.68 & 0.038 & 3.719 & 60.41 & 0.038 & 3.625 & 57.50 \\
MAGIC & -0.060 & 6.004 & 12.11 & -0.107 & 4.588 & 26.46 & 0.070 & 6.220 & 127.78 & -0.069 & 6.010 & 18.02 & 0.059 & 5.364 & 55.90 & -0.021 & 5.637 & 48.05 \\
\bottomrule
\end{tabular}
}
\end{table*}

\section{More Discussion}
\label{appendix_H}
\subsection{Future Direction}
Despite the progress demonstrated by existing methods, our benchmark identifies several directions for future research.

\textbf{Fine-grained semantic discrimination.} Our findings show that the advantages of language-assisted image clustering become less consistent on fine-grained datasets. This suggests that textual semantics do not always capture the subtle distinctions required to separate visually similar categories. Future research should investigate how to combine detailed visual cues with more discriminative linguistic knowledge, including dynamically generated descriptions, to improve fine-grained clustering.

\textbf{Robustness against adversarial perturbations.} Language assistance generally improves post-attack clustering performance but does not systematically reduce sensitivity to adversarial perturbations. Understanding how perturbations affect visual representations and their alignment with textual semantics remains an open problem. Developing methods that explicitly preserve semantic consistency and stable cluster assignments under adversarial conditions is an important research direction.

\textbf{Generalization across diverse distribution shifts.} Although LaIC methods generally exhibit stronger generalization, no single method consistently performs well across all evaluated distribution shifts. Future research should explore how to retain transferable semantic information while reducing reliance on distribution-specific cues. Extending evaluation to long-tailed datasets and specialized domains, such as medical imaging, would further clarify the applicability and limitations of existing approaches.

\textbf{Efficient and scalable language-assisted clustering.} Our experiments indicate that LaIC offers a favorable effectiveness–efficiency trade-off, yet its performance advantages become less consistent at scale. How to preserve the benefits of language assistance as dataset size and category diversity increase remains an open question. Future methods should jointly address computational cost and clustering quality, enabling efficient semantic integration without sacrificing discrimination among increasingly diverse categories.

\subsection{Limitation and Outlook}

\texttt{VLM4Cluster} has several limitations that we aim to address in future work. First, our current evaluation primarily focuses on class-balanced natural image datasets. Second, the current framework relies on fixed lexical databases (e.g., WordNet) to obtain language assistance, which, although a de facto choice in many LaIC methods, may limit semantic coverage and adaptability, particularly for specialized or emerging concepts.

Apart from maintaining and improving the codebase, we plan to further enhance \texttt{VLM4Cluster} in the following ways. To begin with, we plan to broaden our dataset coverage to ensure a more comprehensive evaluation. For example, we will look into long-tailed image datasets and medical image datasets. Secondly, we will plan to incorporate more flexible and dynamically generated linguistic knowledge, such as that provided by MLLMs.

\section{AI use statement} 
We use ChatGPT for grammar and spelling checks only, with prompt ”Proofread the sentences”.

\section{Ethics statement} 
Our study relies solely on publicly available datasets and models. No private or personally identifiable information was used. The work aims to advance the scientific understanding of image clustering while upholding principles of transparency, fairness, and responsible research.

\section{Reproducibility Statement} 
All pre-trained VLMs used in this paper are publicly available. To further support reproducibility, we have released the \texttt{VLM4Cluster} codebase at \href{https://github.com/YuanweiHuu/VLM4Cluster}{\textcolor{red}{here}}, which includes the evaluation protocols, baseline methods, datasets, and scripts needed to reproduce our results.

\end{document}